\documentclass{article}

\usepackage{float} 
\usepackage{PRIMEarxiv}
\usepackage{xcolor}
\usepackage{multirow}
\usepackage{adjustbox}
\usepackage[utf8]{inputenc} 
\usepackage[T1]{fontenc}    
\usepackage[hidelinks]{hyperref}       \hypersetup{
    colorlinks=true,
    filecolor=magenta,      
    pdftitle={Overleaf Example},
    pdfpagemode=FullScreen,
    }
\usepackage{url}            
\usepackage{booktabs}       
\usepackage{amsfonts}       
\usepackage{nicefrac}       
\usepackage{microtype}      
\usepackage{lipsum}
\usepackage{fancyhdr}       
\usepackage{graphicx}       
\usepackage{cite}
\graphicspath{{media/}}     

\usepackage{caption}
\usepackage{subcaption}
\usepackage{array}

\usepackage[most]{tcolorbox}

\definecolor{ForestGreen}{RGB}{34,139,34}

\newcommand{\fboxgreen}{\setlength{\fboxrule}{1.5pt} \setlength{\fboxsep}{0pt}\fcolorbox{ForestGreen}{white}}

\newcommand{\fboxred}{\setlength{\fboxrule}{1.5pt} \setlength{\fboxsep}{0pt}\fcolorbox{red!70!white}{white}}

\title{
A Visual Tour Of Current Challenges In \\
Multimodal Language Models 
}

\author{
  Shashank Sonkar, Naiming Liu, Richard G. Baraniuk \\
  Rice University \\
  \texttt{\{ss164, nl35, richb\}@rice.edu} \\
}

\begin{document}
\maketitle

\begin{abstract}
Transformer models trained on massive text corpora have become the de facto models for a wide range of natural language processing tasks.
However, learning effective word representations for {\em function words} remains challenging.
Multimodal learning, which visually grounds transformer models in imagery, can overcome the challenges to some extent; however, there is still much work to be done.
In this study, we explore the extent to which visual grounding facilitates the acquisition of function words using stable diffusion models that employ multimodal models for text-to-image generation.
Out of seven categories of function words, along with numerous subcategories, we find that stable diffusion models effectively model only a small fraction of function words -- a few pronoun subcategories and relatives. 
We hope that our findings will stimulate the development of new datasets and approaches that enable multimodal models to learn better representations of function words.

\end{abstract}

\section{Introduction}
Transformer models \cite{bert,roberta} are currently state-of-the-art across many natural language processing (NLP) tasks such as question answering \cite{luke,xlnet}, information retrieval \cite{formal2021splade}, inference \cite{Wang2021EntailmentAF,albert}, and machine translation \cite{vaswani2017attention,al2019character}.
Transformers are masked language models which use the self-attention mechanism \cite{vaswani2017attention} to output contextualized word embeddings.
However, not all words can be modeled effectively using context information \cite{asher2018content,baroni2012entailment,bernardi2013relatedness,hermann2013not,linzen2016quantificational}. 
Function words like conjunctions, pronouns, prepositions etc are difficult to learn using the masked language modeling loss \cite{kim2019probing,chaves2021look}.

An alternative method to learn the representations of function words is to use multimodal learning to ground the language models visually in natural images.
These multimodal language models (MLMs) \cite{lee2018stacked, clip, jia2021scaling} learn an aligned representation of images and text.
Recently, stable diffusion models (SDMs) \cite{sdm} have gained popularity for text-to-image generation.
SDMs take a natural language prompt as input, encode the prompt using a MLM, and then generate an image capturing the semantics of the prompt.

{\em The key point of this short paper is that SDMs can be used to gain new and useful insights into the workings of MLMs.}
In our study, we use carefully crafted prompts with seven different categories of function words and their sub-categories \cite{klammer2007analyzing} to probe SDMs. 
Next, we visually inspect whether the images capture the semantics of the function words.
Despite the MLM in SDMs being visually grounded, we discover that, for the majority of function words, the images generated do not accurately convey the meaning entailed by the prompts.

Our findings can inspire innovative research in the construction of datasets to improve MLMs' understanding of function words, which are fundamental building blocks of English grammar. 
We also provide the \href{https://github.com/lucy66666/abstract-diffusion}{code} on github for readers to replicate our findings and explore further. 

\section{Background}
\begin{figure}
    \centering
    \includegraphics[width=0.9\textwidth]{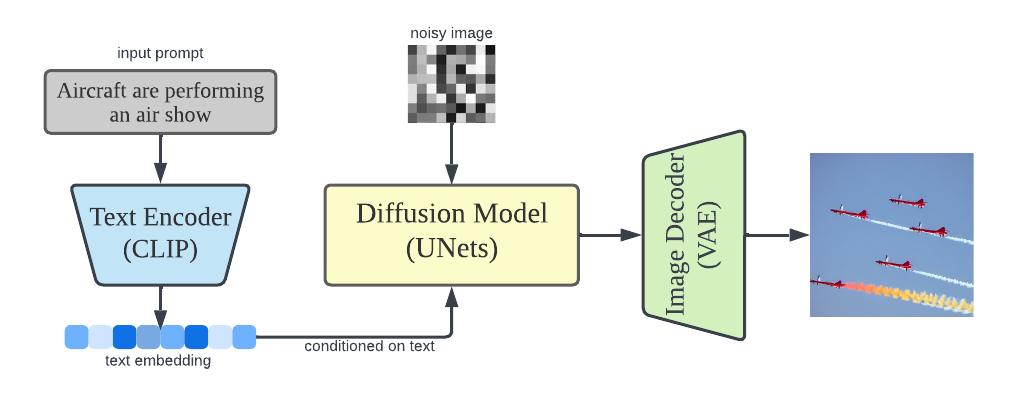}
    \caption{A Stable Diffusion Model (SDM) architecture \cite{sdm} has three main components: a frozen CLIP ViT-L/14 as text encoder, UNets as diffusion model, and variational autoencoders (VAE) as image decoder. The text encoder embeds the input prompt into a high-dimensional representation, which is fed to the diffusion model together with a noisy sample. Then, the image decoder converts the diffusion model's latent space image representation to a real image that captures the semantics of the input prompt.}
    \label{fig:sdm}
\end{figure}
\subsection{Stable Diffusion Model}
Stable diffusion model (SDM) is a trending, open source text-to-image generation model which utilizes latent diffusion model conditioned on the text embeddings. As shown in figure \ref{fig:sdm}, SDM is composed of three main components: 1) a text encoder (a frozen CLIP ViT-L/14 \cite{clip}), 2) a diffusion model (U-Nets \cite{unet}) and 3) an image decoder (variational autoencoder \cite{VAE}). The text encoder takes a natural language prompt as input and transforms it into a high-dimensional embedding using the self-attention mechanism \cite{vaswani2017attention}. Then, using this embedding and noise sample as input, the diffusion model and the image decoder output the target image.

\subsection{Multimodal Language Models}
We discuss a few specifics of multimodal language models (MLMs) in this section. We focus on CLIP, a type of MLM, since it is used by the SDM that we use for our experiments. CLIP aligns images and their textual descriptions to embed text and image in the same vector space. CLIP contains a text encoder and an image encoder. Both are optimized simultaneously using the principle of contrastive learning \cite{constractive-learning} by maximizing the cosine similarity between paired text and image embeddings while minimizing the cosine similarity between unpaired ones. CLIP representations have many applications including visual question-answering \cite{shen2021much}, automatic image captioning \cite{mokady2021clipcap} and object navigation \cite{khandelwal2022simple}.

\subsection{Linguistics Review: Function Words}

Words can be broadly classified into two categories --- function words like determiners, prepositions, pronouns, etc and content words like nouns, verbs, adjectives, etc.
It is hard to capture the semantics of content words as compared to function words using the contextual information \cite{asher2018content,baroni2012entailment,bernardi2013relatedness,hermann2013not,linzen2016quantificational}.
Since language models using context to predict the missing/next word in the pre-training objectives, they have been shown to perform poorly on function words \cite{kim2019probing,chaves2021look}. We believe that grounding text encoder in natural images can alleviate these shortcomings, but the question remains to what extent. As mentioned before, we primarily focus on the CLIP text encoder and explore how visually-grounded language model performs as text encoder of SDM when it comes to modeling functions words.

\begin{table}[t!]
\begin{minipage}[t]{0.48\textwidth}
\centering
\begin{tabular}{ccc}
\toprule
\textbf{Category} & \textbf{Sub-category} & \textbf{Examples} \\ 
\midrule
\multicolumn{1}{l}{\multirow{5}{*}{\textbf{Pronouns}}} & Subject & he, she, they \\ 
\multicolumn{1}{l}{} & Object & him, her, them \\ 
\multicolumn{1}{l}{} & Possessive & his, her, our \\ 
\multicolumn{1}{l}{} & Indefinite & nobody, everyone \\ 
\multicolumn{1}{l}{} & Reflexive & himself, herself \\ 
\midrule
\multicolumn{1}{l}{\textbf{Conjunctions}} &  & and, but, yet, or \\ \midrule
\multicolumn{1}{l}{\textbf{Interrogatives}} &  & who, which, where \\ \bottomrule
\end{tabular}
\end{minipage} \hfill
\begin{minipage}[t]{0.5\textwidth}
\begin{tabular}{ccc}
\toprule
\textbf{Category} & \textbf{Sub-category} & \textbf{Examples} \\ 
\midrule
\multicolumn{1}{l}{\multirow{3}{*}{\textbf{Determiners}}} & Article & a, an, the \\ 
\multicolumn{1}{l}{} & Numeral & one, two, ten \\ 
\multicolumn{1}{l}{} & Quantifier & little, many, few \\ 
\midrule
\multicolumn{1}{l}{\textbf{Qualifiers}} &  & not, always, very \\ \midrule
\multicolumn{1}{l}{\multirow{3}{*}{\textbf{Prepositions}}} & Place & in, on, under \\ 
\multicolumn{1}{l}{} & Movement & up, down, towards \\ 
\multicolumn{1}{l}{} & Particle & on, off, with \\ 
\bottomrule
\end{tabular}
\end{minipage}
\vspace{3mm}
\caption{Categories of function words with examples. For each category, we carefully design prompts to probe stable diffusion models to explore their understanding of function words.}
\label{function-words}
\vspace{-5mm}
\end{table}

\section{Experiments}
\label{ss:exps}
In this section, we divide function words into seven categories (listed in table~\ref{function-words}) and visually inspect each category to check if it can be modeled through SDMs.
We list out language prompts used to probe SDM\footnote{We use "\textit{Stable Diffusion v1-4}" model released at \url{https://huggingface.co/CompVis/stable-diffusion-v1-4}} for each category and present a figure that contains multiple images that spans all its subcategories. Green/red border around the images are used to identify if SDM successfully/unsuccessfully outputs an image that captures the semantics of the input prompt.

Note that we provide only a sample of images in the experiment section. Please refer to the appendix for more samples.

\begin{figure}
     \centering
     \begin{subfigure}[b]{0.32\textwidth}
         \centering
         \fboxgreen{\includegraphics[width=\textwidth]{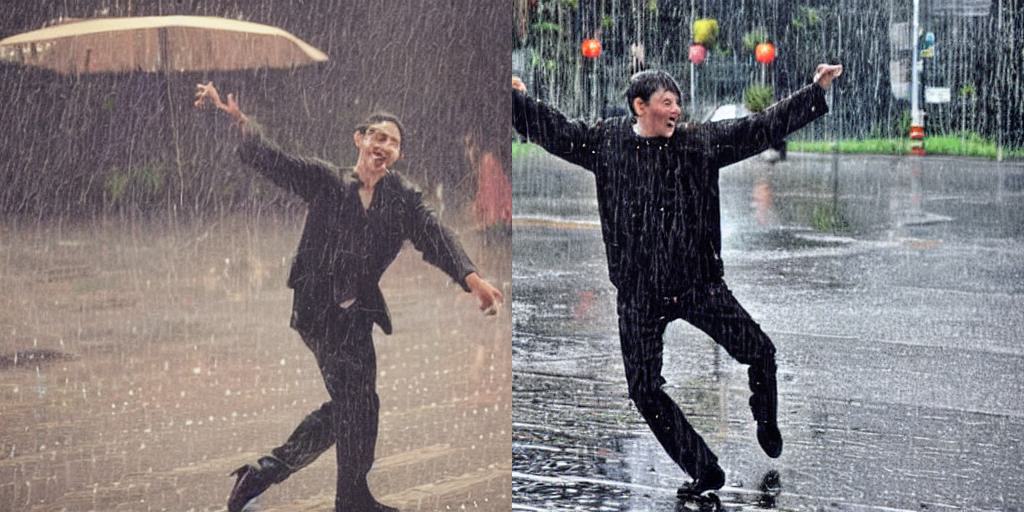}}
         \caption{\textit{He} is dancing in the rain}
         \label{fig:he}
     \end{subfigure}
     \hfill
     \begin{subfigure}[b]{0.32\textwidth}
         \centering
         \fboxgreen{\includegraphics[width=\textwidth]{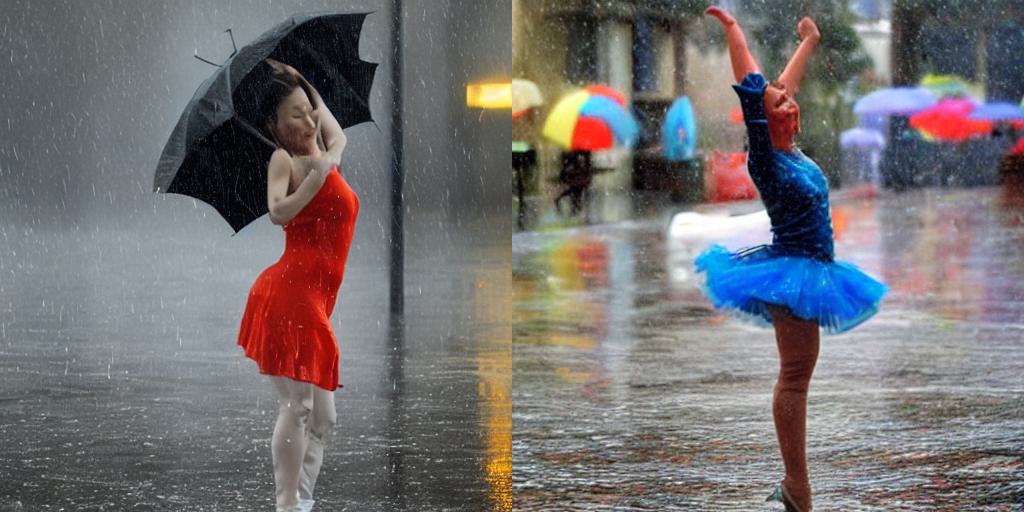}}
         \caption{\textit{She} is dancing in the rain}
         \label{fig:she}
     \end{subfigure}
     \hfill
     \begin{subfigure}[b]{0.32\textwidth}
         \centering
         \fboxgreen{\includegraphics[width=\textwidth]{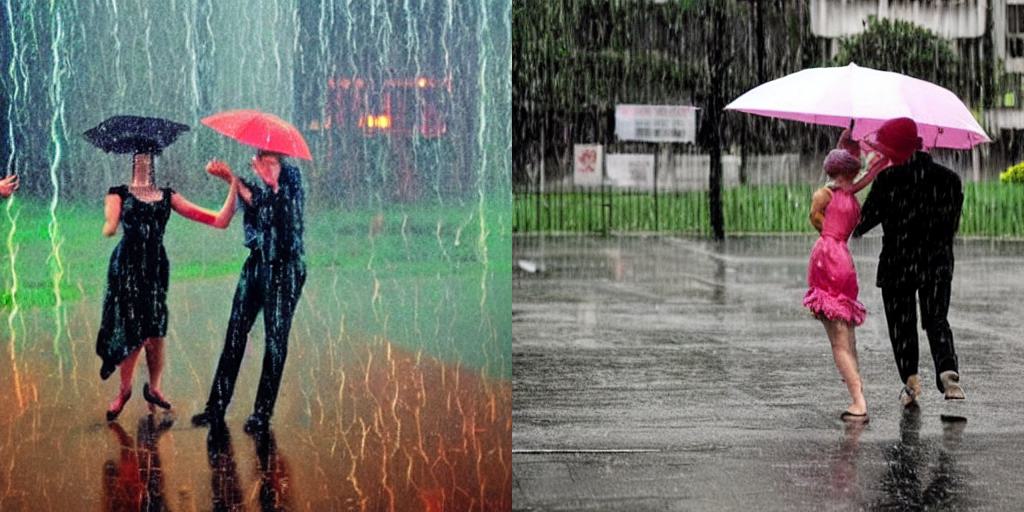}}
         \caption{\textit{We} are dancing in the rain}
         \label{fig:we}
     \end{subfigure}
    \centering
     \begin{subfigure}[b]{0.32\textwidth}
         \centering
         \fboxred{\includegraphics[width=\textwidth]{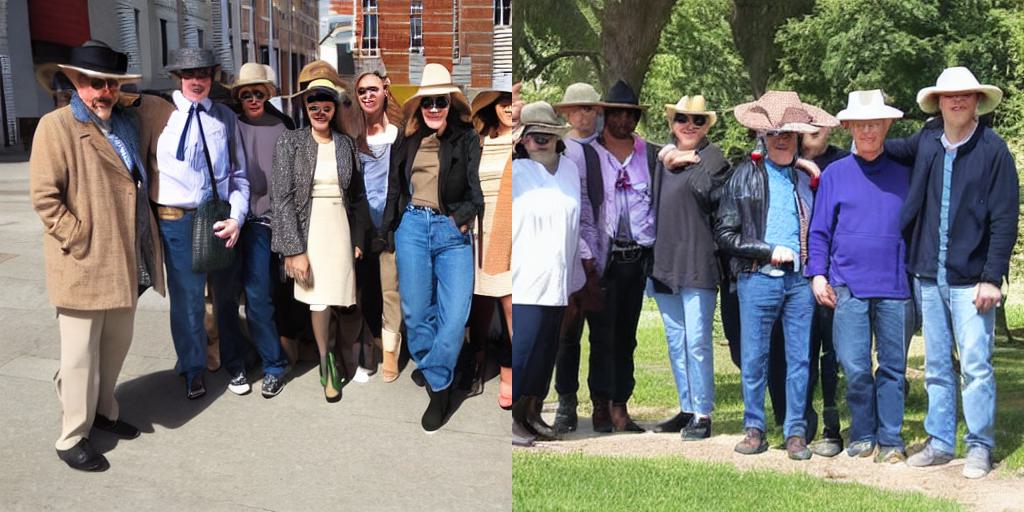}}
         \caption{\textit{Nobody} in the group is wearing a hat}
         \label{fig:nobody}
     \end{subfigure}
     \hfill
     \begin{subfigure}[b]{0.32\textwidth}
         \centering
         \fboxred{\includegraphics[width=\textwidth]{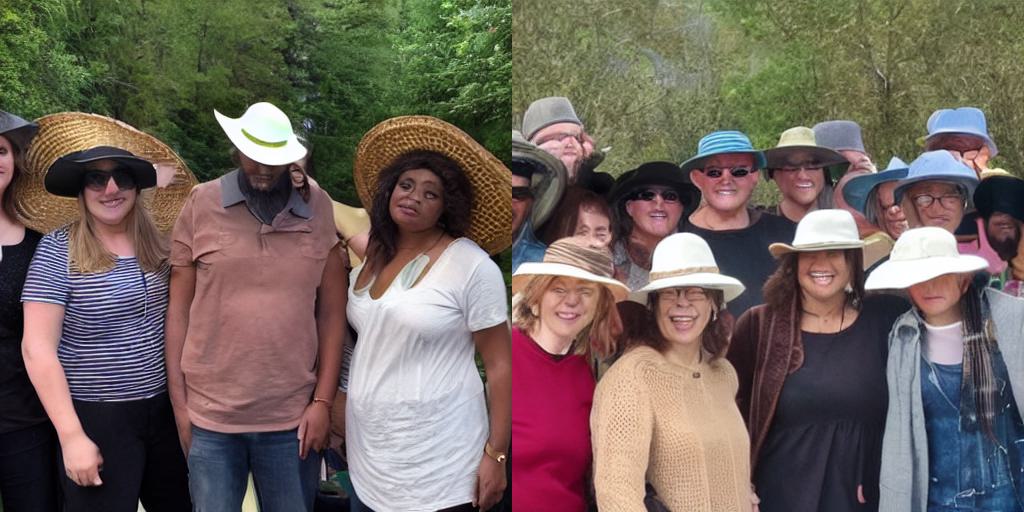}}
         \caption{\textit{Some} in the group are wearing a hat}
         \label{fig:some}
     \end{subfigure}
     \hfill
     \begin{subfigure}[b]{0.32\textwidth}
         \centering
         \fboxgreen{\includegraphics[width=\textwidth]{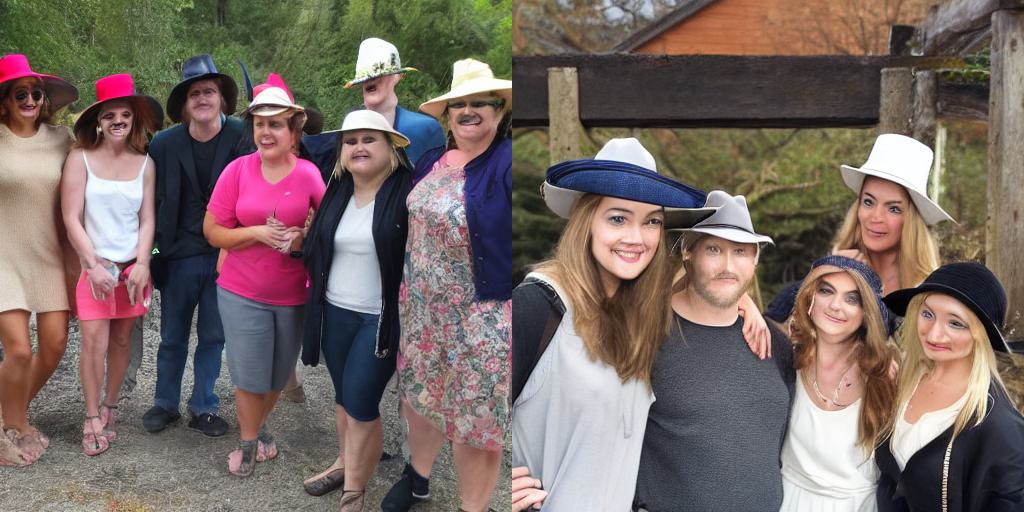}}
         \caption{\textit{Everyone} in the group is wearing a hat}
         \label{fig:everyone}
     \end{subfigure}
    \centering
     \begin{subfigure}[b]{0.32\textwidth}
         \centering
         \fboxred{\includegraphics[width=\textwidth]{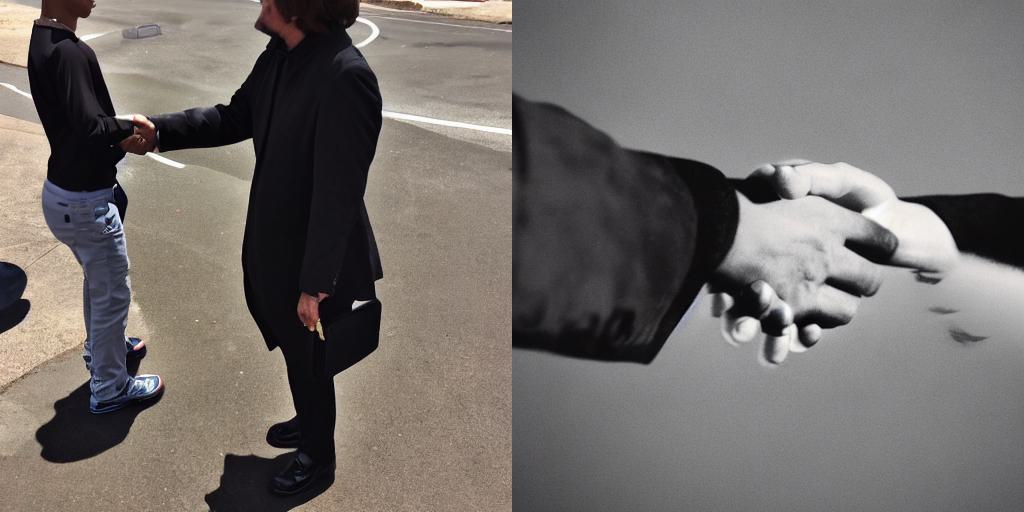}}
         \caption{I shook hands with \textit{myself}}
         \label{fig:myself}
     \end{subfigure}
     \hfill
     \begin{subfigure}[b]{0.32\textwidth}
         \centering
         \fboxred{\includegraphics[width=\textwidth]{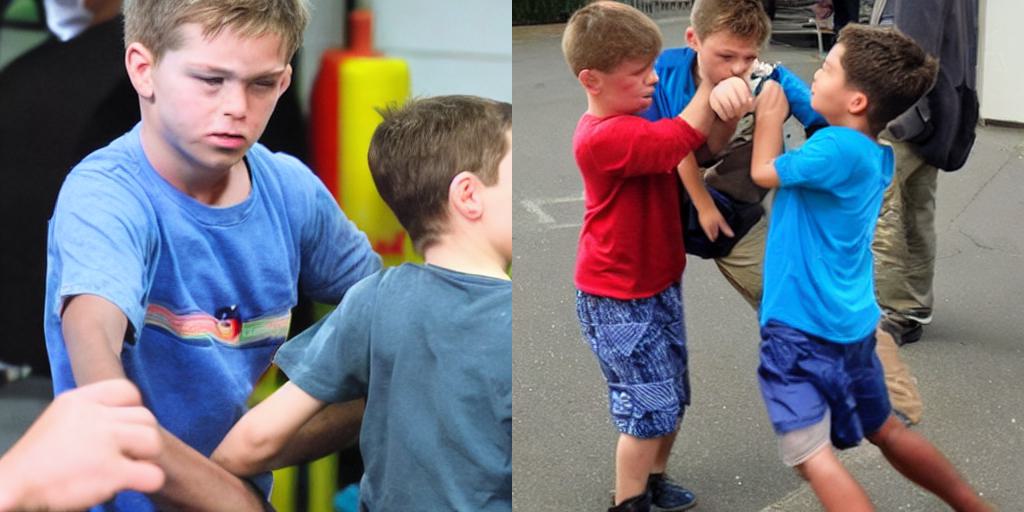}}
         \caption{The boy punched \textit{himself} in the face}
         \label{fig:himself}
     \end{subfigure}
     \hfill
     \begin{subfigure}[b]{0.32\textwidth}
         \centering
         \fboxred{\includegraphics[width=\textwidth]{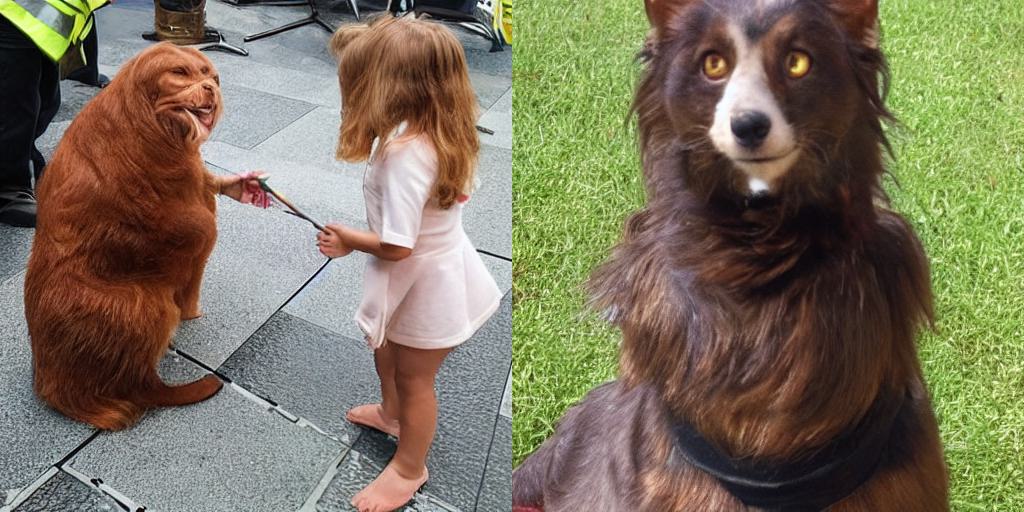}}
         \caption{She patted \textit{herself} for a job well done}
         \label{fig:herself}
     \end{subfigure}
     \caption{Sample images depicting SDM's success (green border) and failure (red border) in capturing the semantics of different subcategories of {\bf pronouns}. (a)--(c) show that the information about gender and count implicit in subject pronouns like he, she, we is accurately depicted. But, for indefinite pronouns, SDMs fail to capture the notion of negatives ((d) nobody), existenial ((e) some), and universals ((f) everyone). Likewise SDMs fail to capture the meaning of reflexive pronouns like (g) myself, (h) himself, (i) herself.}
     \label{fig:pronouns}
     \vspace{-4mm}
\end{figure}

\subsection{Pronouns}
Pronouns are used in English grammar as a substitute for nouns. They are divided into five categories: 1) subject pronouns e.g., he, she, we 2) object pronouns e.g., him, her, them 3) possessive adjectives e.g., his, her, our 4) indefinite pronouns e.g., few, many, nobody, everyone and 5) reflexive pronouns e.g., himself, herself, ourselves.

Subject pronouns, object pronouns, and possessive pronouns reflect the gender and count of the entity they refer to.
Through language prompts like ``\textit{He} is dancing in the rain', `\textit{She} is dancing in the rain', and `\textit{We} are dancing in the rain', We can test the diffusion model's ability to produce visuals that appropriately depict the gender and count of entities that each language prompt embodies.
Our experiments reveal that, for the most part, the images in figure \ref{fig:he}--\ref{fig:we} did accurately represent the gender and count.

However, that is not the case for indefinite pronouns and reflexive pronouns.
Indefinite pronouns are divided into negatives (none, no one, nobody), assertive existential (some, someone, somebody), and universals (everyone, everybody).
We probed the semantics of indefinite pronouns using prompts like `\textit{No one} in the group is wearing a hat', `\textit{Some} in the group are wearing a hat', and `\textit{Everyone} in the group is wearing a hat'.
We found that the diffusion model is not able to differentiate amongst the three subcategories of indefinite pronouns as can be seen in the figure \ref{fig:nobody}--\ref{fig:everyone}.

Reflexive pronouns make reference to the formerly mentioned noun and include words that end with which $-self$ and $-selves$.
We investigated these pronouns using prompts like `The boy punched \textit{himself} in the face', `She patted \textit{herself} for a job well done.' and `I shook hands with \textit{myself}'. We found that images in figure \ref{fig:myself}--\ref{fig:herself} did not reflect the reflexive nature of these pronouns.

\begin{figure}[t!]
     \centering
     \begin{subfigure}[b]{0.24\textwidth}
         \centering
         \fboxgreen{\includegraphics[width=\textwidth]{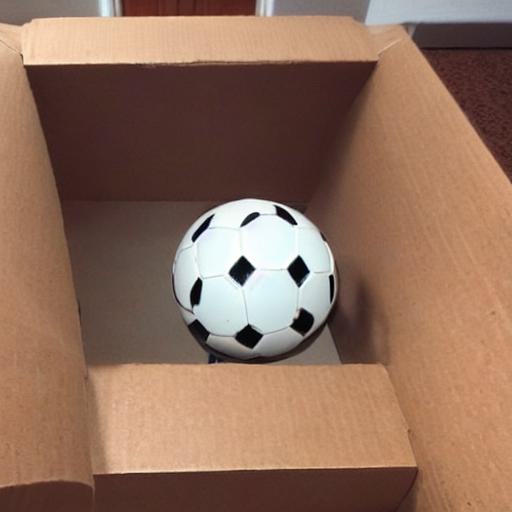}}
         \caption{The ball is \textit{in} the box}
         \label{fig:in}
     \end{subfigure}
     \hfill
     \begin{subfigure}[b]{0.24\textwidth}
         \centering
         \fboxred{\includegraphics[width=\textwidth]{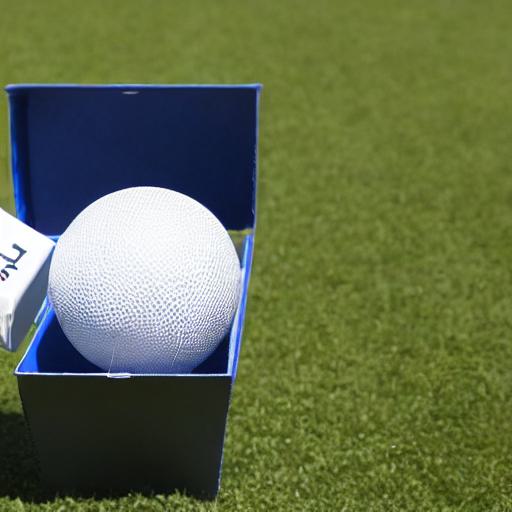}}
         \caption{The ball is \textit{on} the box}
         \label{fig:on}
     \end{subfigure}
     \hfill
     \begin{subfigure}[b]{0.24\textwidth}
         \centering
         \fboxred{\includegraphics[width=\textwidth]{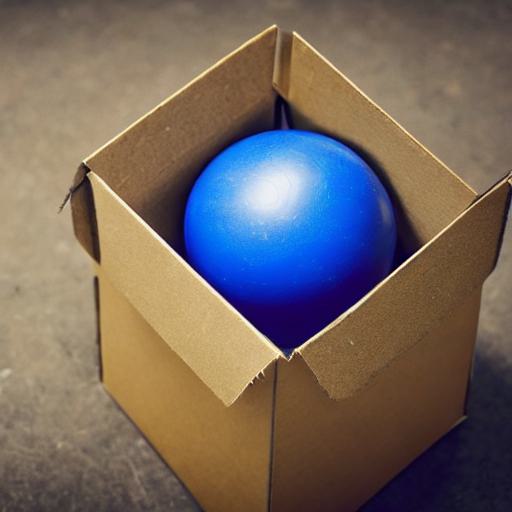}}
         \caption{The ball is \textit{next to} the box}
         \label{fig:next}
     \end{subfigure}
     \hfill
     \begin{subfigure}[b]{0.24\textwidth}
         \centering
         \fboxred{\includegraphics[width=\textwidth]{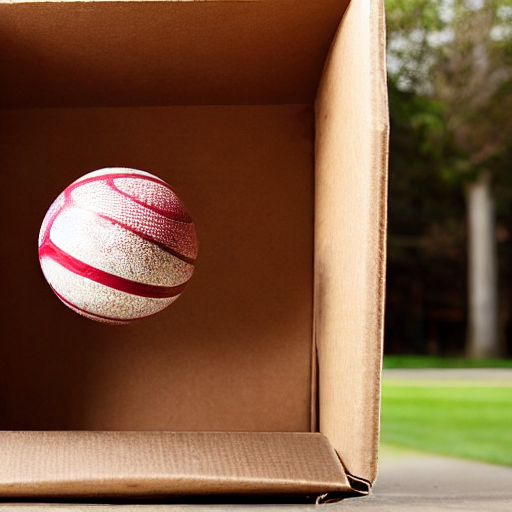}}
         \caption{The ball is \textit{behind} the box}
         \label{fig:behind}
     \end{subfigure}
    \centering
     \begin{subfigure}[b]{0.24\textwidth}
         \centering
         \fboxgreen{\includegraphics[width=\textwidth]{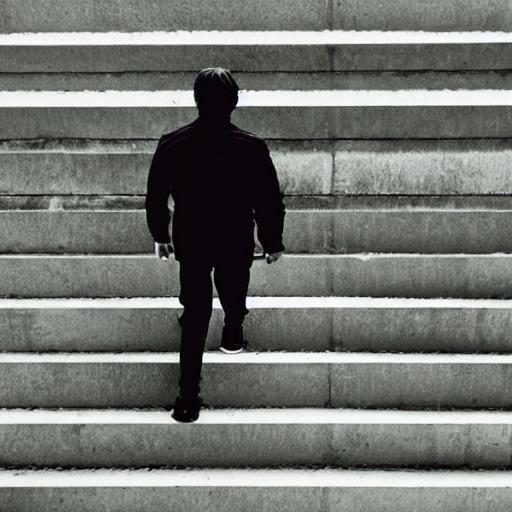}}
         \caption{He is walking \textit{up} the stairs}
         \label{fig:up}
     \end{subfigure}
     \hfill
     \begin{subfigure}[b]{0.24\textwidth}
         \centering
         \fboxred{\includegraphics[width=\textwidth]{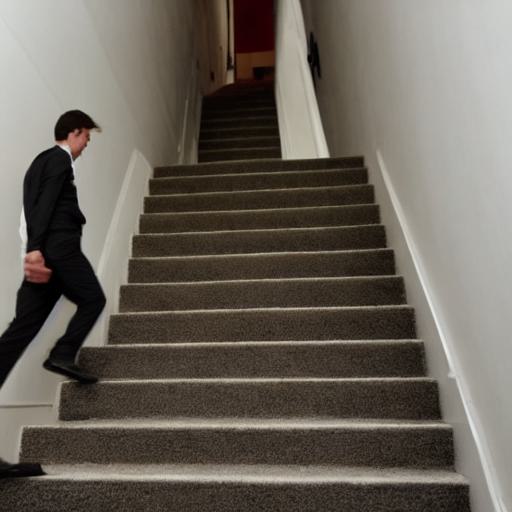}}
         \caption{He is walking \textit{down} the stairs}
         \label{fig:down}
     \end{subfigure}
     \hfill
     \begin{subfigure}[b]{0.24\textwidth}
         \centering
         \fboxred{\includegraphics[width=\textwidth]{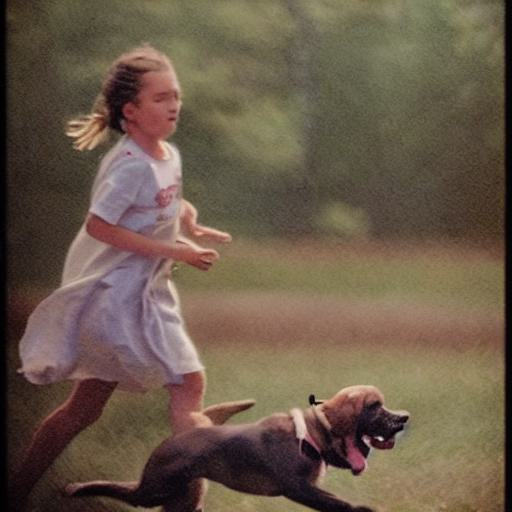}}
         \caption{The girl ran \textit{from} the dog}
         \label{fig:from}
     \end{subfigure}
     \hfill
     \begin{subfigure}[b]{0.24\textwidth}
         \centering
         \fboxred{\includegraphics[width=\textwidth]{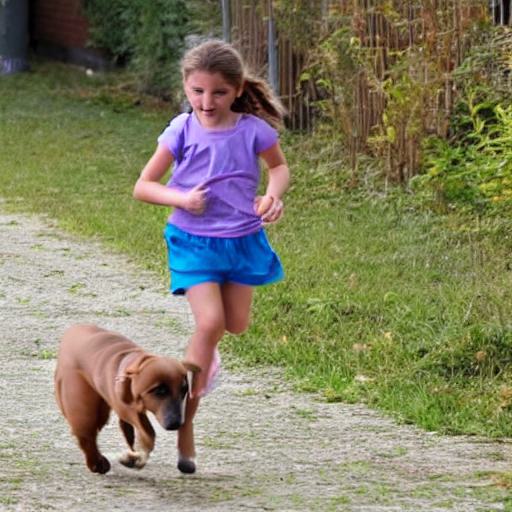}}
         \caption{The girl ran \textit{towards} the dog}
         \label{fig:towards}
     \end{subfigure}
    \centering
     \begin{subfigure}[b]{0.24\textwidth}
         \centering
         \fboxgreen{\includegraphics[width=\textwidth]{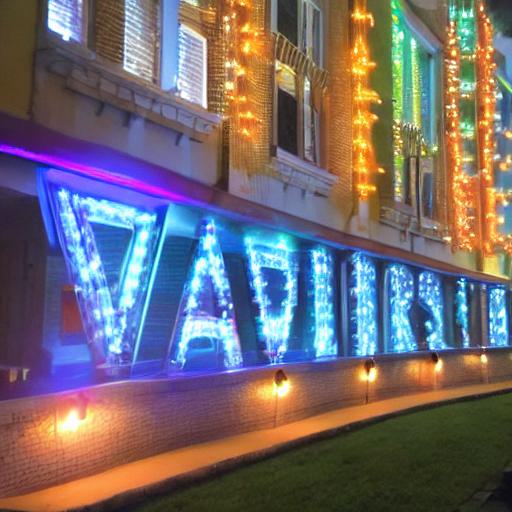}}
         \caption{The lights are \textit{on}}
         \label{fig:lightson}
     \end{subfigure}
     \hfill
     \begin{subfigure}[b]{0.24\textwidth}
         \centering
         \fboxred{\includegraphics[width=\textwidth]{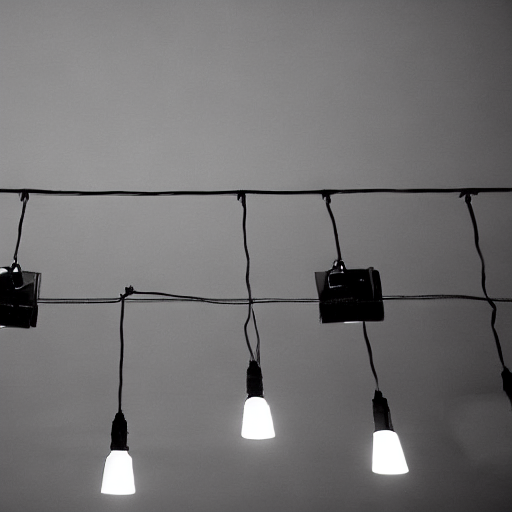}}
         \caption{The lights are \textit{off}}
         \label{fig:lightsoff}
     \end{subfigure}
     \hfill
     \begin{subfigure}[b]{0.24\textwidth}
         \centering
         \fboxred{\includegraphics[width=\textwidth]{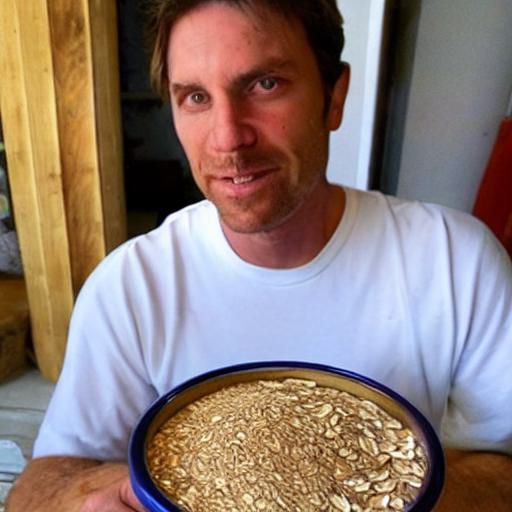}}
         \caption{He eats oats \textit{with} milk}
         \label{fig:with}
     \end{subfigure}
     \hfill
     \begin{subfigure}[b]{0.24\textwidth}
         \centering
         \fboxgreen{\includegraphics[width=\textwidth]{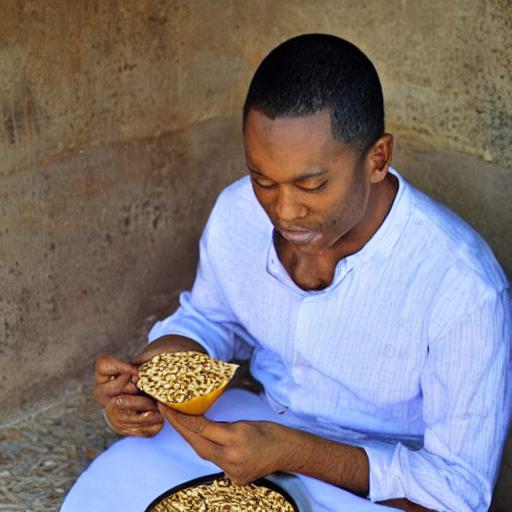}}
         \caption{He eats oats \textit{without} milk}
         \label{fig:without}
     \end{subfigure}
     \caption{Sample images depicting SDM's success (green border) and failure (red border) in capturing the semantics of different subcategories of {\bf prepositions}. Even for prepositions of place which can be learned easily through visual grounding, images for (a) in, (b) on, (c) next, and (d) behind show that SDMs do not understand this subcategory of preposition. Same analysis holds for prepositions of movement like (e) up, (f) down, (g) from, and (h) towards. Unsurprisingly, SDMs also fail for the hardest abstract category of particles, which include (i) on, (j) off, (k) with, and (l) without.}
     \label{fig:prepostions}
     \vspace{-3mm}
\end{figure}

\subsection{Prepositions and Particles}
Prepositions are an interesting category of words that are simple to illustrate with visuals since they convey spatial and temporal relationships. Despite their simplicity we noted that modeling prepositions is rather challenging for current diffusion models. Spatial relations are covered by prepositions of place e.g. in, on, under, etc. For testing preposition of place, we used prompts like `The ball is \textit{in} the box', `The ball is \textit{on} the box', `The ball is \textit{under} the box', etc.

We observed that the diffusion models were not able to differentiate amongst prepositions of place as can be seen in the figure \ref{fig:in}--\ref{fig:behind}.
Even though the model successfully outputs the image in figure~\ref{fig:in} for the prompt `The ball is \textit{in} the box', it outputs similar images for other prepositions of place \ref{fig:on}--\ref{fig:behind}, thereby raising the question --- did SDM really understand the meaning of `in' in figure \ref{fig:in}? In this paper, we assume that SDM does not in fact understand the meaning of \textit{in} in figure~\ref{fig:in} since it outputs similar images irrespective of the input.

Likewise for prepositions of movements which model temporal relations, creative prompts like `He is walking \textit{up} the stairs', He is walking \textit{down} the stairs', `The girl ran \textit{towards} the dog', `The girl ran \textit{away} from the dog' were used. In this experiment as well, we concluded from figure \ref{fig:up}--\ref{fig:towards} that the semantic properties of prepositions in questions were not accounted for.

We also consider a special type of prepositions: particles which reflect the state of an entity. Examples include on, off, with, without, etc. Using inventive prompts like `The lights are \textit{on}', `The lights are \textit{off}', `He eats oats \textit{with} milk.', `He eats oats \textit{without} milk.', we notice that diffusion models are ineffective at modeling particles either as seen in figure \ref{fig:lightson}--\ref{fig:without}.

\begin{figure}
     \centering
     \begin{subfigure}[b]{0.32\textwidth}
         \centering
         \fboxred{\includegraphics[width=\textwidth]{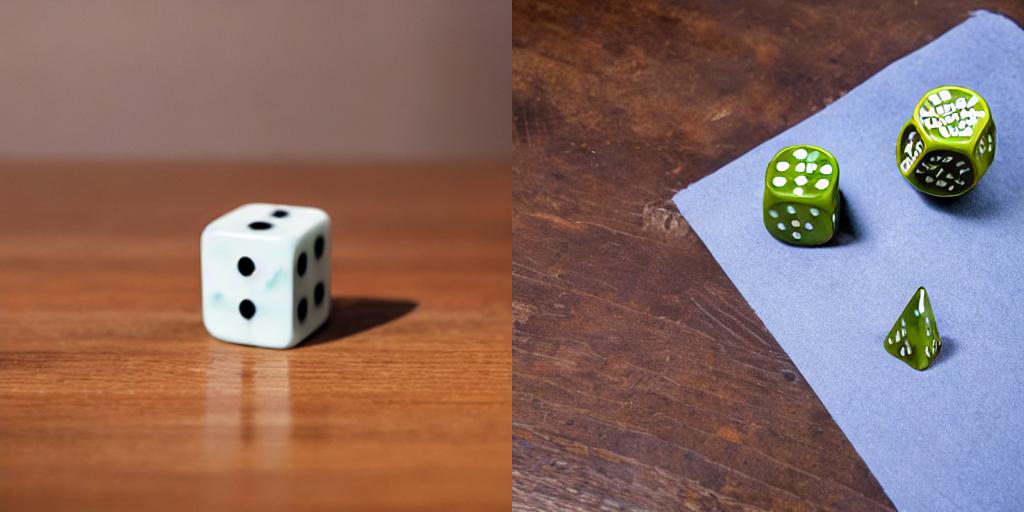}}
         \caption{\textit{A} dice rolled on the table}
         \label{fig:a}
     \end{subfigure}
     \hfill
     \begin{subfigure}[b]{0.32\textwidth}
         \centering
         \fboxgreen{\includegraphics[width=\textwidth]{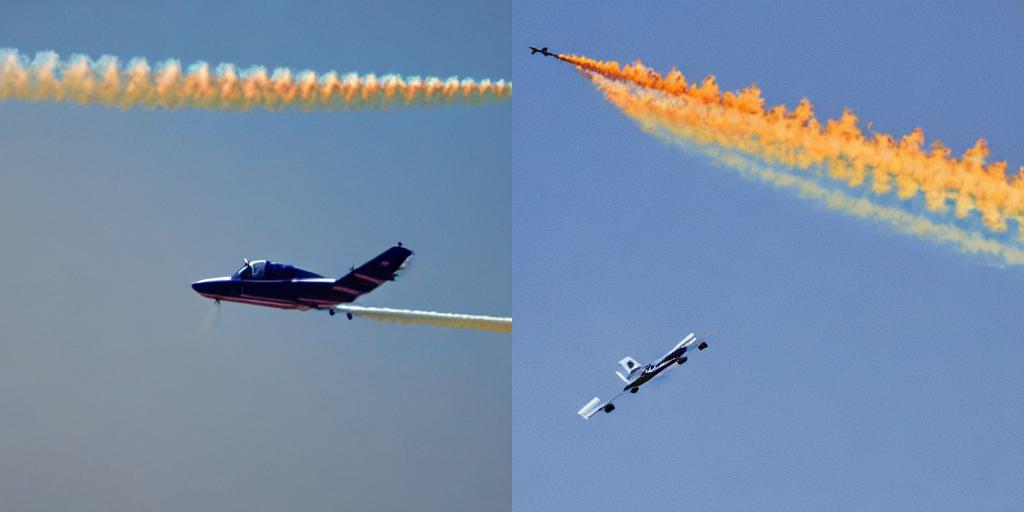}}
         \caption{\textit{An} aircraft performing an air show}
         \label{fig:an}
     \end{subfigure}
     \hfill
     \begin{subfigure}[b]{0.32\textwidth}
         \centering
         \fboxred{\includegraphics[width=\textwidth]{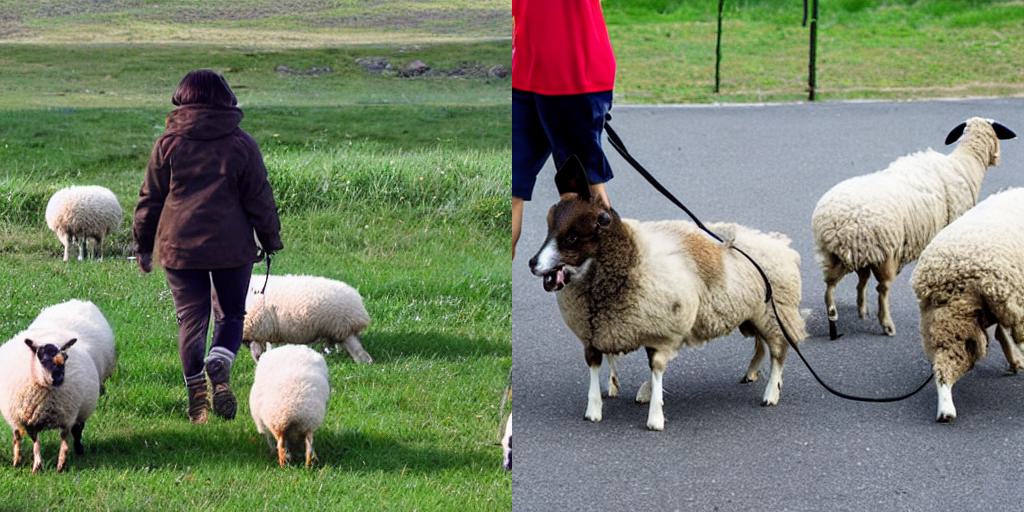}}
         \caption{The dog is guiding \textit{the} sheep}
         \label{fig:the}
     \end{subfigure}
        \label{fig:subj_pronouns}
    \centering
     \begin{subfigure}[b]{0.32\textwidth}
         \centering
         \fboxred{\includegraphics[width=\textwidth]{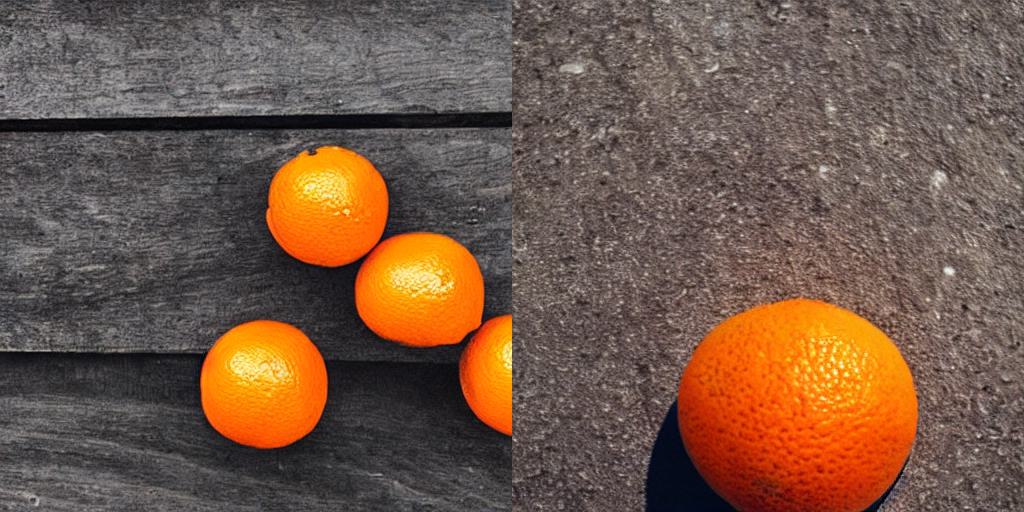}}
         \caption{There is \textit{one} orange in the photo}
         \label{fig:one}
     \end{subfigure}
     \hfill
     \begin{subfigure}[b]{0.32\textwidth}
         \centering
         \fboxred{\includegraphics[width=\textwidth]{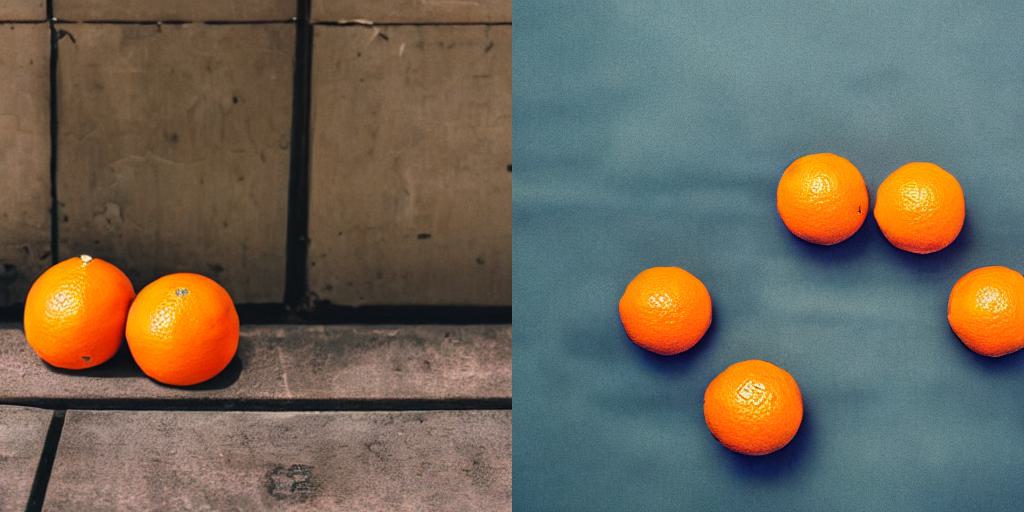}}
         \caption{There are \textit{two} oranges in the photo}
         \label{fig:two}
     \end{subfigure}
     \hfill
     \begin{subfigure}[b]{0.32\textwidth}
         \centering
         \fboxred{\includegraphics[width=\textwidth]{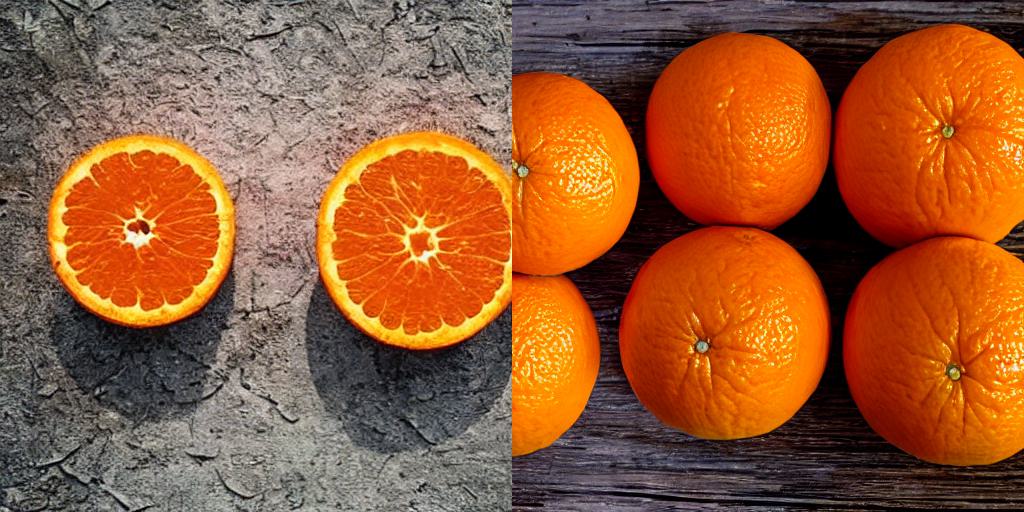}}
         \caption{There are \textit{ten} oranges in the photo}
         \label{fig:ten}
     \end{subfigure}
        \label{fig:indefinite_pronouns}
    \centering
     \begin{subfigure}[b]{0.24\textwidth}
         \centering
         \fboxred{\includegraphics[width=\textwidth]{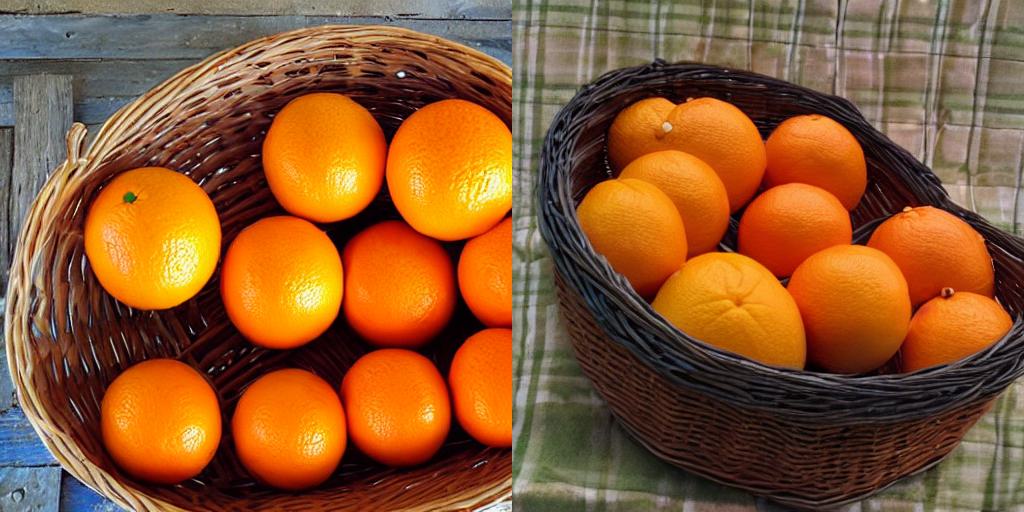}}
         \caption{\textit{Few} oranges in the basket}
         \label{fig:few}
     \end{subfigure}
     \hfill
     \begin{subfigure}[b]{0.24\textwidth}
         \centering
         \fboxgreen{\includegraphics[width=\textwidth]{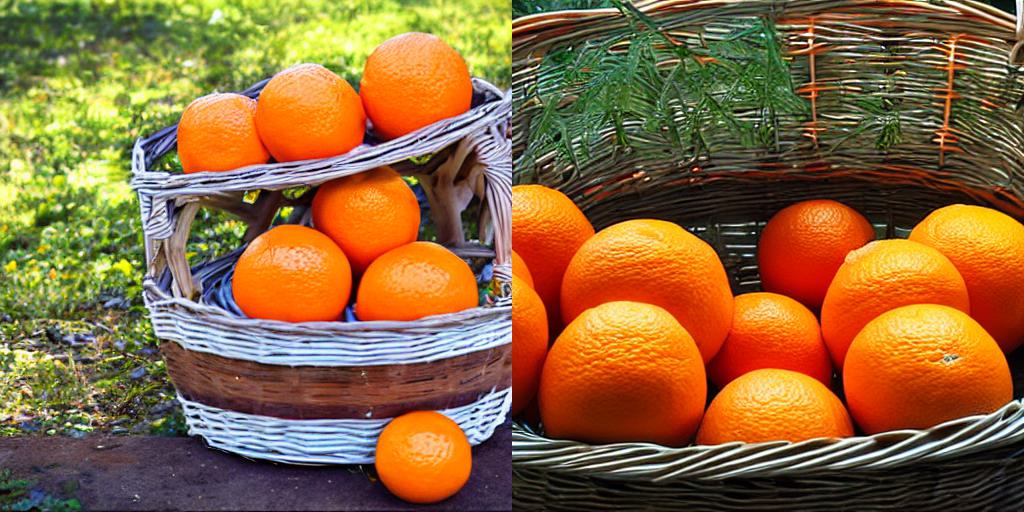}}
         \caption{\textit{Many} oranges in the basket}
         \label{fig:many}
     \end{subfigure}
     \hfill
     \begin{subfigure}[b]{0.24\textwidth}
         \centering
         \fboxgreen{\includegraphics[width=\textwidth]{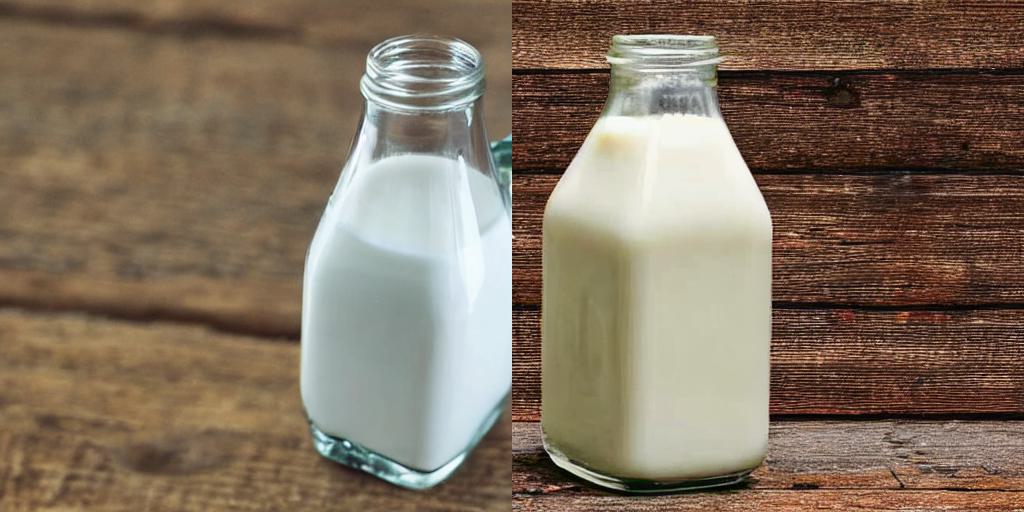}}
         \caption{\textit{Little} milk in the bottle}
         \label{fig:littlemilk}
     \end{subfigure}
     \hfill
     \begin{subfigure}[b]{0.24\textwidth}
         \centering
         \fboxred{\includegraphics[width=\textwidth]{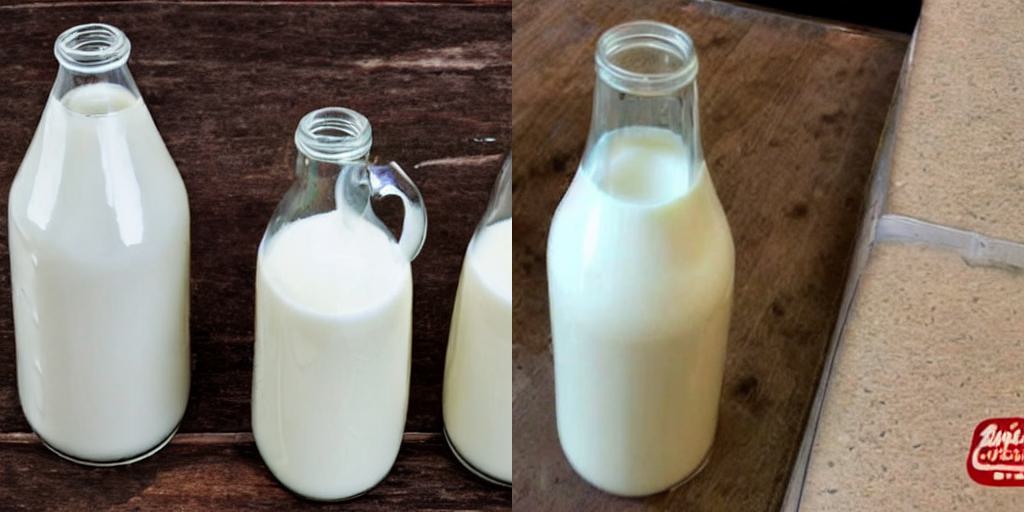}}
         \caption{\textit{Lot} of milk in the bottle}
         \label{fig:lotmilk}
     \end{subfigure}
     \centering
     \begin{subfigure}[b]{0.24\textwidth}
         \centering
         \fboxred{\includegraphics[width=\textwidth]{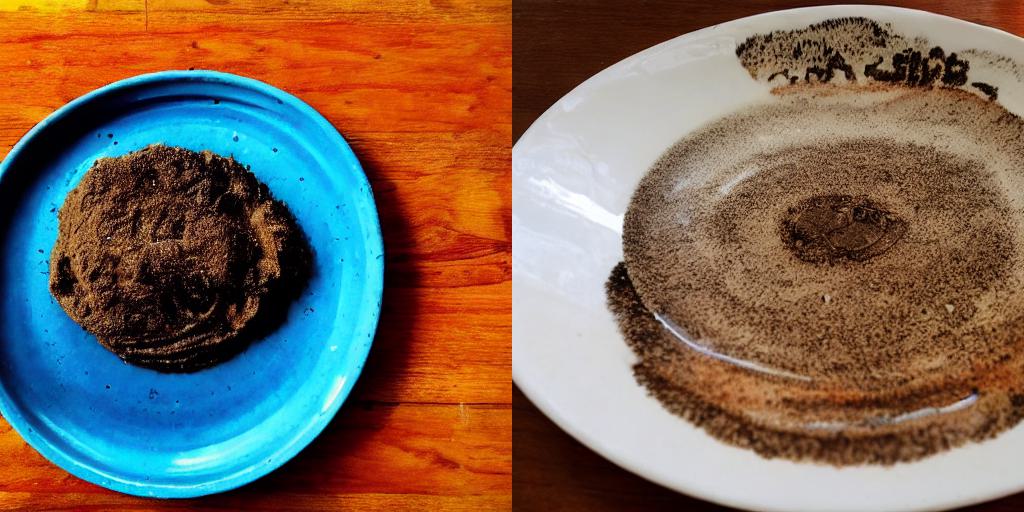}}
         \caption{The plate is a \textit{little} dirty}
         \label{fig:littledirty}
     \end{subfigure}
     \hfill
     \begin{subfigure}[b]{0.24\textwidth}
         \centering
         \fboxred{\includegraphics[width=\textwidth]{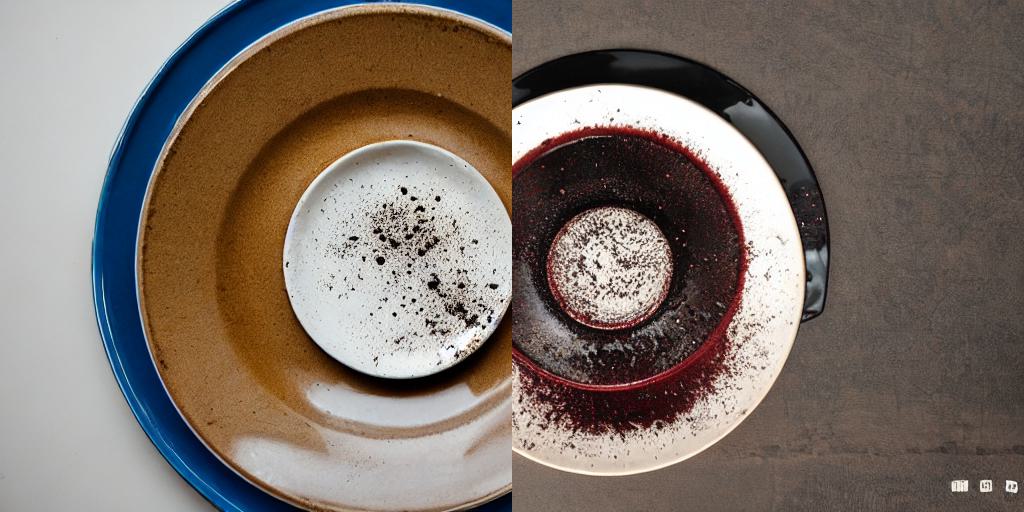}}
         \caption{The plate is \textit{very} dirty}
         \label{fig:verydirty}
     \end{subfigure}
     \hfill
     \begin{subfigure}[b]{0.24\textwidth}
         \centering
         \fboxred{\includegraphics[width=\textwidth]{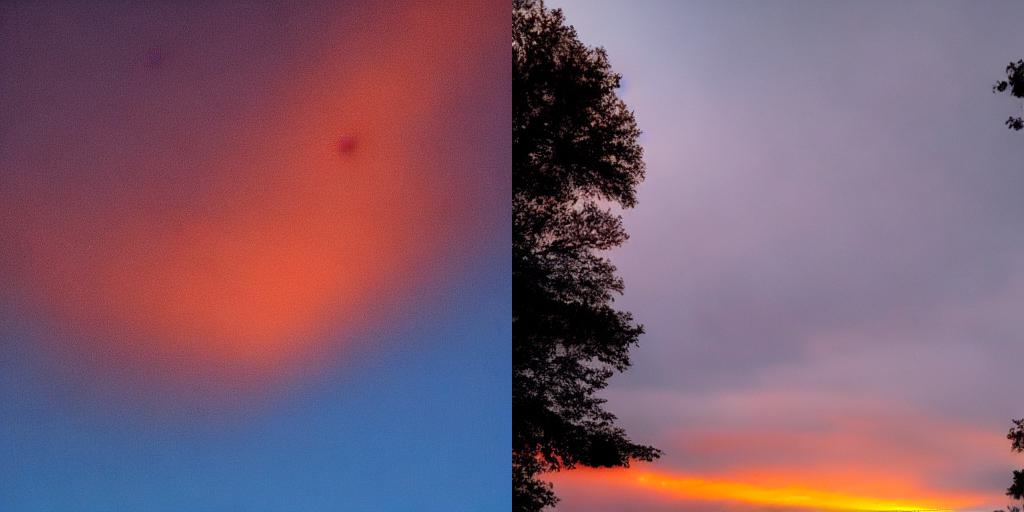}}
         \caption{The sky is \textit{not} orange}
         \label{fig:notorange}
     \end{subfigure}
     \hfill
     \begin{subfigure}[b]{0.24\textwidth}
         \centering
         \fboxgreen{\includegraphics[width=\textwidth]{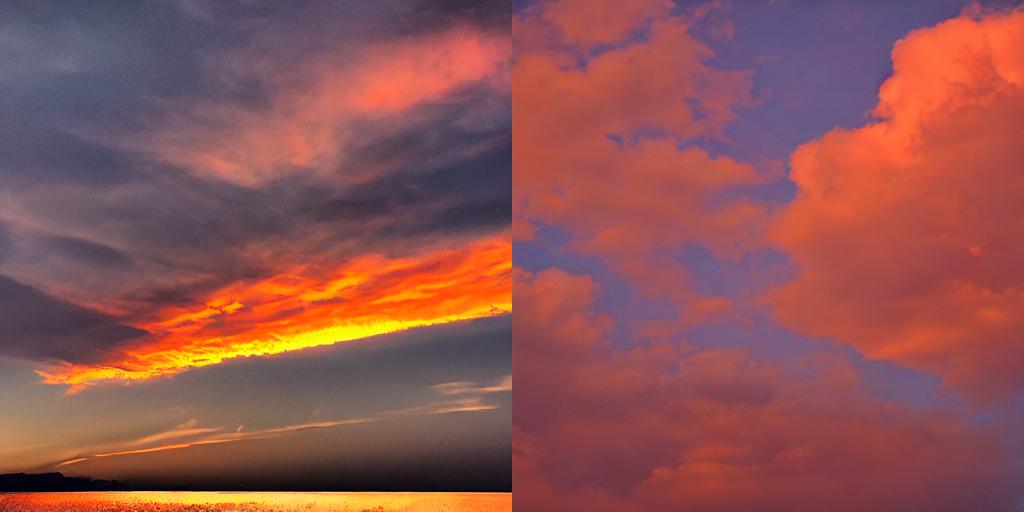}}
         \caption{The sky is \textit{always} orange}
         \label{fig:alwaysorange}
     \end{subfigure}
     \caption{Sample images depicting SDM's success (green border) and failure (red border) in capturing the semantics of different subcategories of {\bf determiners} and {\bf qualifiers}. Unlike the case of subject pronouns, images (a)--(c) show that SDMs cannot capture the notion of singularity implicit in articles like a, an, and the. They also exhibit weak understanding of cardinal numerals like (d) one, (e) two, and (f) ten. Concept of \textit{less} and \textit{more} suggested by quantifiers like (g) few, and (h) many for countable nouns and quantifiers like (i) little, and (j) lot for uncountable nouns is also not modeled by SDMs. (k)--(n) Qualifiers, which likewise cover the concept of \textit{less} and \textit{more} for adjectives and adverbs, too fail to be modeled by SDMs.}
     \label{fig:determiners}
     \vspace{-5mm}
\end{figure}

\subsection{Determiners and Qualifiers}
Determiners include articles, cardinal numerals, and quantifiers.
Articles like, a, an, and the modify the nouns by placing a restriction on them to show how particular or generic they are. 
Articles generally indicate a single \textit{unit} of noun that is being modified.
Cardinal numerals and quantifiers also modify the nouns by indicating their quantity.

Using nouns that have the same single and plural form, we examine whether the diffusion model captures the singularity implied in the articles. 
Prompts include `\textit{A} dice rolled on the table', `\textit{An} aircraft performing an air show', `The dog is guiding \textit{the} sheep'.
Images in figure \ref{fig:a}--\ref{fig:the} reveal that the diffusion models do not capture the singularity.

Cardinal numerals also offer a similar class of words to test the diffusion models. Simply using number words like one, two, and ten to fill the \textit{mask}  and looking at images generated for prompts `There is/are \textit{mask} orange/oranges in the photo', we observed that diffusion models do not understand the concept of numbers as can be seen in figure \ref{fig:one}--\ref{fig:ten}.

Next interesting category of determiners is quantifiers which includes words like little, few, lot, many etc. Using sentences like `There is \textit{little} milk in the bottle', `There is a \textit{lot of} milk in the bottle', `\textit{Few} oranges in the basket', and \textit{Many} bananas in the basket', we find that the diffusion models lack the comprehension for quantifiers as well as can be seen in figure \ref{fig:few}-- \ref{fig:lotmilk}.

Qualifiers are similar to quantifiers and numbers which also limit or enhance another word's meaning, but are associated with the adjectives and adverbs rather than the nouns. Examples include not, never, always, a little, very, etc. We designed few prompts like `The sky is \textit{not} orange', `The sky is \textit{never} orange', `The sky is \textit{always} orange', `The plate is a \textit{little} dirty', `The plate is \textit{very} dirty'. Unsurprisingly as with the case with quantifiers and numbers, figure \ref{fig:littledirty}--\ref{fig:alwaysorange} shows that the diffusion models fails to recognize the negation as well as cannot model the intensity of the qualifier .

\begin{figure}
     \centering
     \begin{subfigure}[b]{0.32\textwidth}
         \centering
         \fboxred{\includegraphics[width=\textwidth]{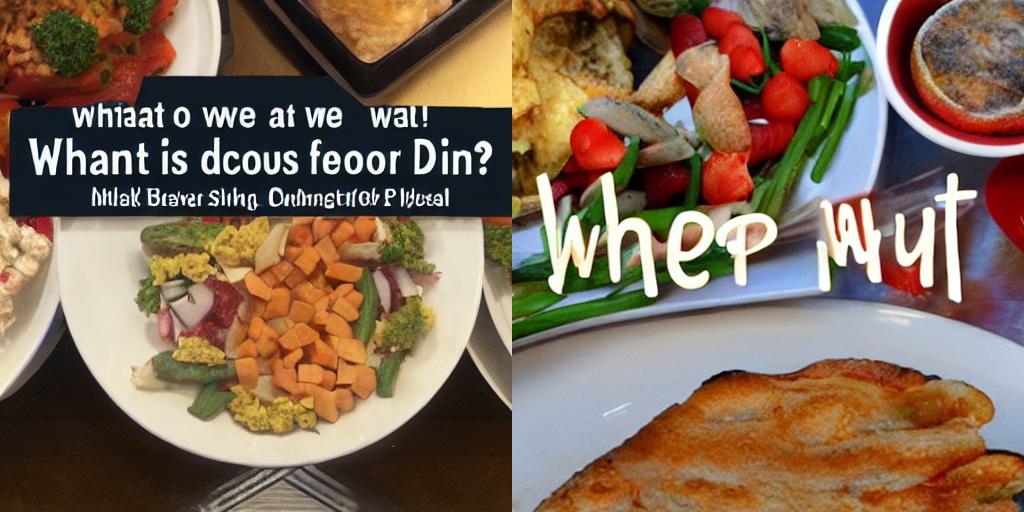}}
         \caption{\textit{What} are we eating for dinner}
         \label{fig:what}
     \end{subfigure}
     \hfill
     \begin{subfigure}[b]{0.32\textwidth}
         \centering
         \fboxred{\includegraphics[width=\textwidth]{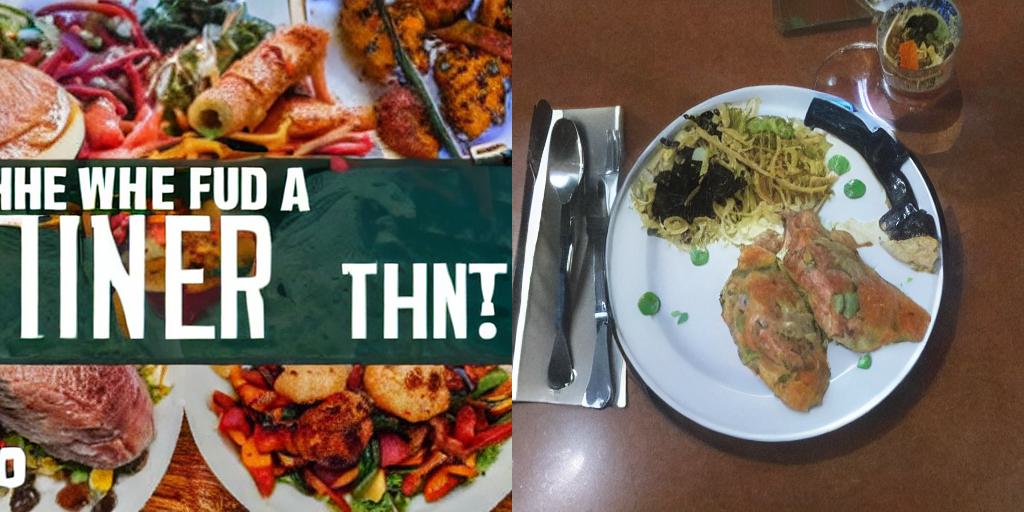}}
         \caption{\textit{Where} did we eat the dinner}
         \label{fig:where}
     \end{subfigure}
     \hfill
     \begin{subfigure}[b]{0.32\textwidth}
         \centering
         \fboxred{\includegraphics[width=\textwidth]{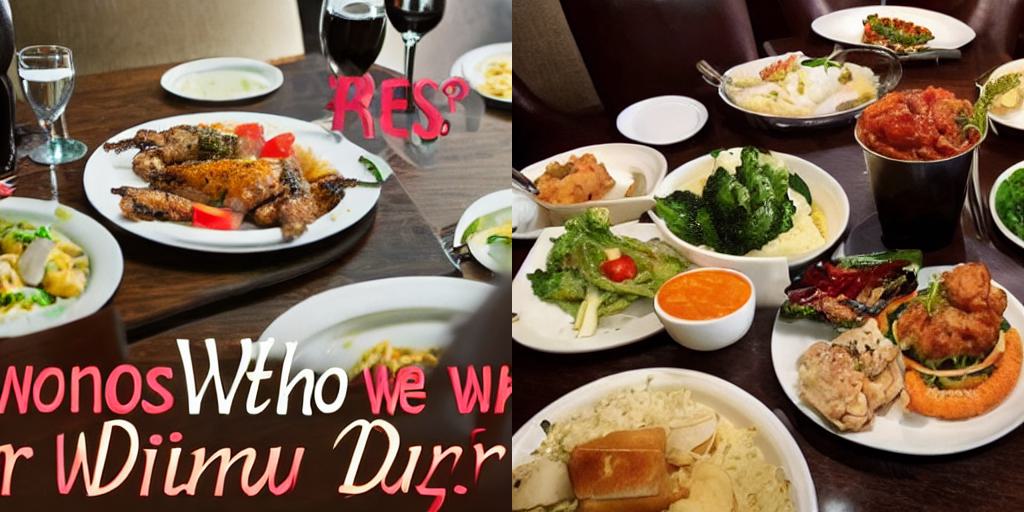}}
         \caption{\textit{Who} are we eating with for dinner}
         \label{fig:who}
     \end{subfigure}
    \centering
     \begin{subfigure}[b]{0.32\textwidth}
         \centering
         \fboxgreen{\includegraphics[width=\textwidth]{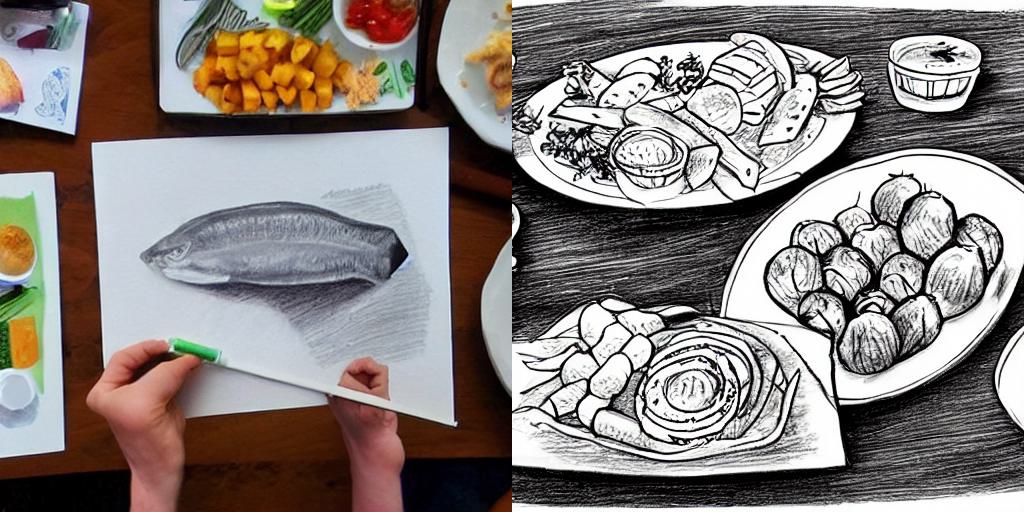}}
         \caption{Draw a picture of \textit{what} we had for the dinner}
         \label{fig:dwhat}
     \end{subfigure}
     \hfill
     \begin{subfigure}[b]{0.32\textwidth}
         \centering
         \fboxgreen{\includegraphics[width=\textwidth]{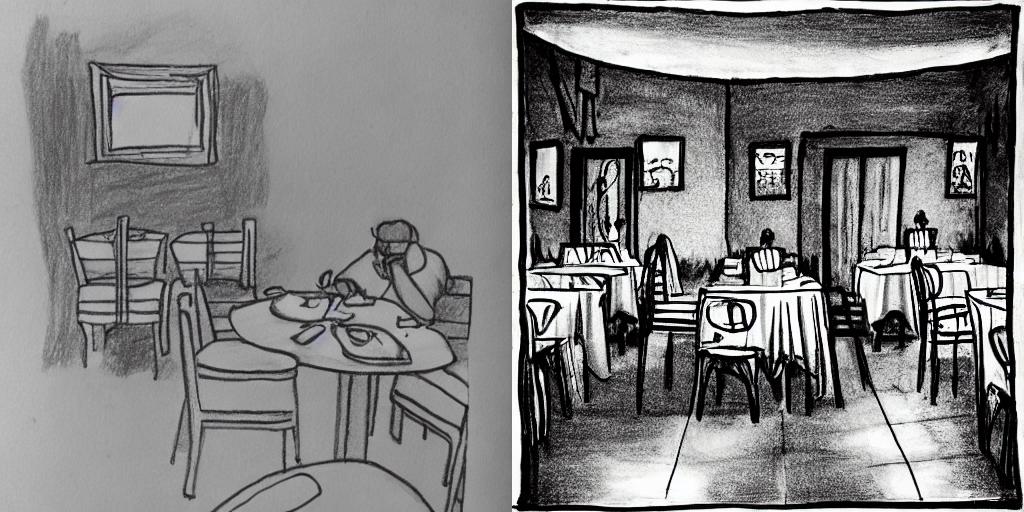}}
         \caption{Draw a picture of \textit{where} we had the dinner}
         \label{fig:dwhere}
     \end{subfigure}
     \hfill
     \begin{subfigure}[b]{0.32\textwidth}
         \centering
         \fboxgreen{\includegraphics[width=\textwidth]{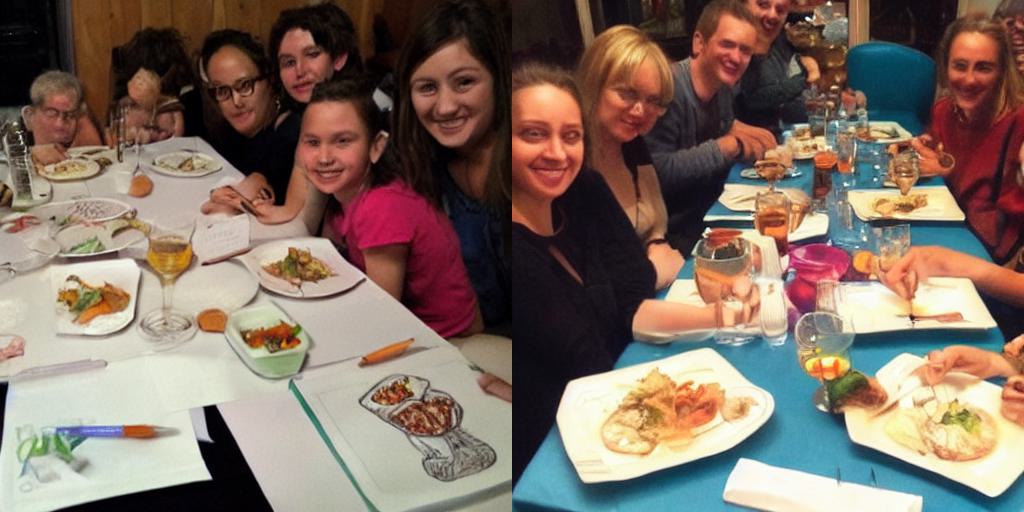}}
         \caption{Draw a picture of \textit{who} we had the dinner with}
         \label{fig:dwho}
     \end{subfigure}
     \caption{Sample images depicting SDM's success (green border) and failure (red border) in capturing the semantics of {\bf interrogatives} and {\bf relatives}. SDMs are unable to understand that the answers to questions like (a) what, (b) where, and (c) who are respectively, things, locations, and persons. However, when (d) what, (e) where, and (f) who are used as relatives, SDMs show that MLMs can capture their essence. This category of function words is the simplest of all function words since it comprises words that co-occur with different contexts in texts, making it understandable even without multimodal learning.}
     \label{fig:interrogatives}
\end{figure}
\begin{figure}
     \centering
     \begin{subfigure}[b]{0.45\textwidth}
         \centering
         \fboxred{\includegraphics[width=\textwidth]{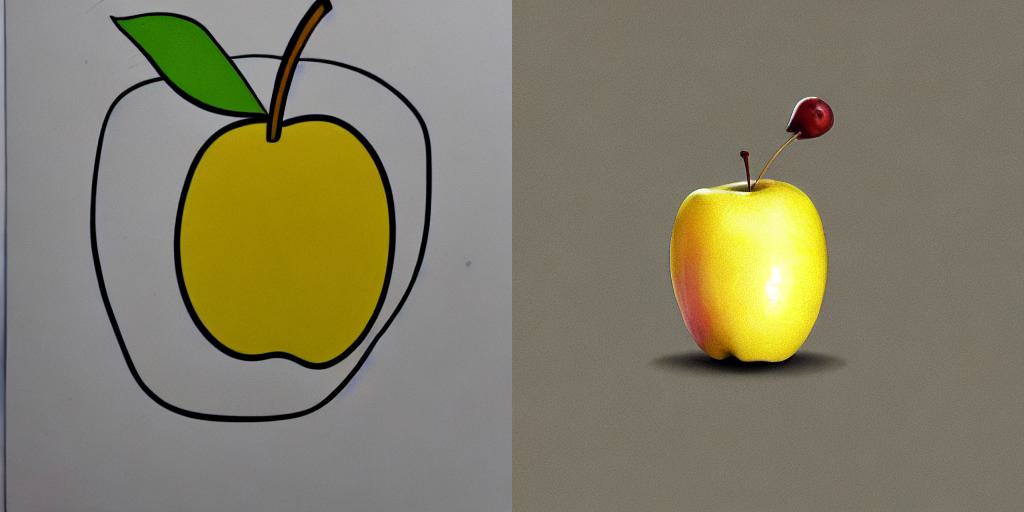}}
         \caption{Draw an apple \textit{and} a banana}
         \label{fig:either}
     \end{subfigure}
     \hfill
     \begin{subfigure}[b]{0.45\textwidth}
         \centering
         \fboxgreen{\includegraphics[width=\textwidth]{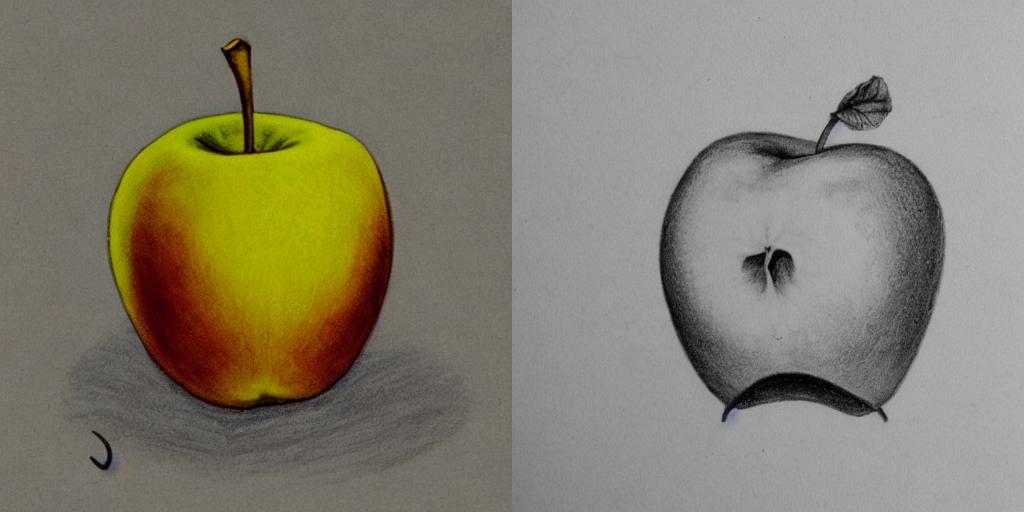}}
         \caption{Draw \textit{either} an apple \textit{or} a banana}
         \label{fig:or}
     \end{subfigure}
     \caption{Sample images depicting SDM's success (green border) and failure (red border) in capturing the semantics of {\bf conjunctions}. This category offers an intriguing scenario because it explores the area of logic and reasoning. Conjunctions \textit{either/ or} and \textit{and} express a decision choice, however the images (a) and (b) show that SDM is unable to comprehend the choice implied by the conjunctions.}
     \label{fig:conjunctions}
     \vspace{-5mm}
\end{figure}

\subsection{Interrogatives and Relatives}
What, Where, and Who are examples of interrogative words used to pose a query. The reason we use these words to probe the diffusion model lies in the answer to these queries. The answer for what, where, and who are objects, places, or person respectively. `Wh-' words can also act as relatives in descriptive sentences where they do not pose a question.

Thus, to probe the diffusion model with `wh-' words  we use both questions as well as descriptive prompts like `\textit{What} are we eating for dinner', `\textit{Where} did we eat the dinner?', `\textit{Who} are we eating dinner with?', `Draw a picture of \textit{what} we had for the dinner' `Draw a picture of \textit{where} we had the dinner.', and `Draw a picture of \textit{who} we had the dinner with'. We conclude from the images that the diffusion model does not understand the semantics of the answers to the interrogative (figure \ref{fig:what}--\ref{fig:who}). But surprisingly when used in descriptive sentence, it was able to comprehend the answers (figure \ref{fig:dwhat}--\ref{fig:dwho}).

\subsection{Conjunctions}
Conjunctions play an important role in english grammar by connecting two sentences. Examples include and, but, yet, either, or etc. We use \textit{or} and \textit{and} since they provide a choice between two alternatives. With language prompts like `Draw an apple and a banana' and `Draw either an apple or a banana', we can probe the diffusion model to understand if it can understand the notion of a choice implicit in \textit{or} vs \textit{and}. We find that the model generates just an apple both the cases (figure \ref{fig:either} and \ref{fig:or}).


\section{Conclusions}
We have explored the limitations of learned representations of function words in multimodal language models by a visual tour of images generated by stable diffusion models.
Our results indicate that the semantics of function words are poorly understood by these language models.
In particular, stable diffusion models only work for select pronoun subcategories and the category of relatives out of the seven categories of function words and their multiple subcategories.
Future work by the research community should focus on methods to remedy these shortcomings, such as the construction of function word datasets.

\bibliographystyle{unsrt}  
\bibliography{references}

\newpage
\appendix
\section{Appendix}
In the appendix, we layout creative prompts that could demonstrate the characteristics of each category of the function words. For each prompt, we provide four additional images generated by the stable diffusion model.

\subsection{Pronouns}
Pronouns contains five categories: subject pronouns (SP), object pronouns (OP),  possessive adjectives (PA), indefinite pronouns (IP), reflexive pronouns (RP). Table~\ref{fig:app_pronouns_1} and Table~\ref{fig:app_pronouns_2} shows additional prompts for each category and their corresponding generated images. 

\begin{table}[!ht]
\centering
    \begin{tabular}[t]{@{} c m{0.47\linewidth} @{}}
    \textbf{Prompts} & \centering\arraybackslash\textbf{Images} \\ 
    \midrule \addlinespace
    {\color{red} \textbf{He}} is dancing in the rain. \textbf{(SP)} & \includegraphics[width=\linewidth]{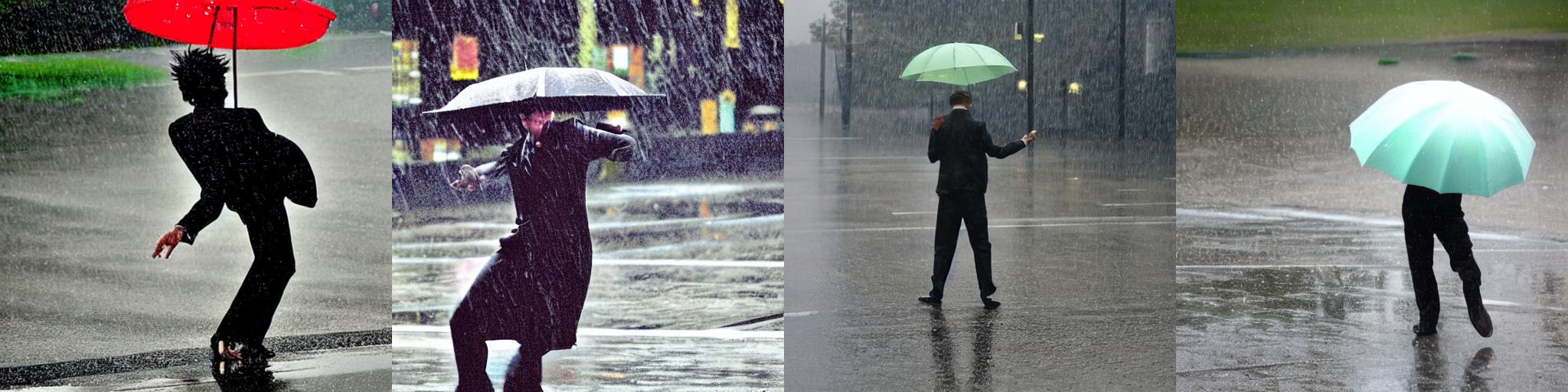} 
    \\ \addlinespace
    {\color{red}\textbf{She}} is dancing in the rain. \textbf{(SP)}& \includegraphics[width=\linewidth]{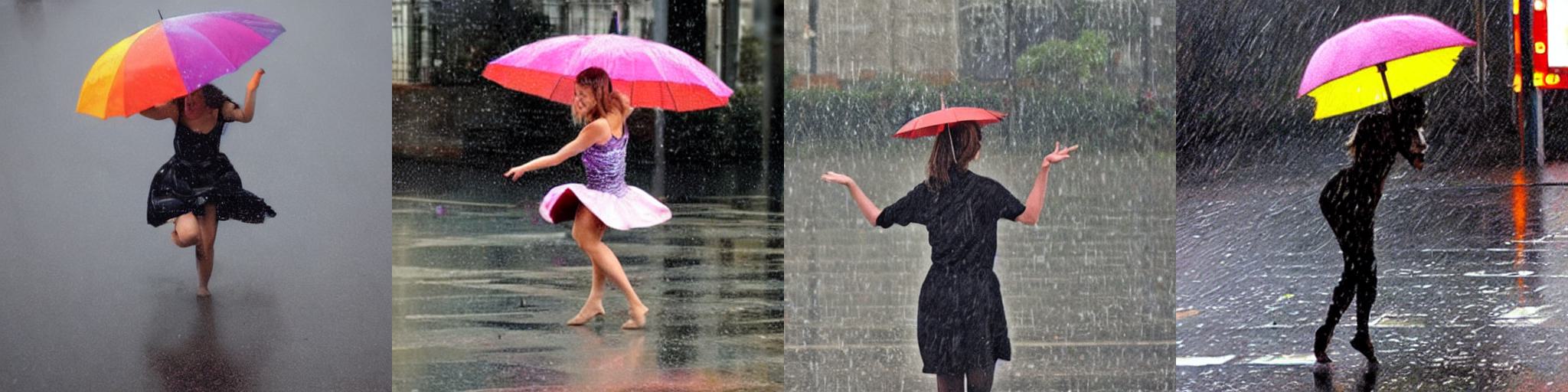} 
    \\ \addlinespace
    {\color{red}\textbf{We}} are dancing in the rain. \textbf{(SP)}& \includegraphics[width=\linewidth]{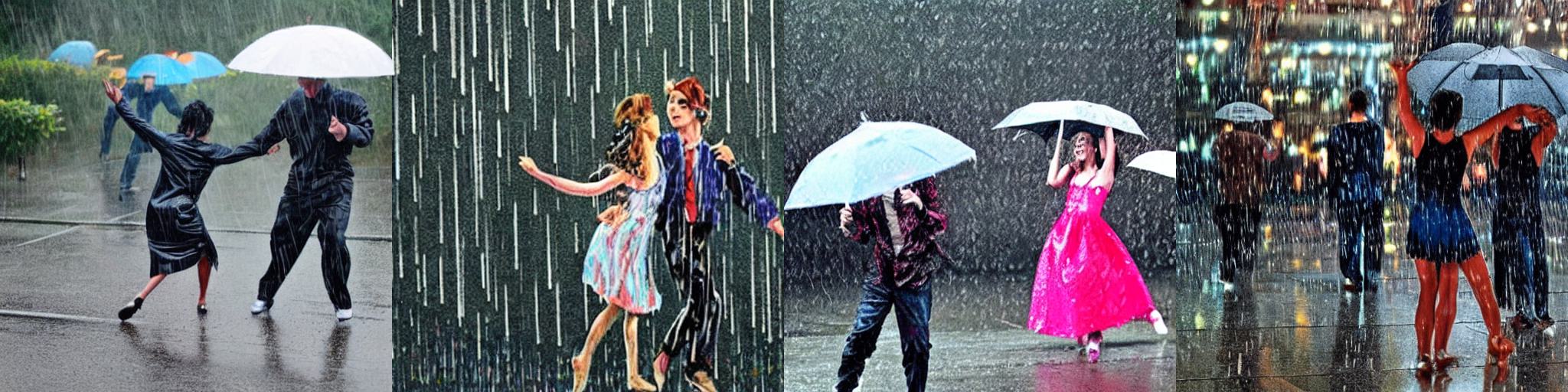} 
    \\  \addlinespace
    \midrule
    \addlinespace
    The teacher hugs {\color{red} \textbf{him}}.  \textbf{(OP)} & {\includegraphics[width=\linewidth]{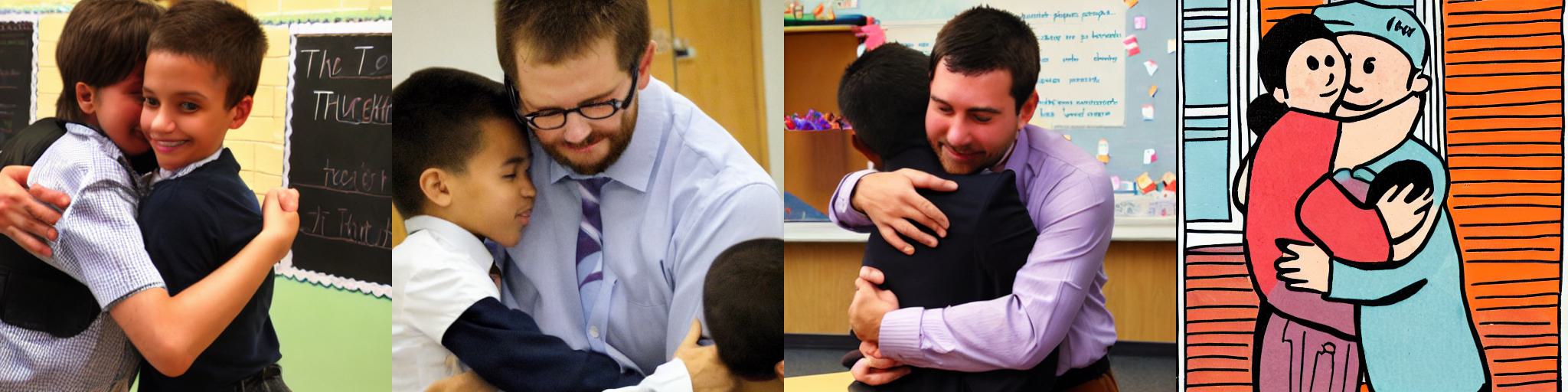}}
    \\ \addlinespace
     The teacher hugs {\color{red} \textbf{her}}. \textbf{(OP)} & \includegraphics[width=\linewidth]{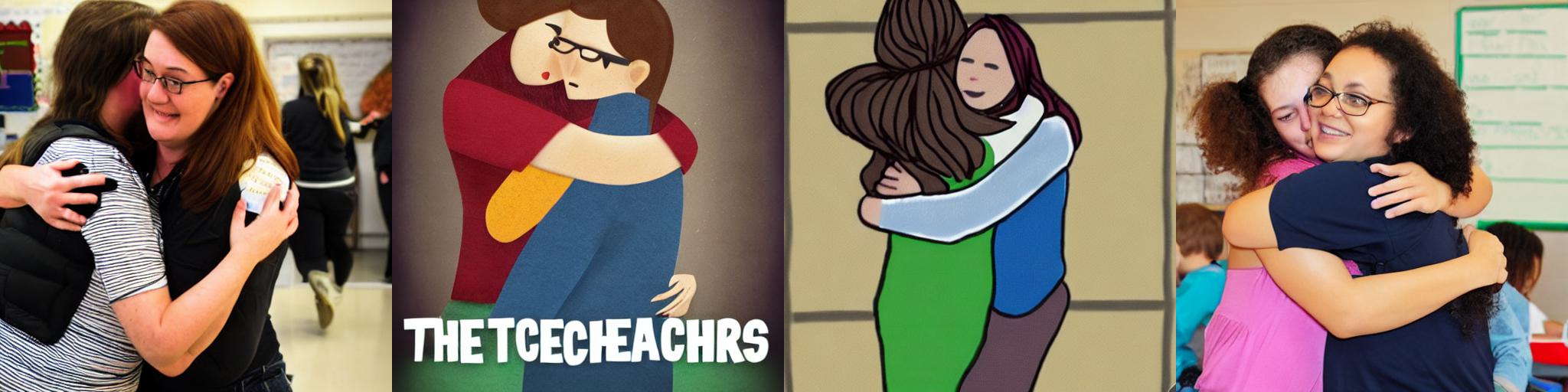}   
    \\  \addlinespace
    \midrule
    \addlinespace
    The boy is holding {\color{red} \textbf{his}} hand. \textbf{(PA)} & \includegraphics[width=\linewidth]{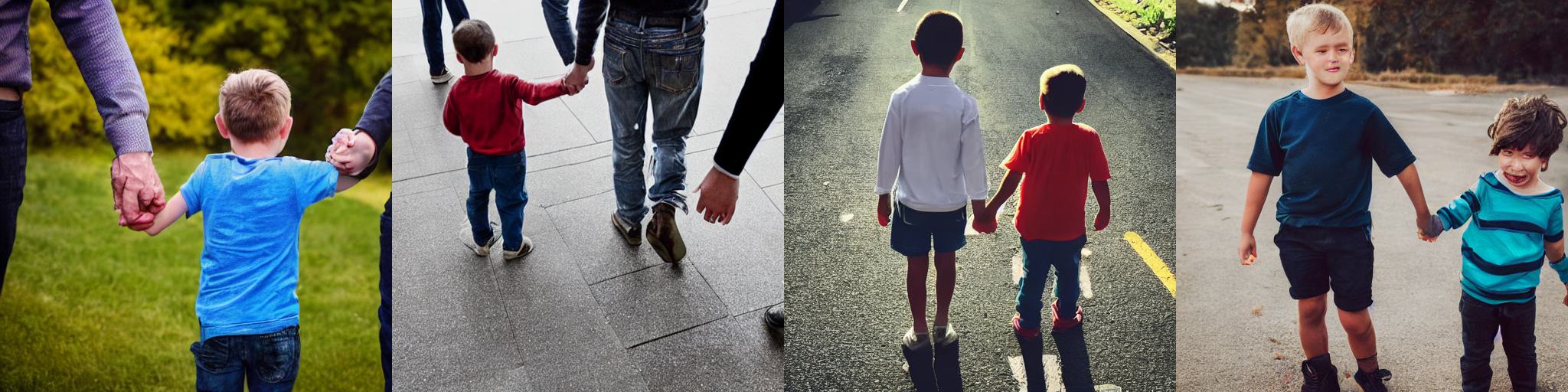} 
    \\ \addlinespace
    The boy is holding {\color{red} \textbf{her}} hand. \textbf{(PA)} & \includegraphics[width=\linewidth]{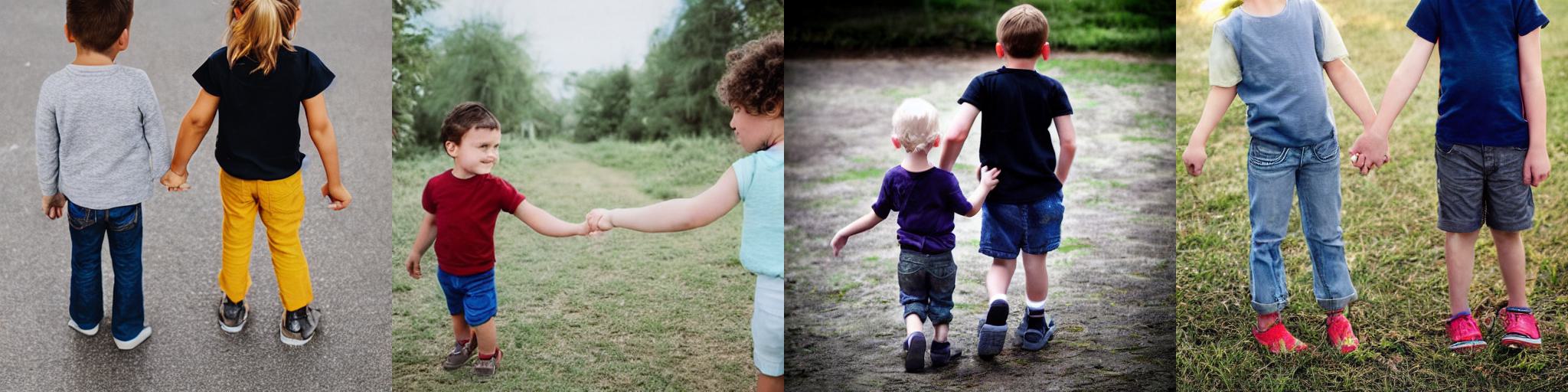}   
    \\ \bottomrule 
    \end{tabular}
    \vspace{4mm}
    \caption{Images generated by Stable Diffusion Model for prompts with pronouns. This table covers first three subcategories of pronouns: subject pronouns (SP), object pronouns (OP), possessive adjectives (PA). The pronoun words are colored in red. }
    \label{fig:app_pronouns_1}
\end{table}

\begin{table}[H]
\centering
    \begin{tabular}[t]{@{} c m{0.47\linewidth} @{}}
    \textbf{Prompts} & \centering\arraybackslash\textbf{Images} \\ 
    \midrule \addlinespace
    {\color{red} \textbf{Nobody}} in the group is wearing a hat. \textbf{(IP)} & \includegraphics[width=\linewidth]{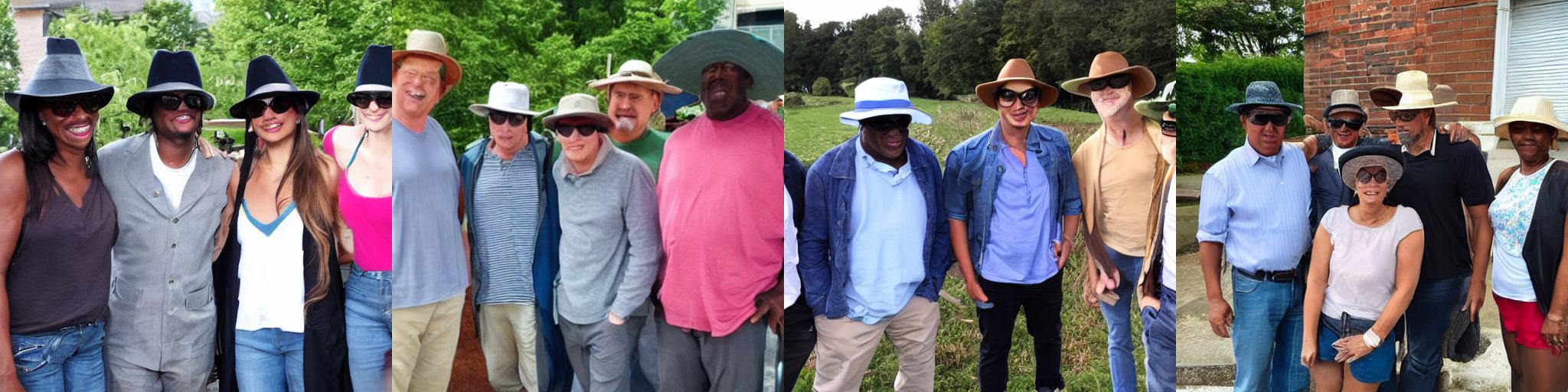} 
    \\ \addlinespace
    {\color{red} \textbf{No one}} in the group is wearing a hat. \textbf{(IP)} & \includegraphics[width=\linewidth]{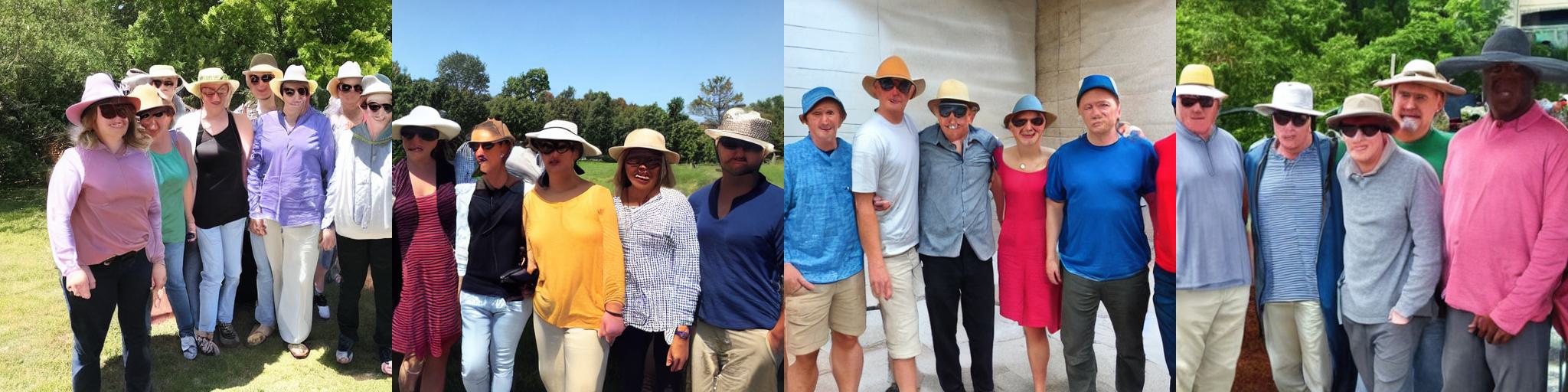}   
    \\ \addlinespace
    {\color{red} \textbf{Some}} in the group are wearing hats. \textbf{(IP)} & \includegraphics[width=\linewidth]{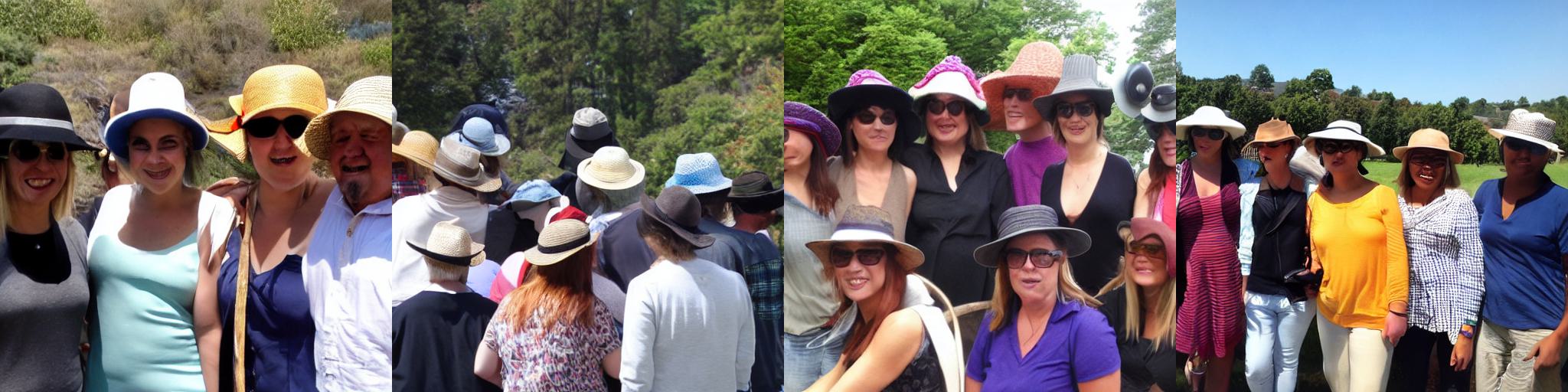}  
    \\ \addlinespace
    {\color{red} \textbf{Someone}} in the group is wearing a hat. \textbf{(IP)} & \includegraphics[width=\linewidth]{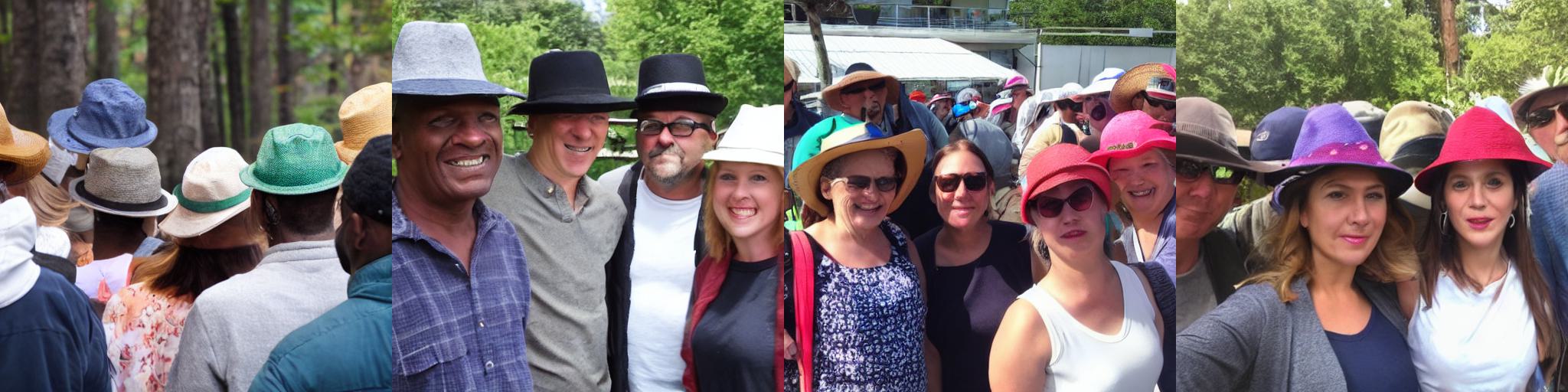}  
    \\  \addlinespace
    {\color{red} \textbf{Everyone}} in the group is wearing a hat. \textbf{(IP)} & \includegraphics[width=\linewidth]{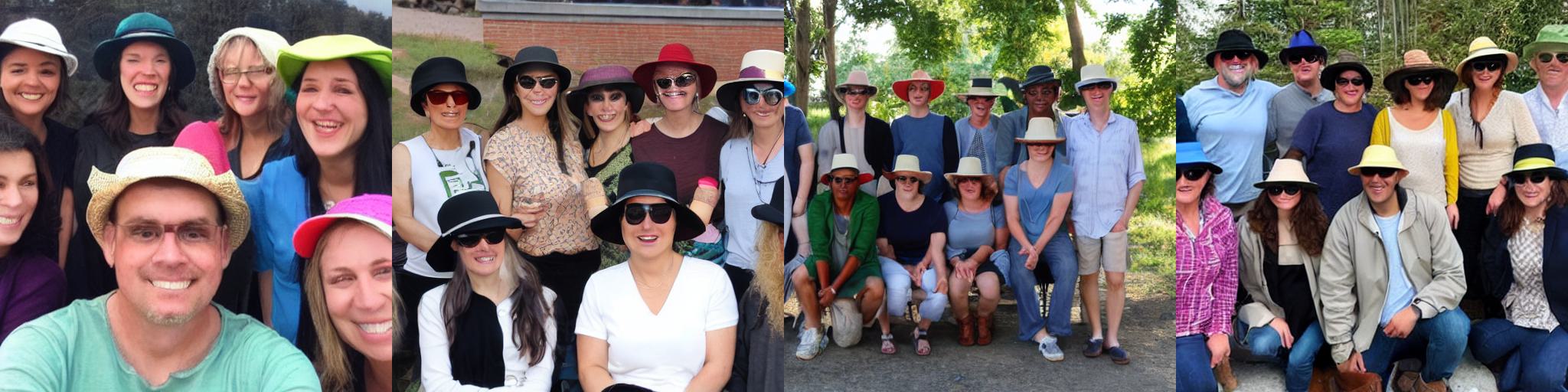}  
    \\  \addlinespace
    {\color{red} \textbf{Everybody}} in the group is wearing a hat. \textbf{(IP)} & \includegraphics[width=\linewidth]{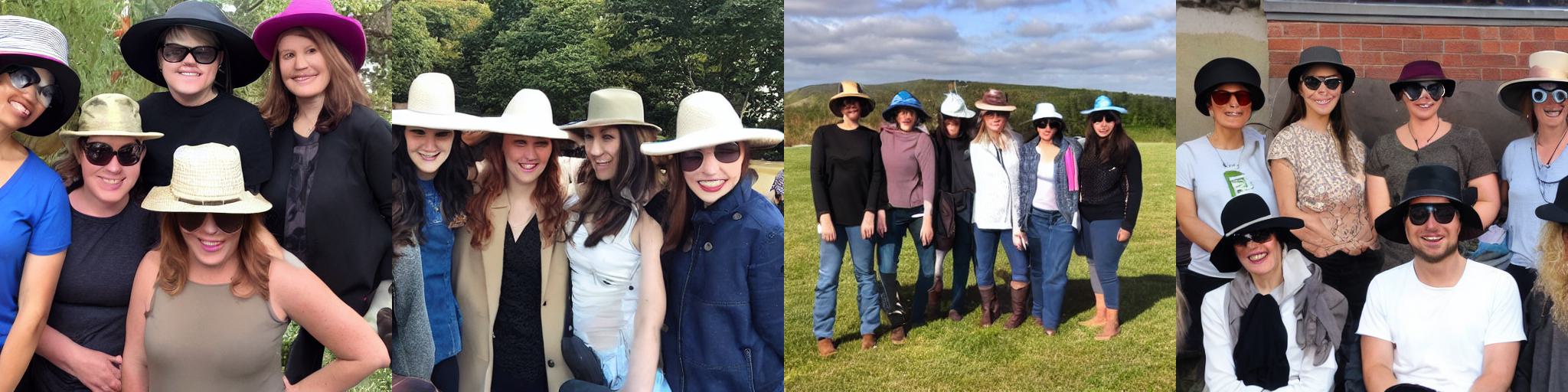}  
    \\  \addlinespace
    \midrule
    \addlinespace
    The boy punched {\color{red} \textbf{himself}} in the face. \textbf{(RP)} & \includegraphics[width=\linewidth]{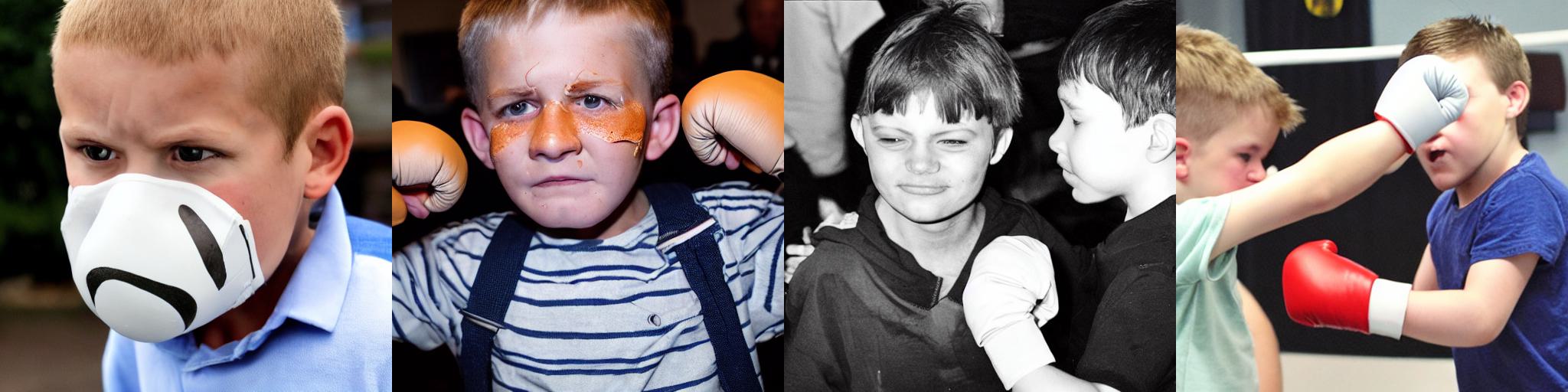}  
    \\ \addlinespace
    She patted {\color{red} \textbf{herself}}  for a job well done. \textbf{(RP)} & \includegraphics[width=\linewidth]{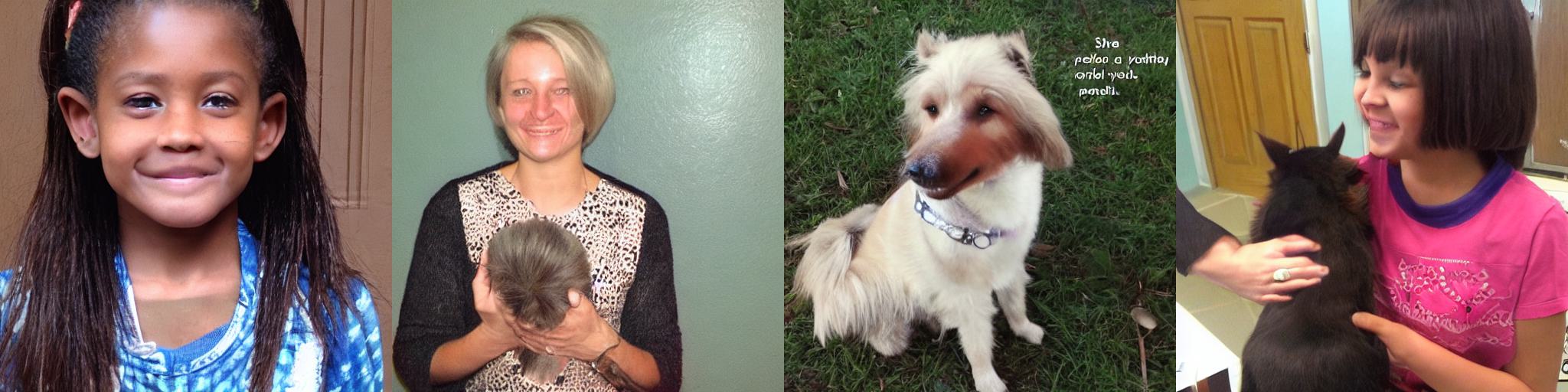}
    \\ \addlinespace
    I shook hands with {\color{red} \textbf{myself}}. \textbf{(RP)} & \includegraphics[width=\linewidth]{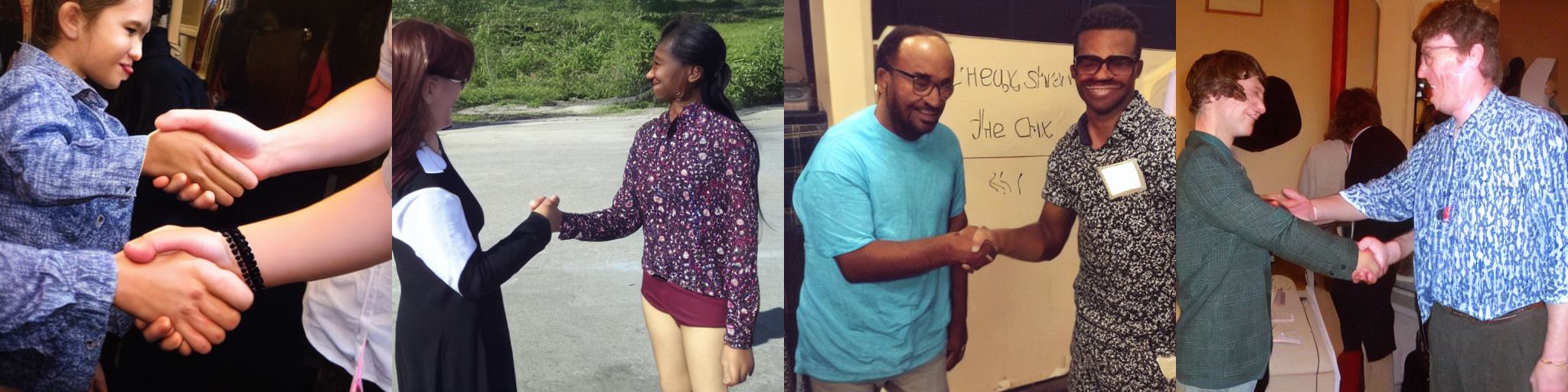}
    \\ \bottomrule 
    \end{tabular}
    \vspace{4mm}
    \caption{Images generated by Stable Diffusion Model for prompts with pronouns. This table covers last two subcategories of pronouns: indefinite pronouns (IP), reflexive pronouns (RP). The pronoun words are colored in red. }
    \label{fig:app_pronouns_2}
\end{table}

\subsection{Prepositions and Particles}
Prepositions include two categories, prepositions of place (PoP), which models entity positions and prepositions of movement (PoM) which signify temporal relations. We also include a special type of prepositions -- particle (Par), which shows the state of an object. Table~\ref{fig:app_prepositions_1} and Table~\ref{fig:app_prepositions_2} shows prompts and generated images for propositions.

\begin{table}[!ht]
\centering
    \begin{tabular}[t]{@{} c m{0.52\linewidth} @{}}
    \textbf{Prompts} & \centering\arraybackslash\textbf{Images} \\ 
    \midrule \addlinespace
    The ball is {\color{red} \textbf{in}} the box. \textbf{(PoP)} & 
    \includegraphics[width=\linewidth]{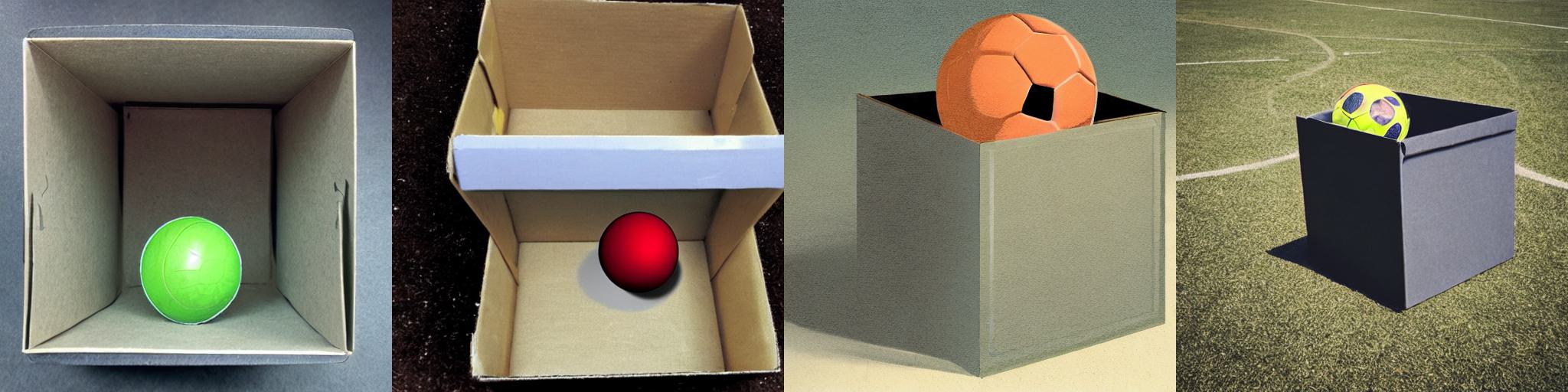}
    \\ \addlinespace
    The ball is {\color{red} \textbf{on}} the box. \textbf{(PoP)} & \includegraphics[width=\linewidth]{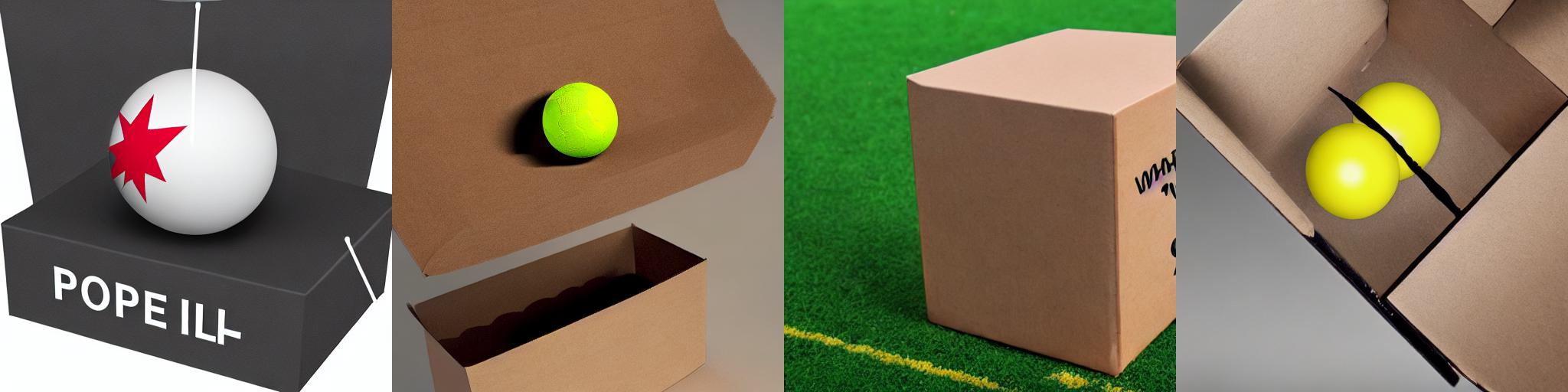}   
    \\ \addlinespace
   The ball is {\color{red} \textbf{under}} the box. \textbf{(PoP)} & \includegraphics[width=\linewidth]{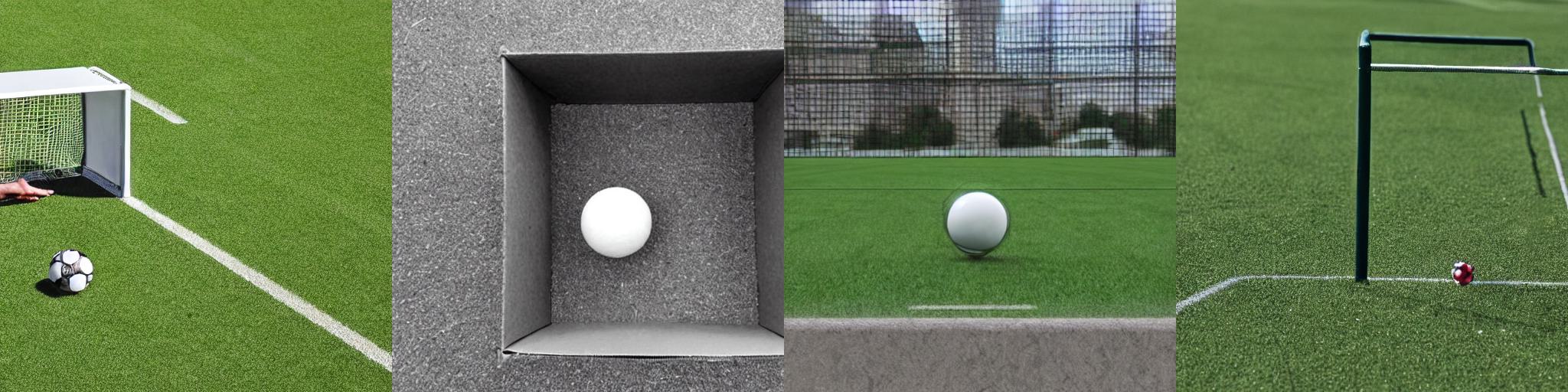}  
    \\ \addlinespace
    The ball is {\color{red} \textbf{next to}} the box. \textbf{(PoP)} & \includegraphics[width=\linewidth]{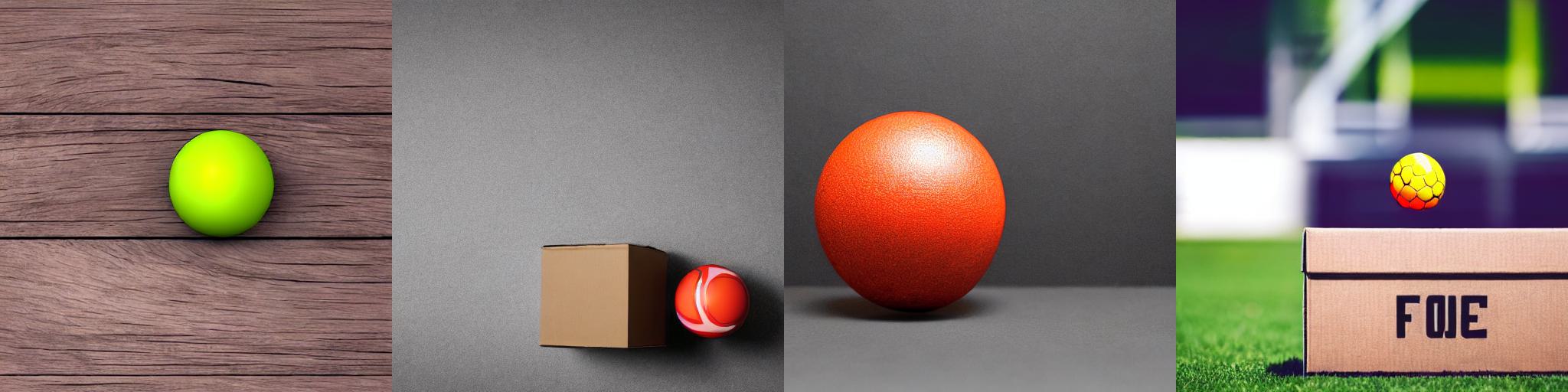}  
    \\  \addlinespace
    The ball is {\color{red} \textbf{behind}} the box. \textbf{(PoP)} & \includegraphics[width=\linewidth]{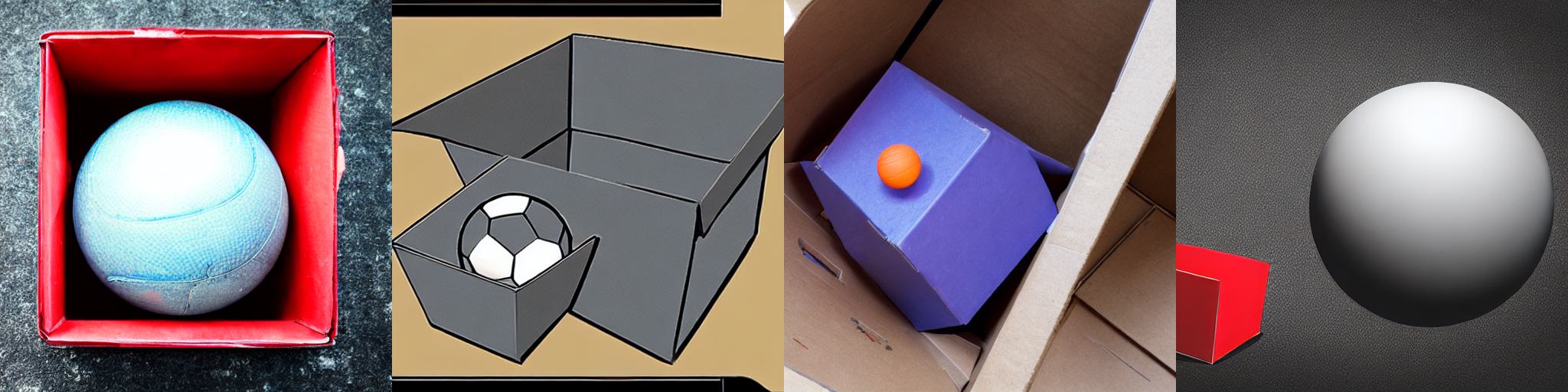}  
    \\  \addlinespace
    The ball is {\color{red} \textbf{in front of}} the box. \textbf{(PoP)} & \includegraphics[width=\linewidth]{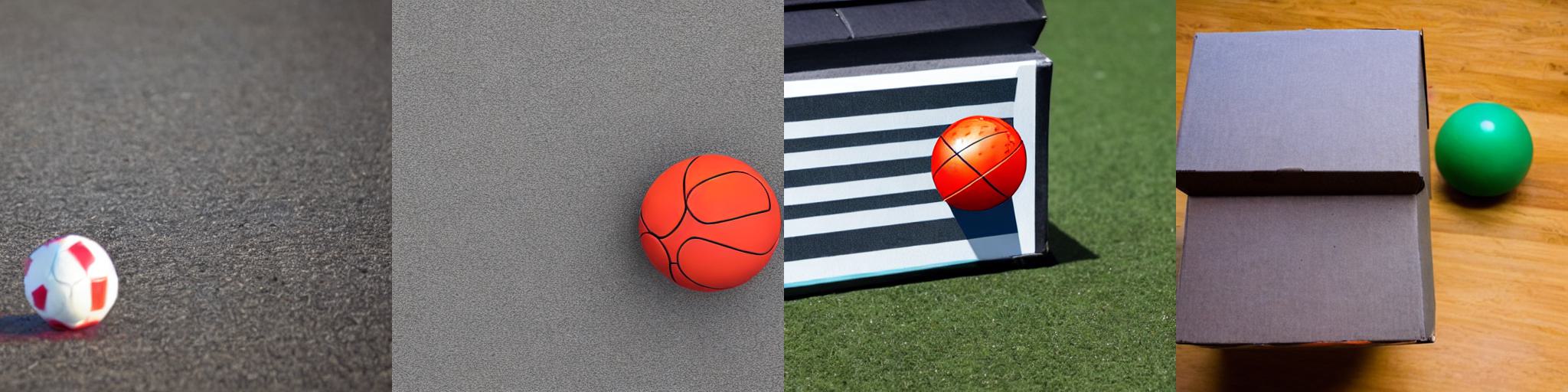}  
    \\  \addlinespace
    The ball is {\color{red} \textbf{in between}} the boxes. \textbf{(PoP)} & \includegraphics[width=\linewidth]{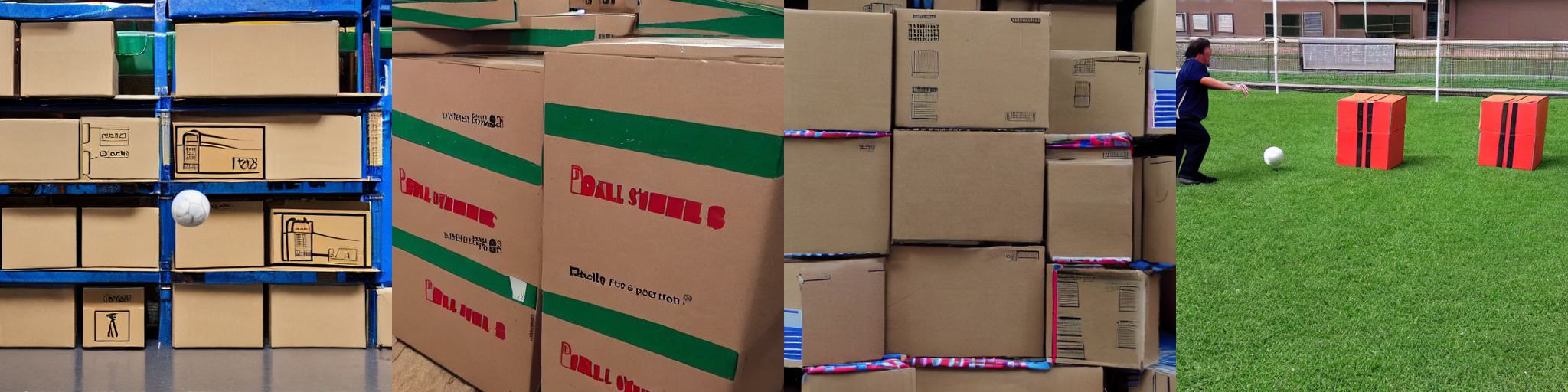}
    \\ \bottomrule 
    \end{tabular}
    \vspace{4mm}
    \caption{Images generated by Stable Diffusion Model for prompts with prepositions. This table covers the first subcategory of prepositions: preposition of place (PoP). The preposition words are colored in red. }
    \label{fig:app_prepositions_1}
\end{table}

\begin{table}[H]
\centering
    \begin{tabular}[t]{@{} c m{0.44\linewidth} @{}}
    \textbf{Prompts} & \centering\arraybackslash\textbf{Images} \\ 
    \midrule \addlinespace
    He is walking {\color{red} \textbf{up}} the stairs. \textbf{(PoE)} & \includegraphics[width=\linewidth]{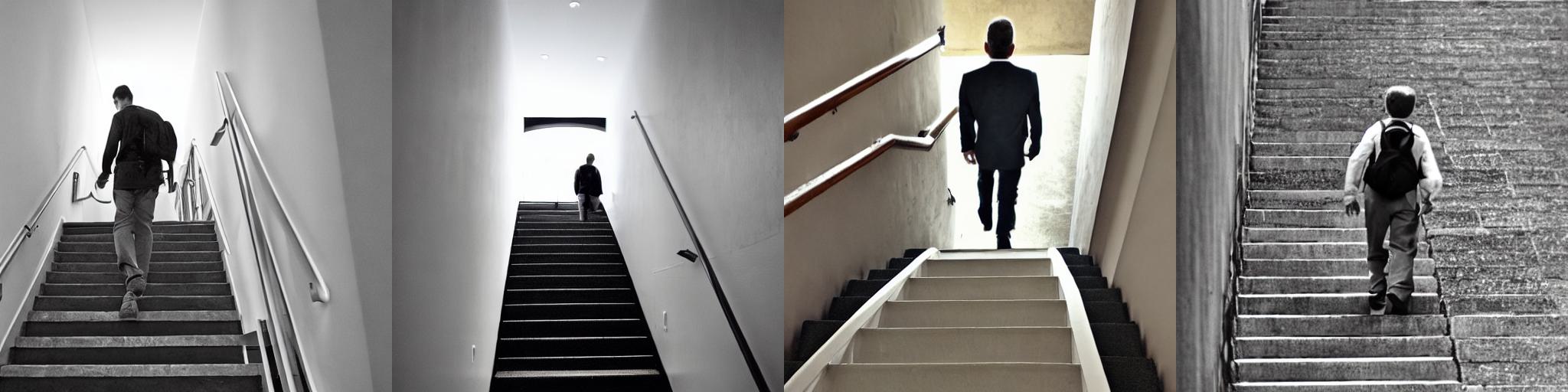} 
    \\ \addlinespace
    He is walking {\color{red} \textbf{down}} the stairs. \textbf{(PoE)} & \includegraphics[width=\linewidth]{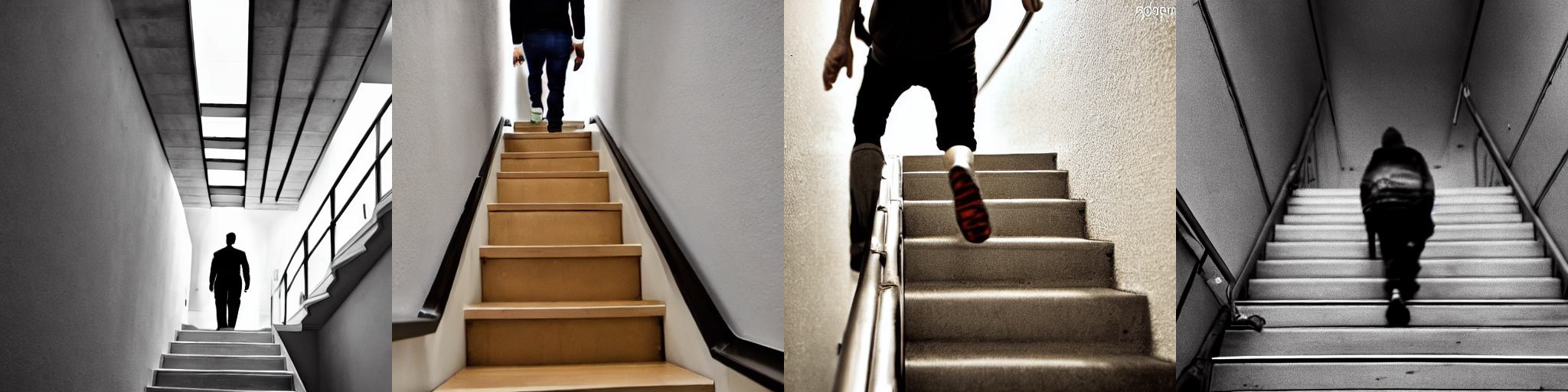}   
    \\ \addlinespace
    The girl ran {\color{red} \textbf{towards}} the dog. \textbf{(PoE)} & \includegraphics[width=\linewidth]{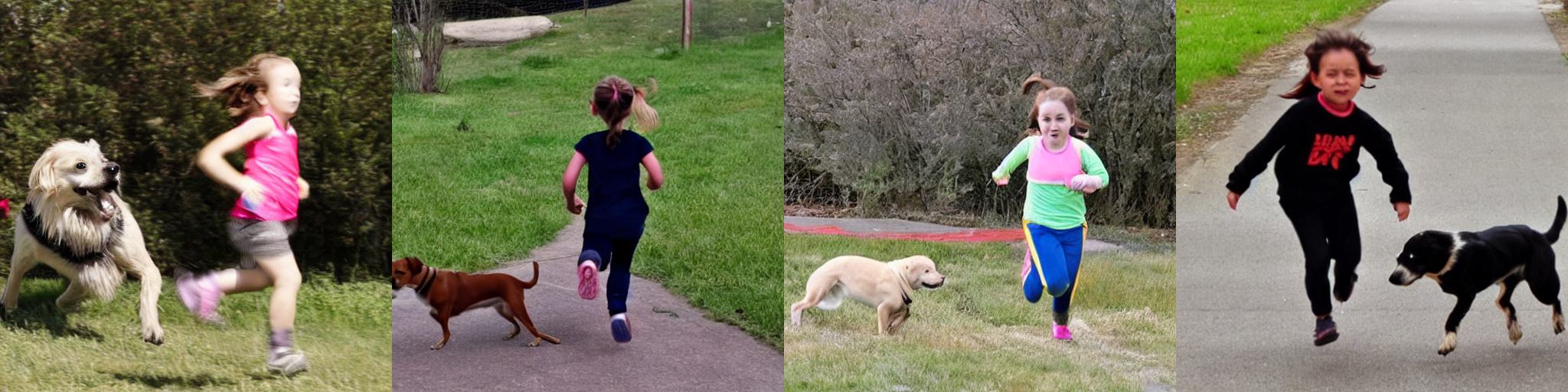}  
    \\ \addlinespace
    The girl ran {\color{red} \textbf{from}} the dog. \textbf{(PoE)} & \includegraphics[width=\linewidth]{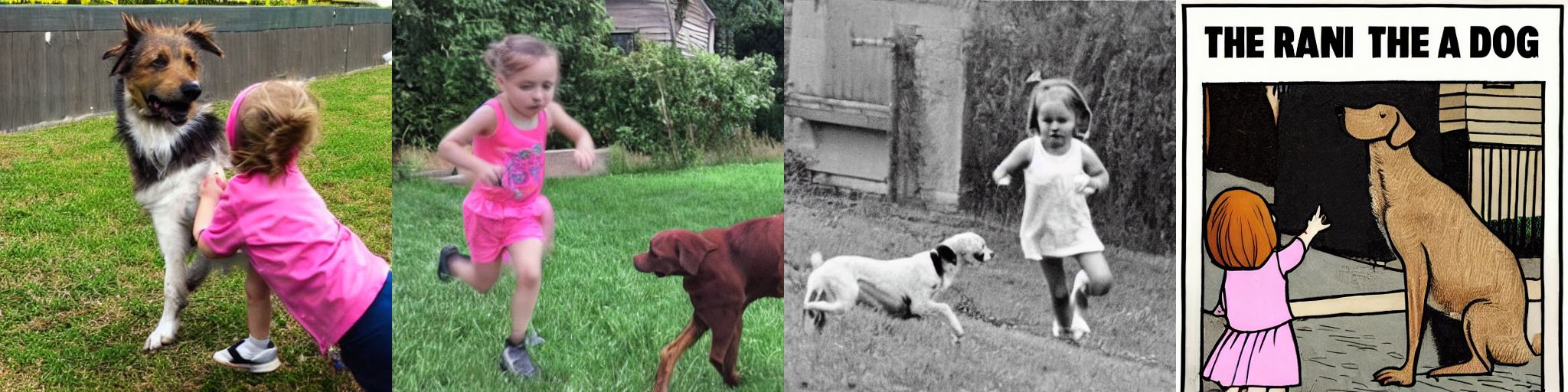}  
    \\  \addlinespace
    \midrule
    \addlinespace
    The light is {\color{red} \textbf{on}}. \textbf{(Par)} & \includegraphics[width=\linewidth]{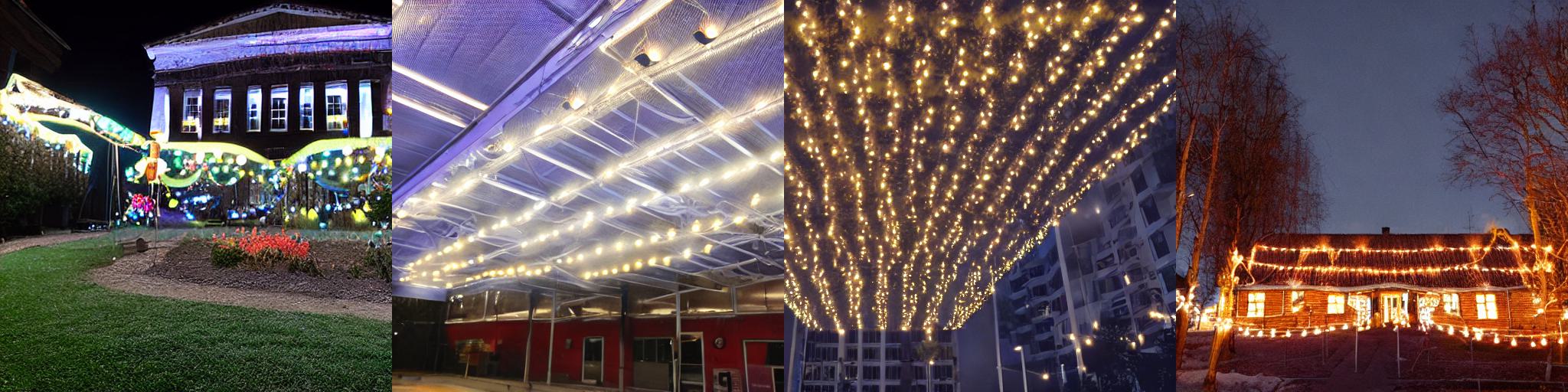}  
    \\  \addlinespace
    The light is {\color{red} \textbf{off}}. \textbf{(Par)} & \includegraphics[width=\linewidth]{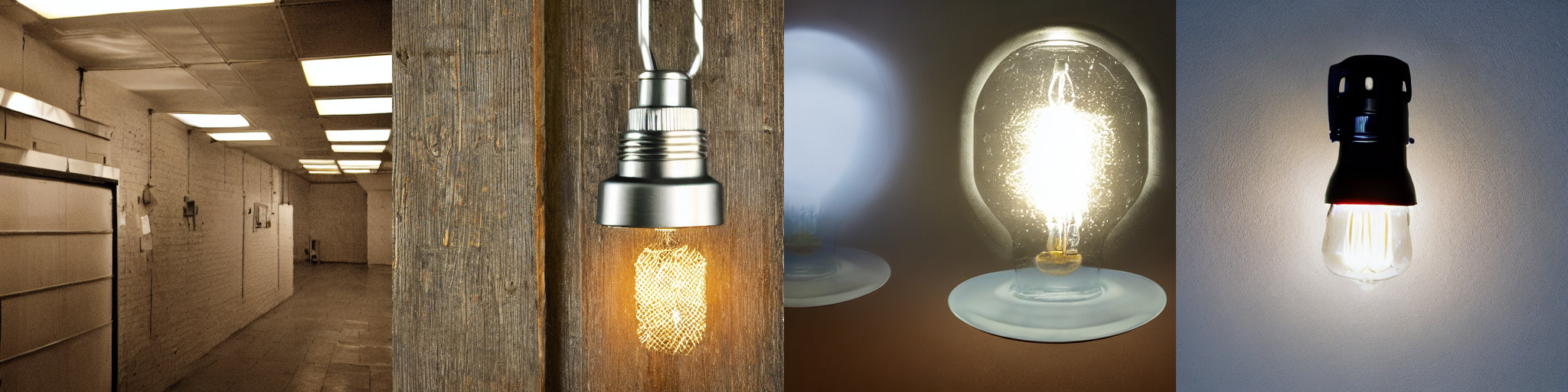}
    \\  \addlinespace
    Oats  {\color{red} \textbf{with}} milk. \textbf{(Par)} & \includegraphics[width=\linewidth]{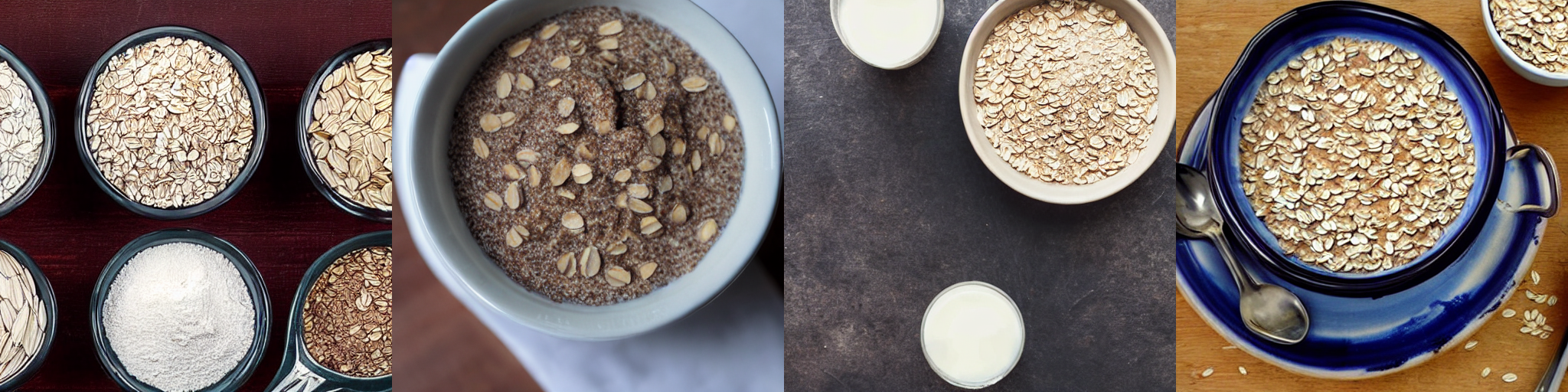}
    \\  \addlinespace
    Oats {\color{red} \textbf{without}} milk. \textbf{(Par)} & \includegraphics[width=\linewidth]{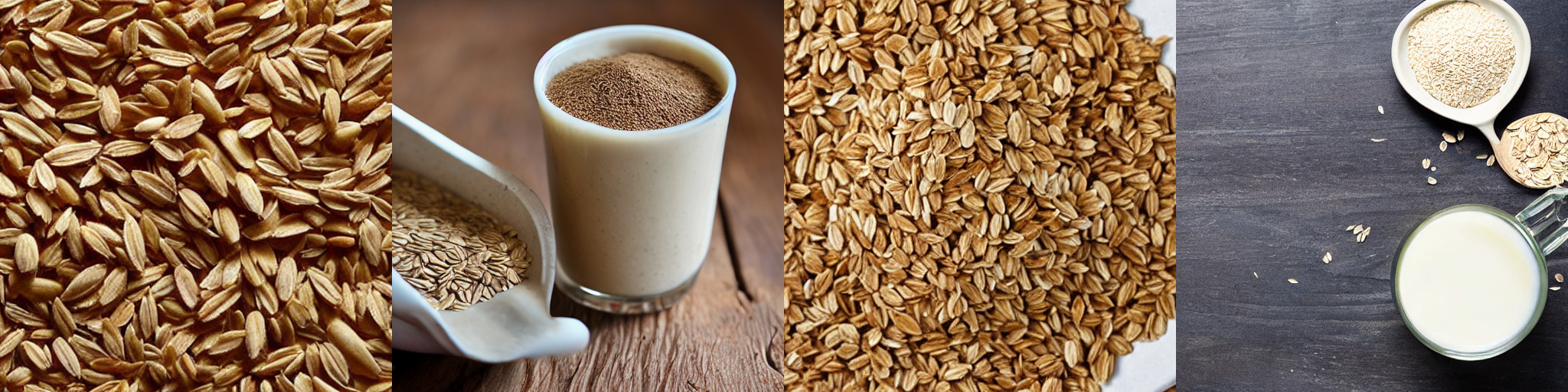}
    \\  \addlinespace
    Coffee {\color{red} \textbf{with}} creamer. \textbf{(Par)} & \includegraphics[width=\linewidth]{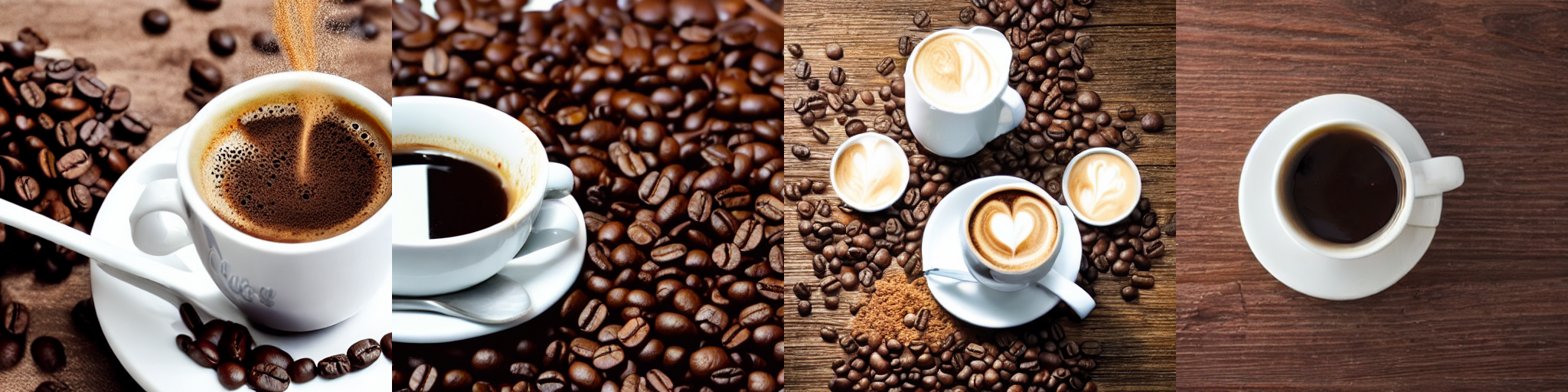}
    \\  \addlinespace
    Coffee {\color{red} \textbf{without}} creamer. \textbf{(Par)} & \includegraphics[width=\linewidth]{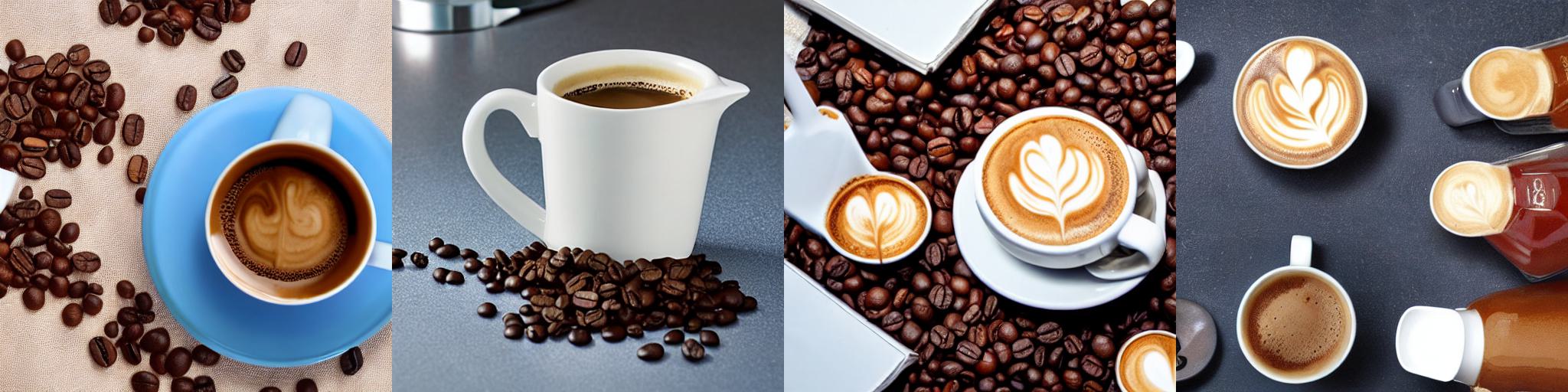}
    \\ \bottomrule 
    \end{tabular}
    \vspace{4mm}
    \caption{Images generated by Stable Diffusion Model for prompts with prepositions. This table covers the last two subcategory of prepositions: preposition of movement (PoE) and Particles (Par). The prepositions are colored in red. }
    \label{fig:app_prepositions_2}
\end{table}

\subsection{Determiners and Qualifiers}
Determiners is composed of three types: articles (AR), cardinal numerals (CN), and quantifiers (QUAN). Table~\ref{fig:app_determiner_1} and Table~\ref{fig:app_determiner_2} demonstrates some examples in this category. Another interesting type of function words is qualifiers (QUAL), where some creative prompts and images are presented in Table~\ref{fig:app_determiner_2}.

\begin{table}[H]
\centering
    \begin{tabular}[t]{@{} c m{0.46\linewidth} @{}}
    \textbf{Prompts} & \centering\arraybackslash\textbf{Images} \\ 
    \midrule \addlinespace
    {\color{red} \textbf{A}} dice rolled on the table. \textbf{(AR)} & \includegraphics[width=\linewidth]{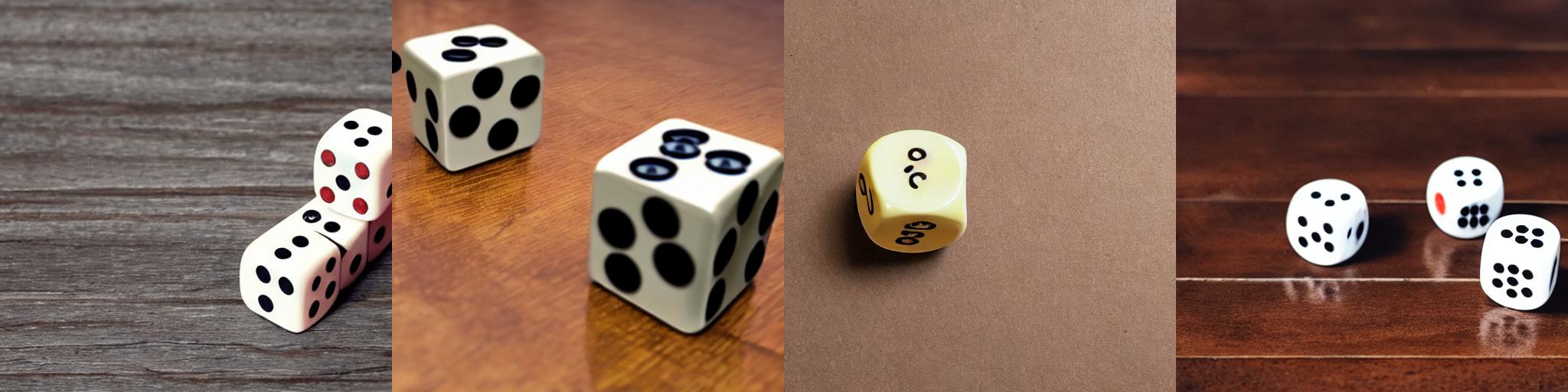} 
    \\ \addlinespace
    {\color{red} \textbf{An}} aircraft performing an air show. \textbf{(AR)} & \includegraphics[width=\linewidth]{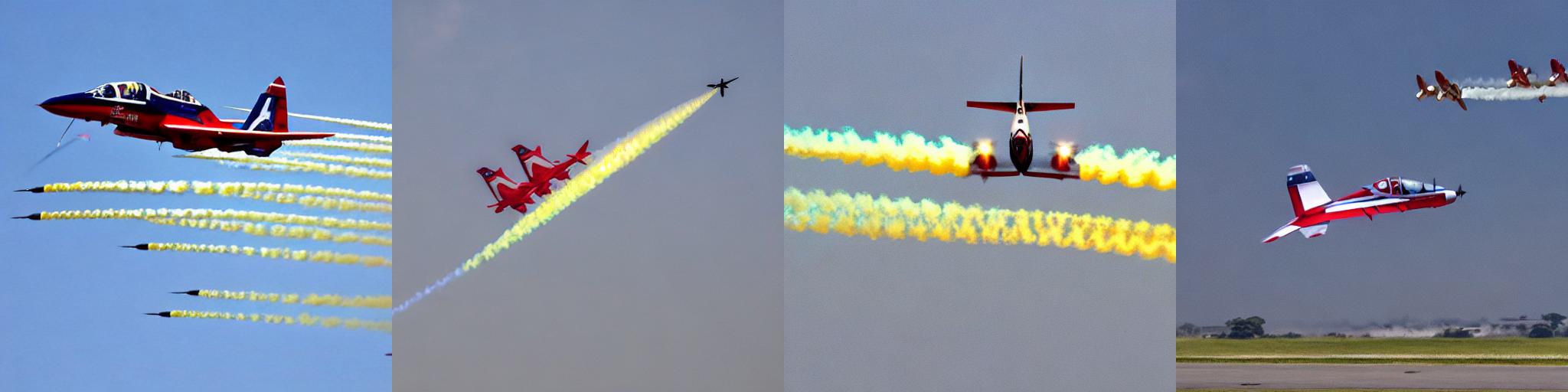}  
    \\  \addlinespace
    {\color{red} \textbf{The}} sheep is eating grass. \textbf{(AR)} & \includegraphics[width=\linewidth]{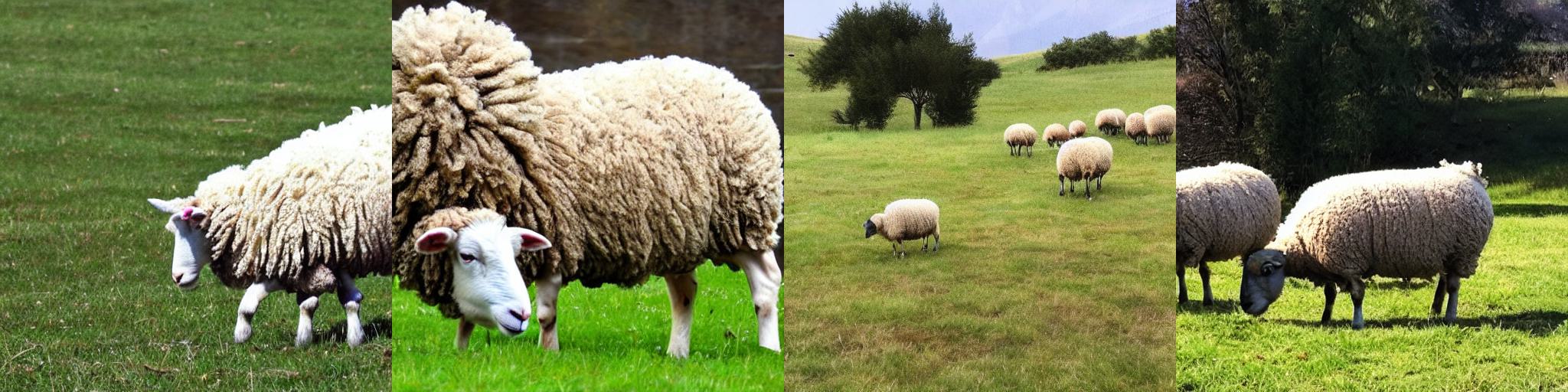}
    \\ \addlinespace
    \midrule
    \addlinespace
    There is {\color{red} \textbf{one}} orange in the photo. \textbf{(CN)} & \includegraphics[width=\linewidth]{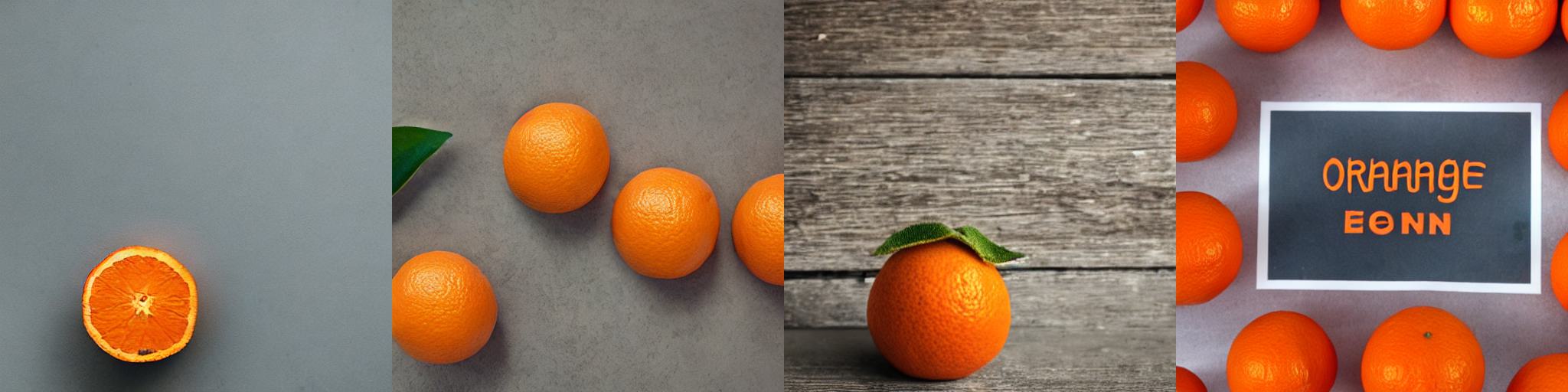} 
    \\ \addlinespace
    There are {\color{red} \textbf{three}} oranges in the photo. \textbf{(CN)} & \includegraphics[width=\linewidth]{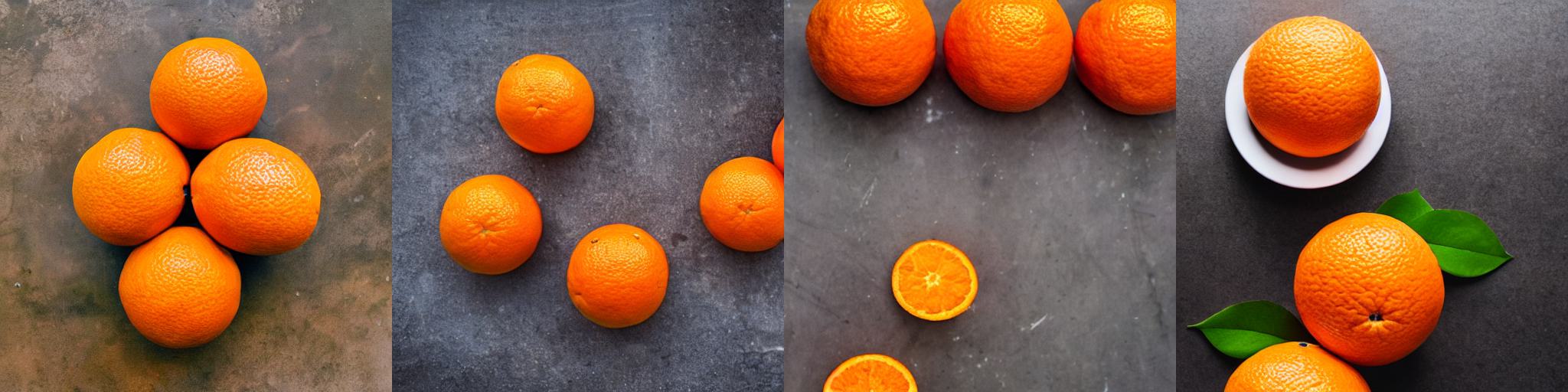}   
    \\ \addlinespace
    There are {\color{red} \textbf{ten}} oranges in the photo. \textbf{(CN)} & \includegraphics[width=\linewidth]{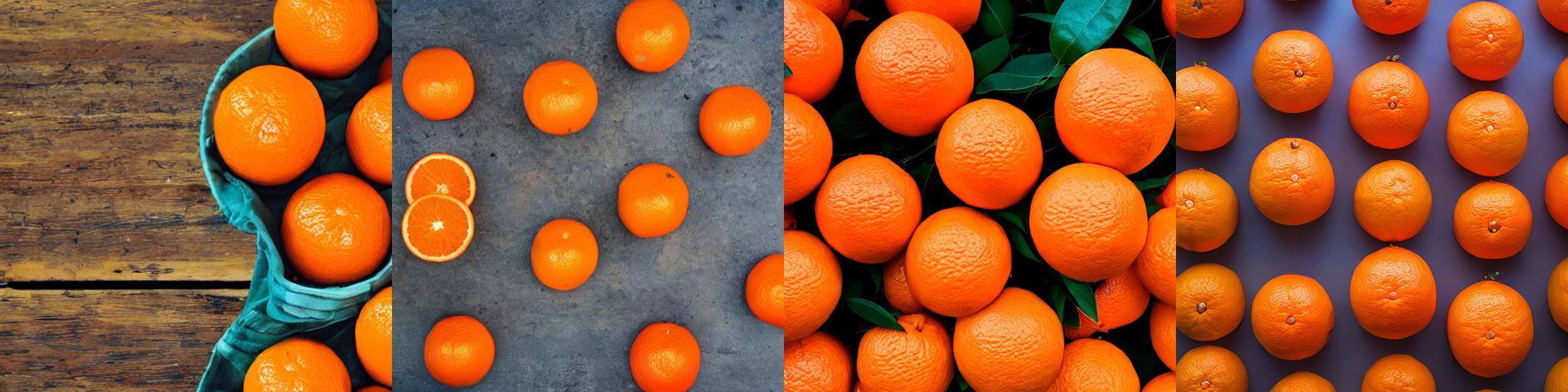}  
    \\  \addlinespace
    \midrule
    \addlinespace
    There is {\color{red} \textbf{little}} milk in the bottle. \textbf{(Quan)} & \includegraphics[width=\linewidth]{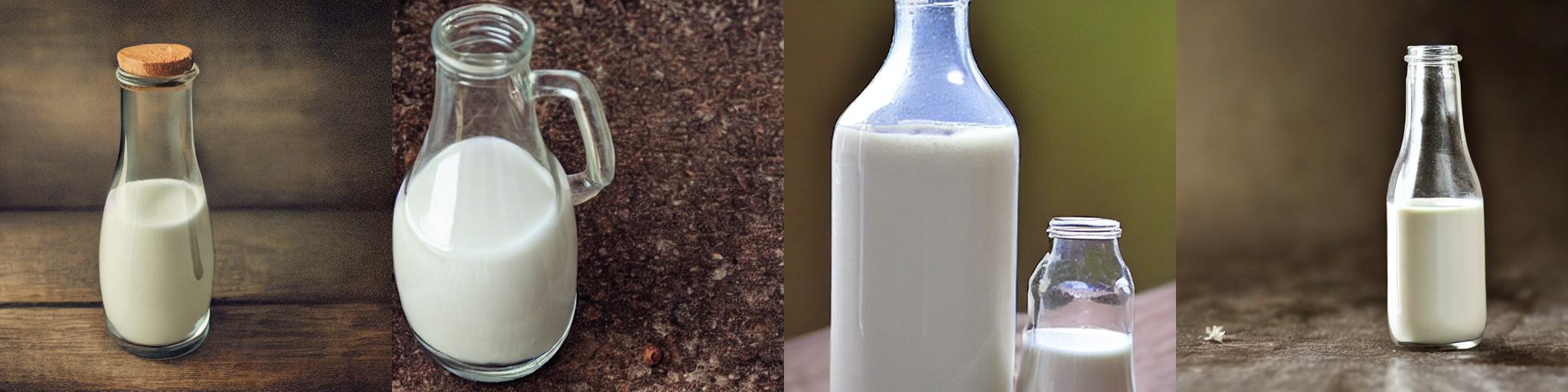}  
    \\  \addlinespace
    There is {\color{red} \textbf{a lot of}} milk in the bottle. \textbf{(Quan)} & \includegraphics[width=\linewidth]{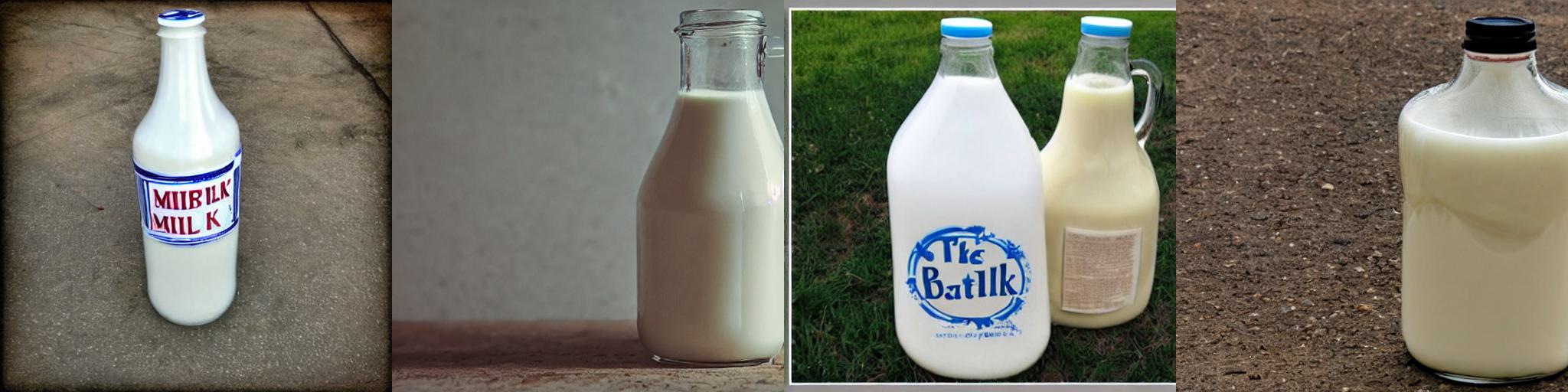}  
    \\ \bottomrule 
    \end{tabular}
    \vspace{4mm}
    \caption{Images generated by Stable Diffusion Model for prompts with determiners. This table covers the first three subcategories of determiners, Article (AR), Numeral Cardinals (NC) and Quantifiers (Quan). The determiner words are colored in red. }
    \label{fig:app_determiner_1}
\end{table}

\begin{table}[H]
\centering
    \begin{tabular}[t]{@{} c m{0.48\linewidth} @{}}
    \textbf{Prompts} & \centering\arraybackslash\textbf{Images} \\ 
    \midrule \addlinespace
    There are {\color{red} \textbf{few}} bananas on the table. \textbf{(Quan)} & \includegraphics[width=\linewidth]{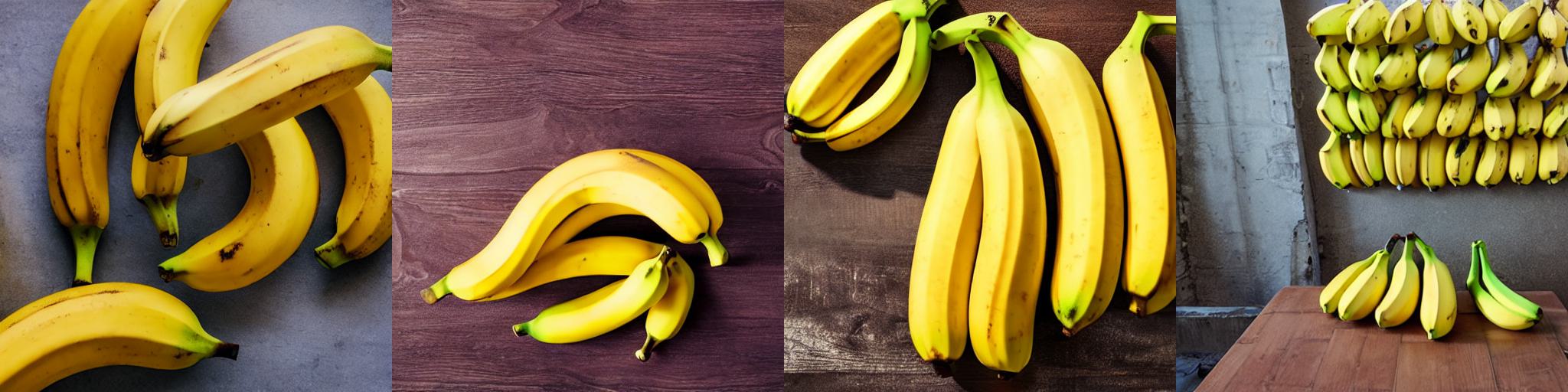}
    \\  \addlinespace
    There are {\color{red} \textbf{many}} bananas on the table. \textbf{(Quan)} & \includegraphics[width=\linewidth]{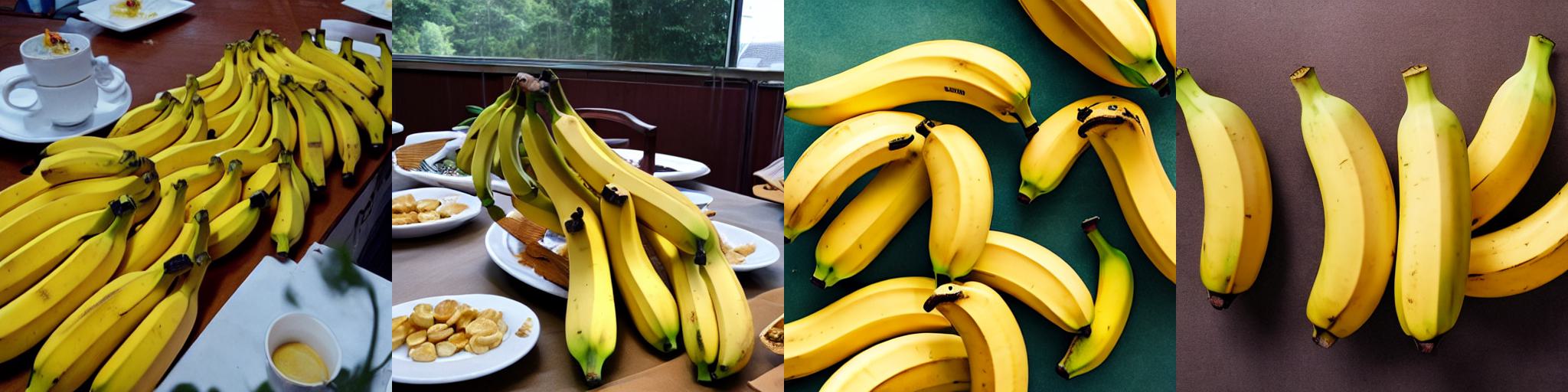}
    \\  \addlinespace
    {\color{red} \textbf{Few}} oranges in the basket. \textbf{(Quan)} & \includegraphics[width=\linewidth]{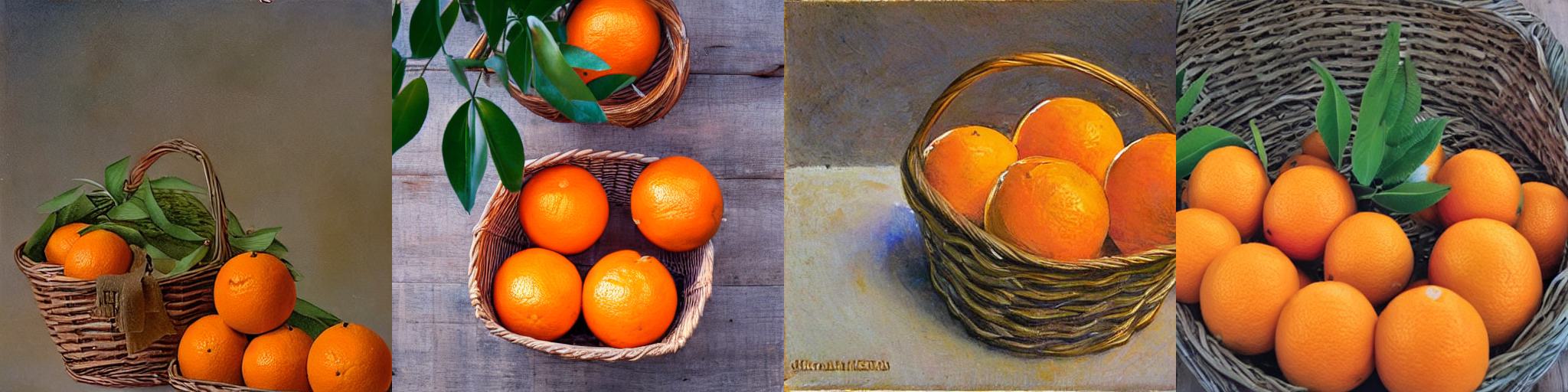}
    \\  \addlinespace
    {\color{red} \textbf{Many}} oranges in the basket. \textbf{(Quan)} & \includegraphics[width=\linewidth]{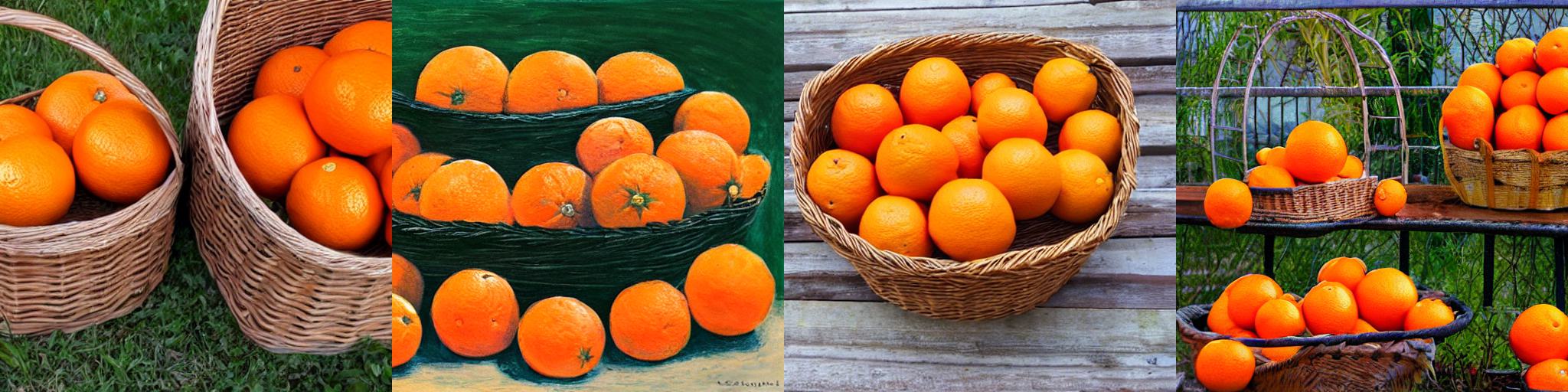}
    \\ \addlinespace
    \midrule
    \addlinespace
    The sky is {\color{red} \textbf{not}} green. \textbf{(Qual)} & \includegraphics[width=\linewidth]{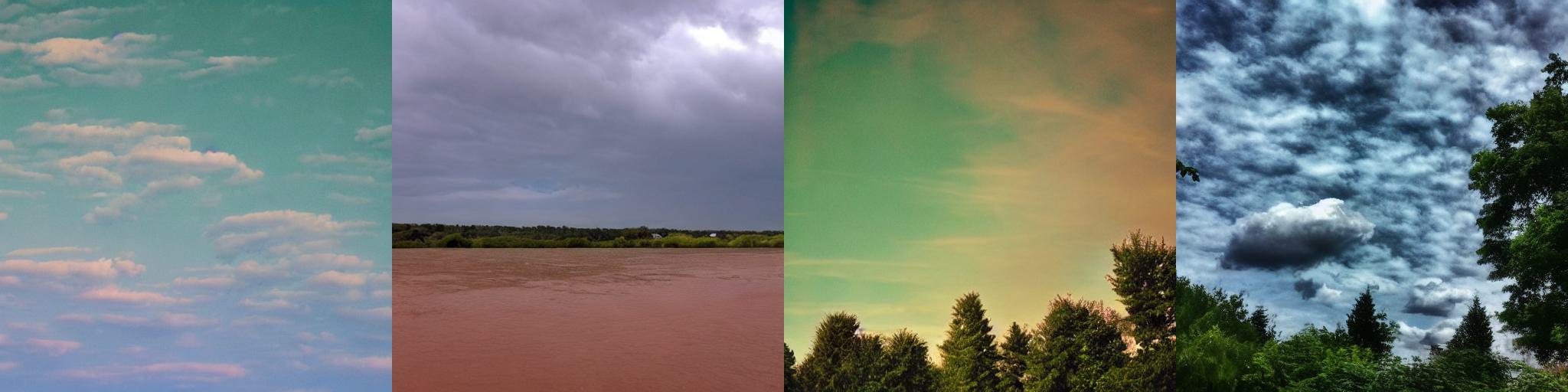} 
    \\ \addlinespace
    The sky is {\color{red} \textbf{never}} green. \textbf{(Qual)} & \includegraphics[width=\linewidth]{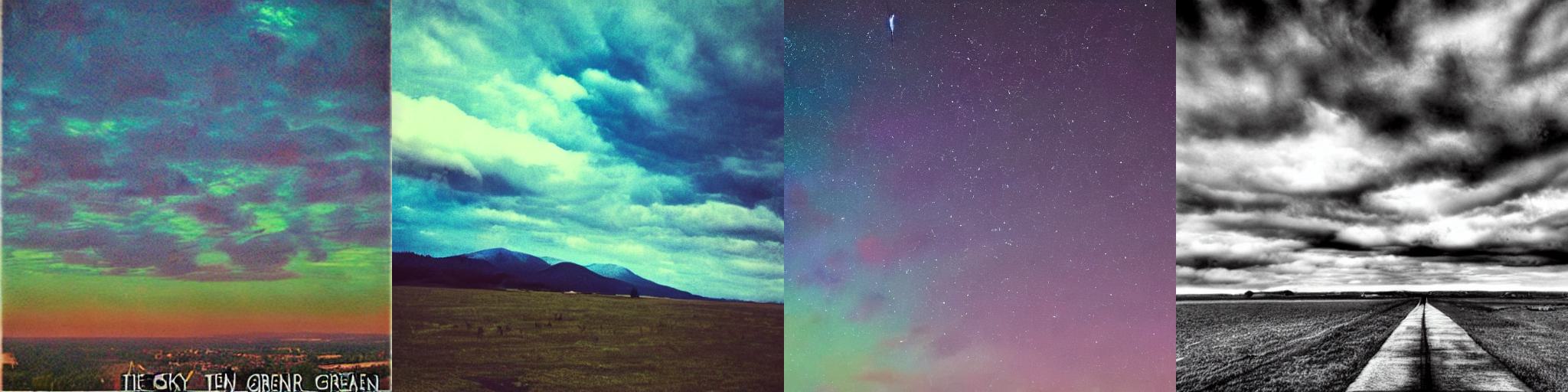}   
    \\ \addlinespace
    The sky is {\color{red} \textbf{always}} green. \textbf{(Qual)} & \includegraphics[width=\linewidth]{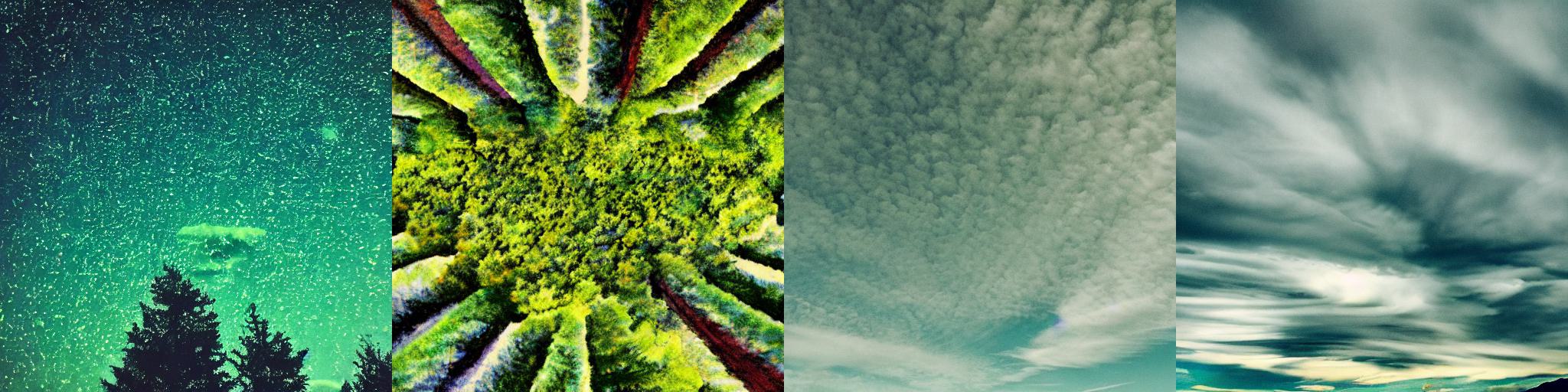}  
    \\ \addlinespace
    The plate is {\color{red} \textbf{a little}} dirty. \textbf{(Qual)}  & \includegraphics[width=\linewidth]{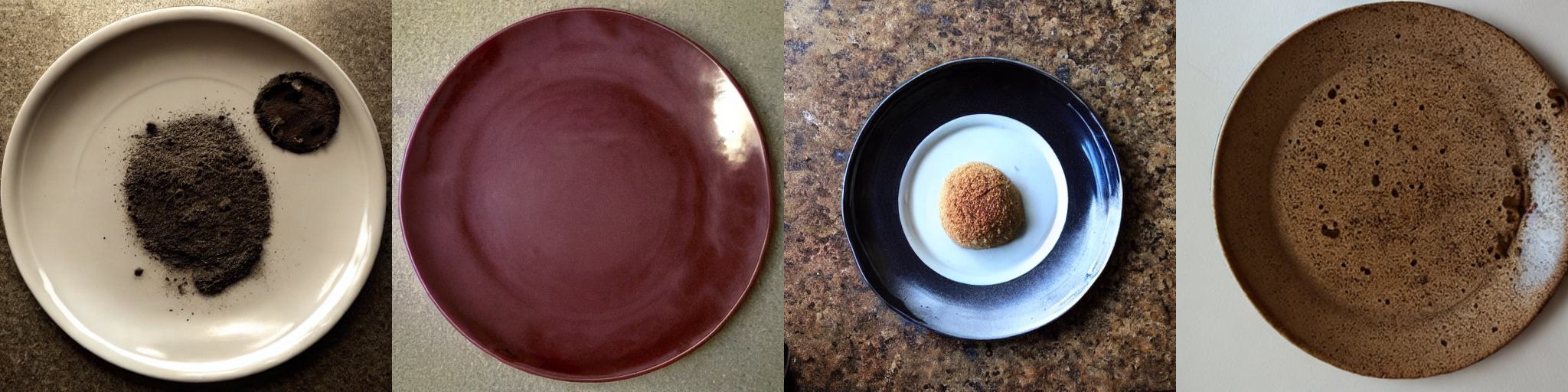}  
    \\  \addlinespace
    The plate is {\color{red} \textbf{very}} dirty. \textbf{(Qual)} & \includegraphics[width=\linewidth]{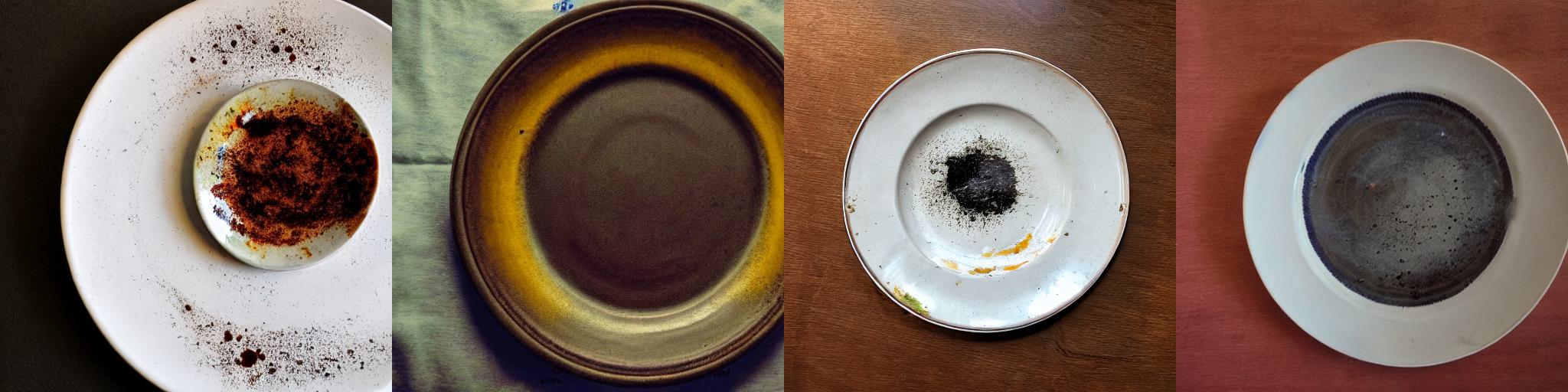}  
    \\ \bottomrule 
    \end{tabular}
    \vspace{4mm}
    \caption{Images generated by Stable Diffusion Model for prompts with determiners. This table covers two subcategories of determiners:  Quantifiers (Quan) and Qualifiers. (Qual). The determiner words are colored in red. }
    \label{fig:app_determiner_2}
\end{table}


\subsection{Interrogatives and Relatives}

Interrogatives and relatives both represent function words that are `Wh-' alike, such as What, Where or Who. Interrogatives (Int) usually raises a question while relatives (Rel) do not. More prompts and generated images are provided in Table~\ref{fig:app_interrogative_1}.

\begin{table}[H]
\centering
    \begin{tabular}[t]{@{} c m{0.45\linewidth} @{}}
    \textbf{Prompts} & \centering\arraybackslash\textbf{Images} \\ 
    \midrule \addlinespace
    {\color{red} \textbf{Where}} did we have the coffee? \textbf{(Int)} & \includegraphics[width=\linewidth]{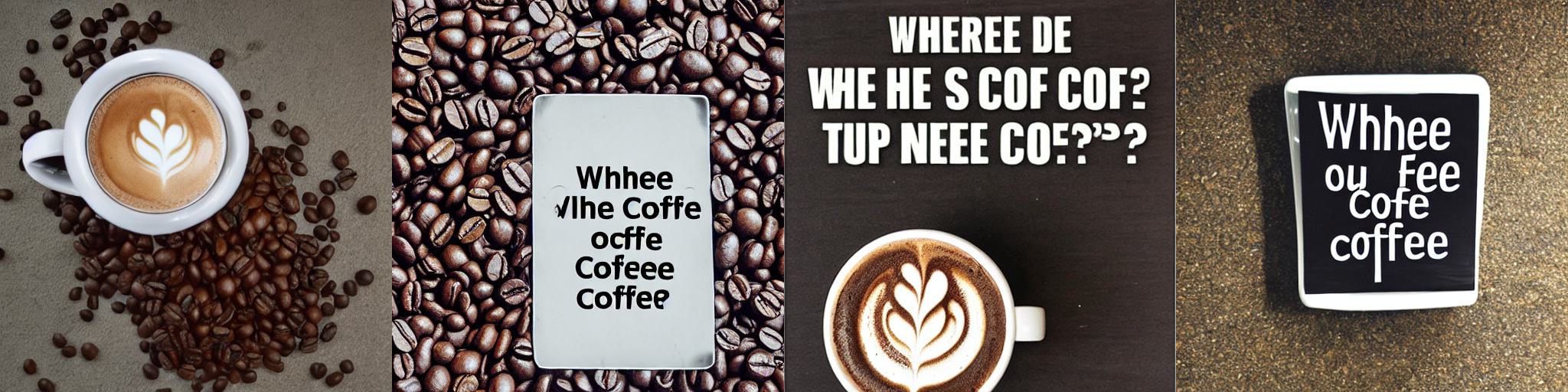} 
    \\ \addlinespace
    {\color{red} \textbf{What}} are we eating for dinner? \textbf{(Int)}& \includegraphics[width=\linewidth]{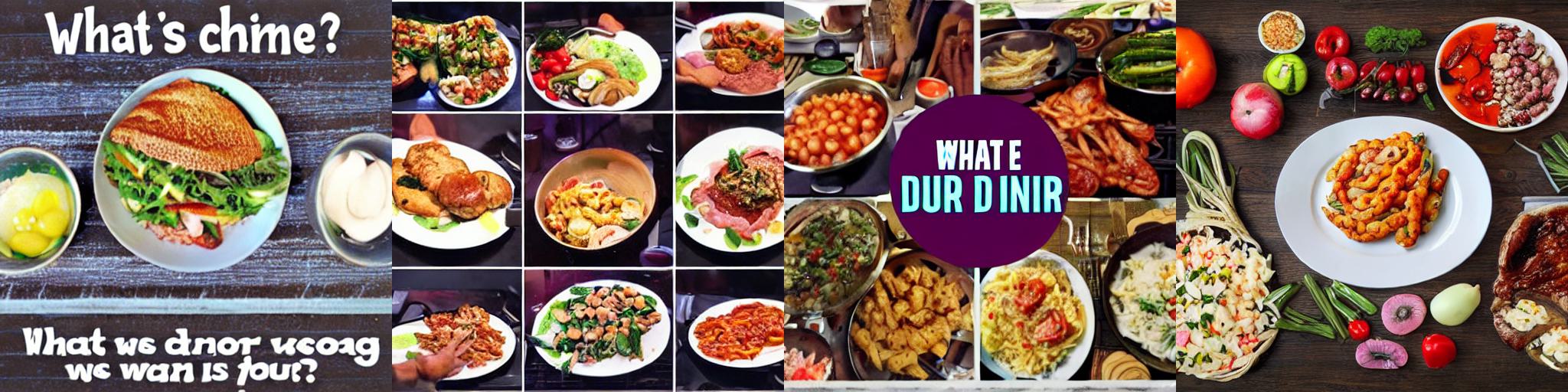}  
    \\ \addlinespace
    {\color{red} \textbf{Who}} are we eating with for dinner? \textbf{(Int)} & \includegraphics[width=\linewidth]{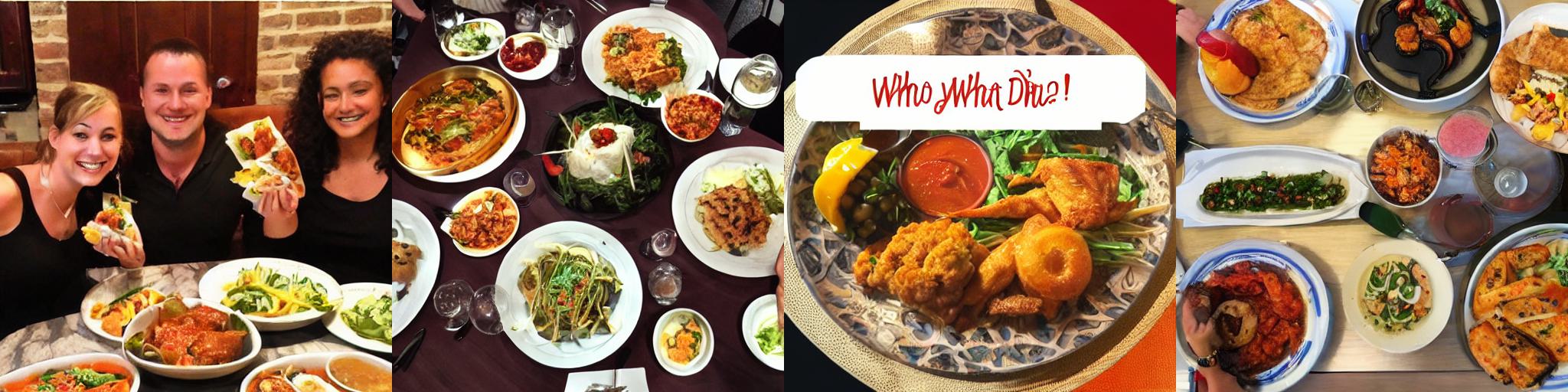} 
    \\ \addlinespace
    {\color{red} \textbf{What}} is climbing the tree? \textbf{(Int)}& \includegraphics[width=\linewidth]{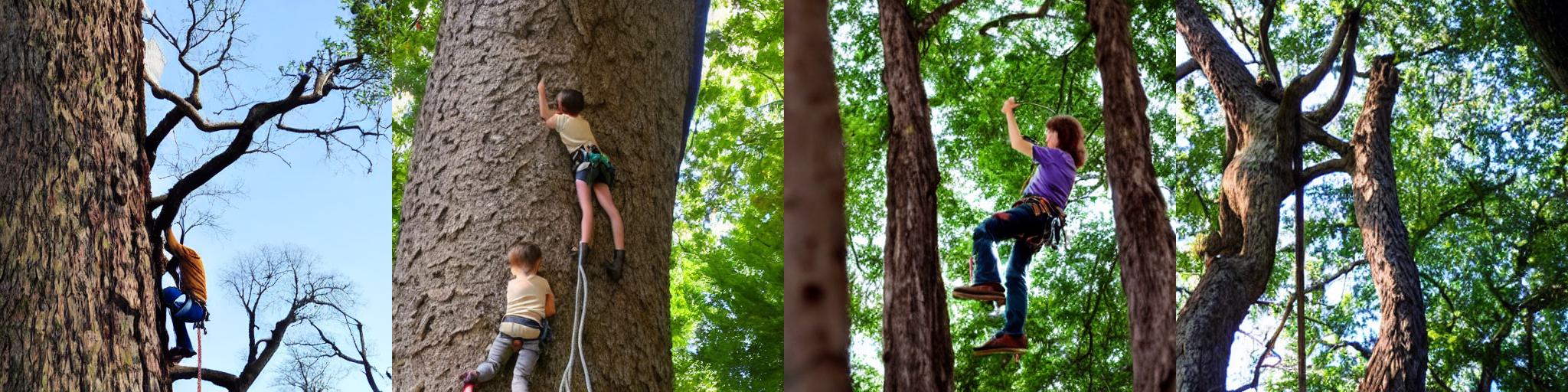} 
    \\ \addlinespace
    {\color{red} \textbf{Who}} is climbing the tree? \textbf{(Int)}& \includegraphics[width=\linewidth]{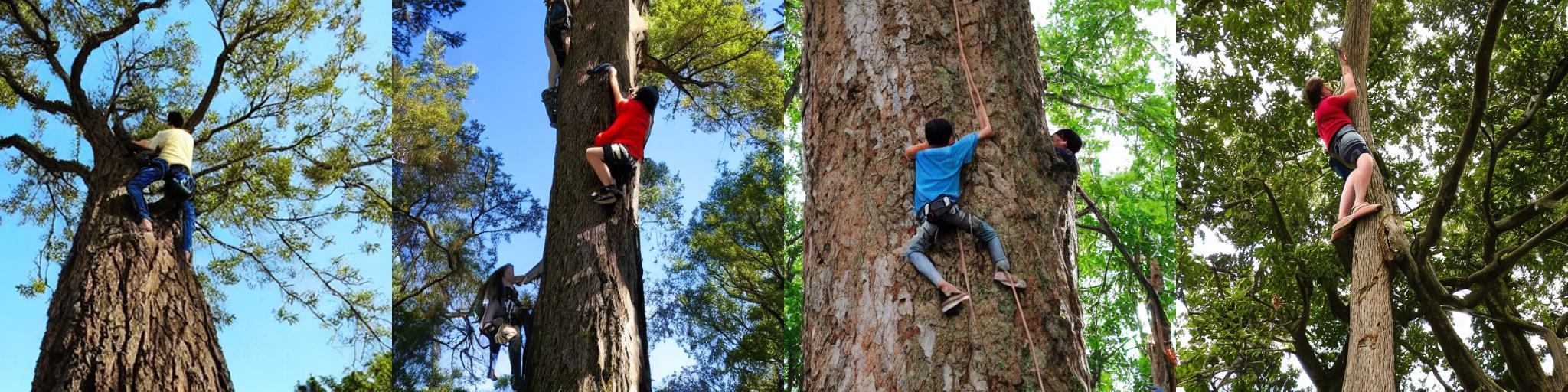} 
    \\ \addlinespace
    \midrule
    \addlinespace
    Draw a picture of {\color{red} \textbf{what}} we had for dinner. \textbf{(Rel)}& \includegraphics[width=\linewidth]{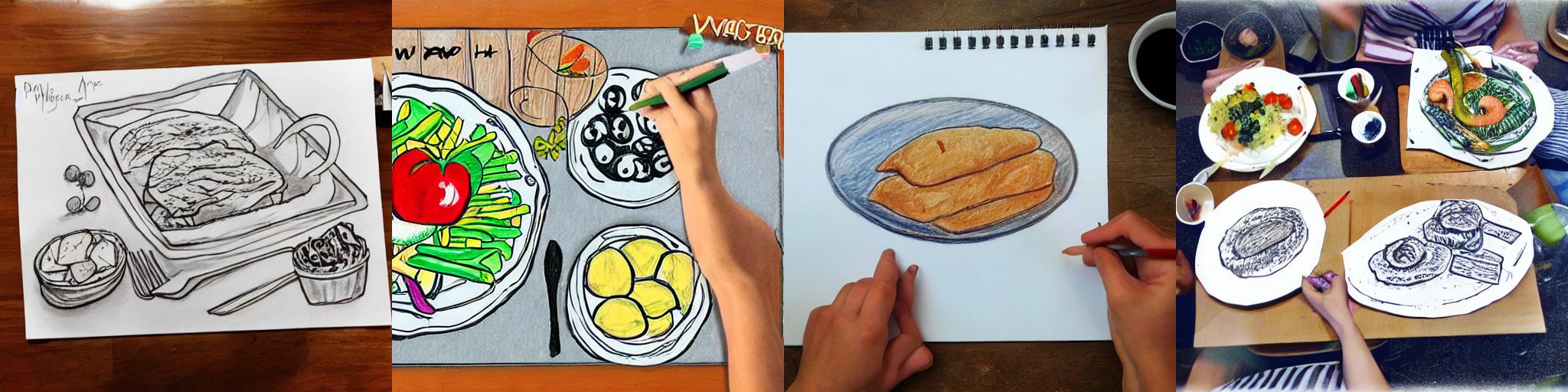} 
    \\ \addlinespace
    Draw a picture of {\color{red} \textbf{who}} is climbing the tree. \textbf{(Rel)}& \includegraphics[width=\linewidth]{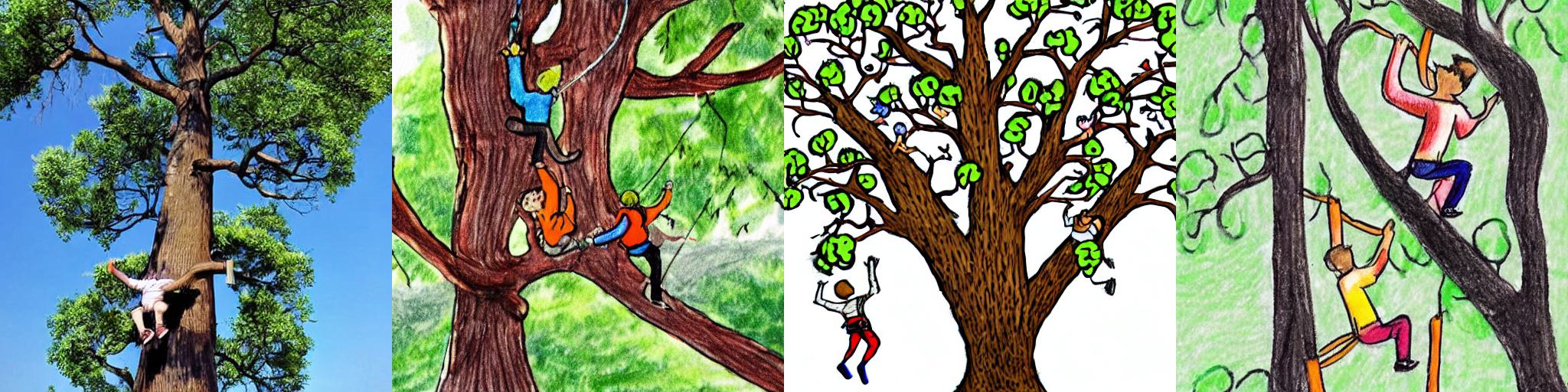}
    \\ \addlinespace
    Draw a picture of {\color{red} \textbf{where}} we had our coffee. \textbf{(Rel)}& \includegraphics[width=\linewidth]{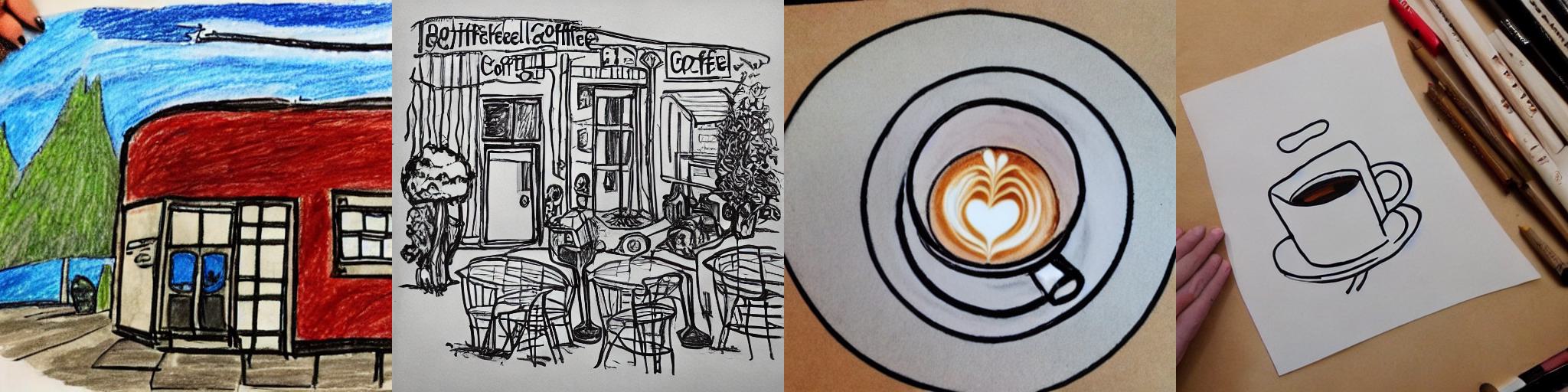} 
    \\ \bottomrule 
    \end{tabular}
    \vspace{4mm}
    \caption{Images generated by Stable Diffusion Model for prompts with interrogatives (Int) and relatives (Rel). The interrogative words are colored in red. }
    \label{fig:app_interrogative_1}
\end{table}

\subsection{Conjunctions}
In this section, it provides some supplementary prompts and images in Table~\ref{fig:app_conjunction_1} to show how stable diffusion model could perform with conjunctions. 

\begin{table}[H]
\centering
    \begin{tabular}[t]{@{} c m{0.5\linewidth} @{}}
    \textbf{Prompts} & \centering\arraybackslash\textbf{Images} \\ 
    \midrule \addlinespace
    Generate an image of {\color{red} \textbf{either}} an apple {\color{red} \textbf{or}} an orange. & \includegraphics[width=\linewidth]{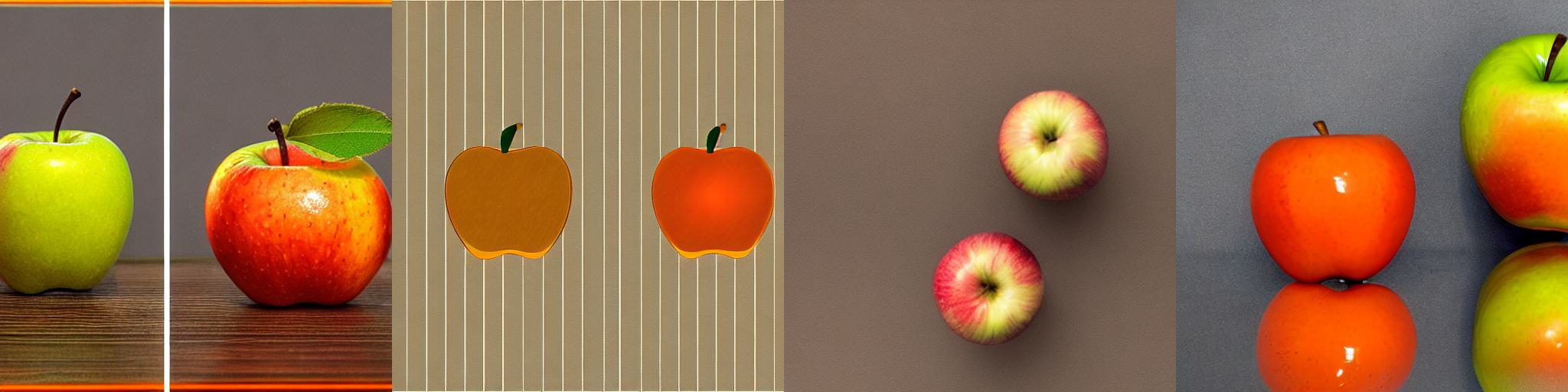} 
    \\ \addlinespace
    Generate an image of apple {\color{red} \textbf{and}} orange. & \includegraphics[width=\linewidth]{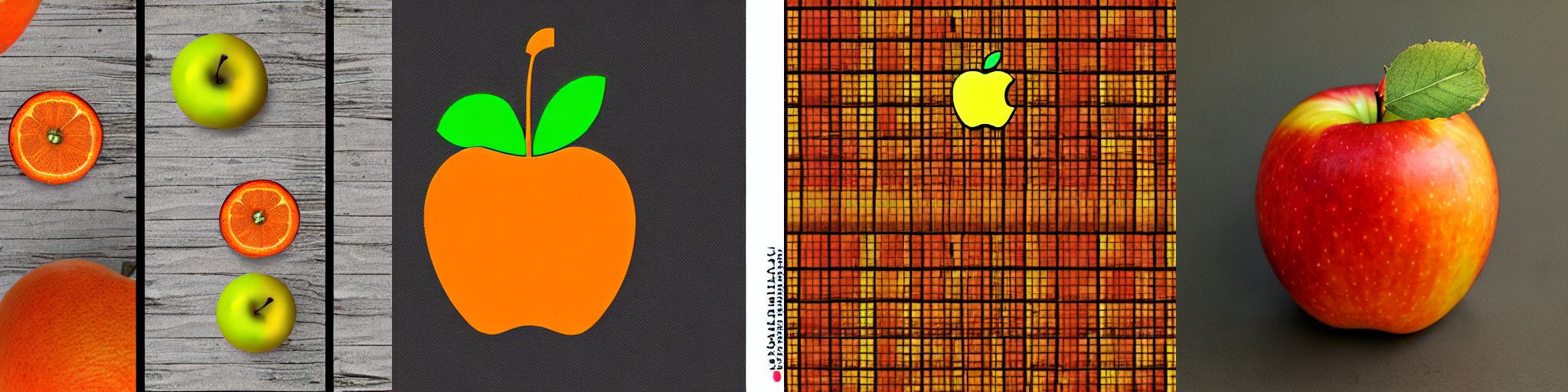}   
    \\ \addlinespace
    Draw an apple {\color{red} \textbf{and}} an orange. & \includegraphics[width=\linewidth]{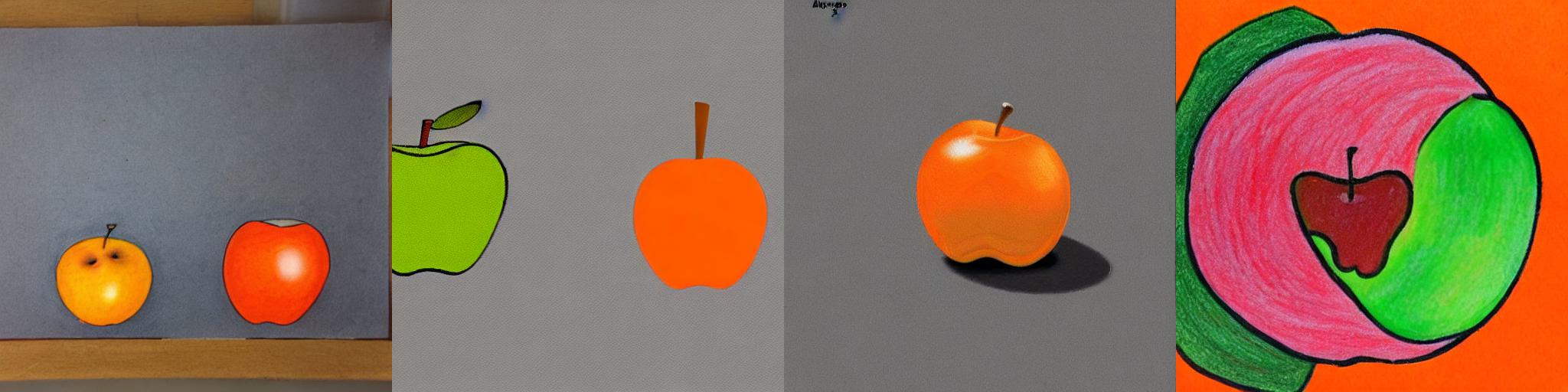}  
    \\ \addlinespace
    Draw {\color{red} \textbf{either}} an apple {\color{red} \textbf{or}} an orange. & \includegraphics[width=\linewidth]{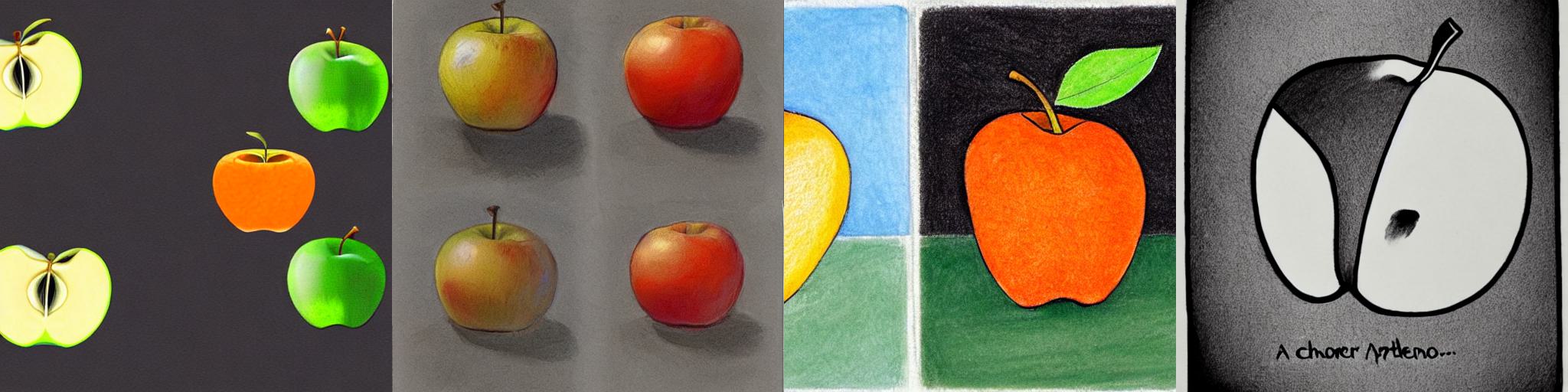}  
    \\  \addlinespace
    It was sunny {\color{red} \textbf{but}} now it is raining. & \includegraphics[width=\linewidth]{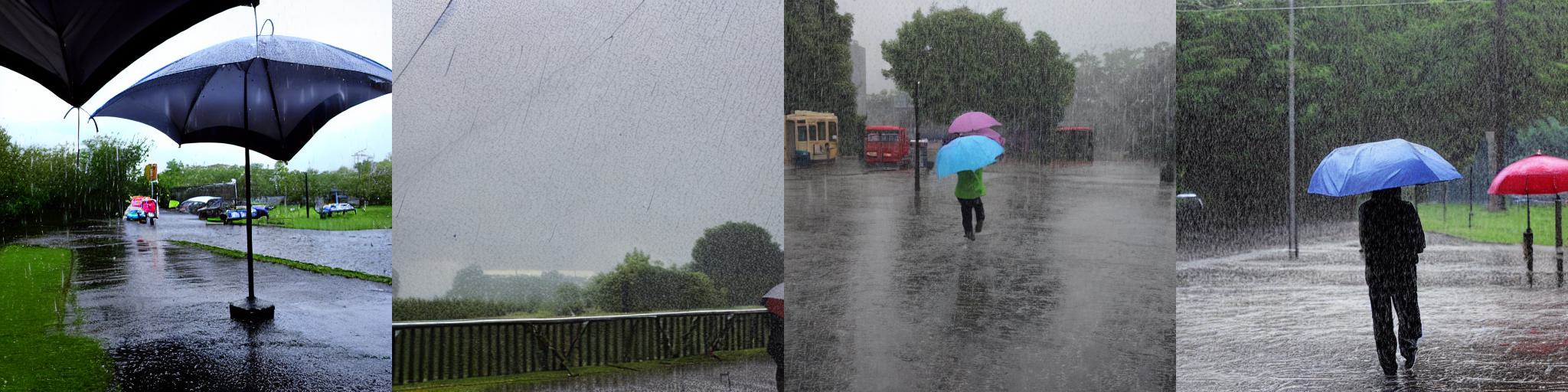}  
    \\  \addlinespace
    It was rainy {\color{red} \textbf{but}} now it is sunny. & \includegraphics[width=\linewidth]{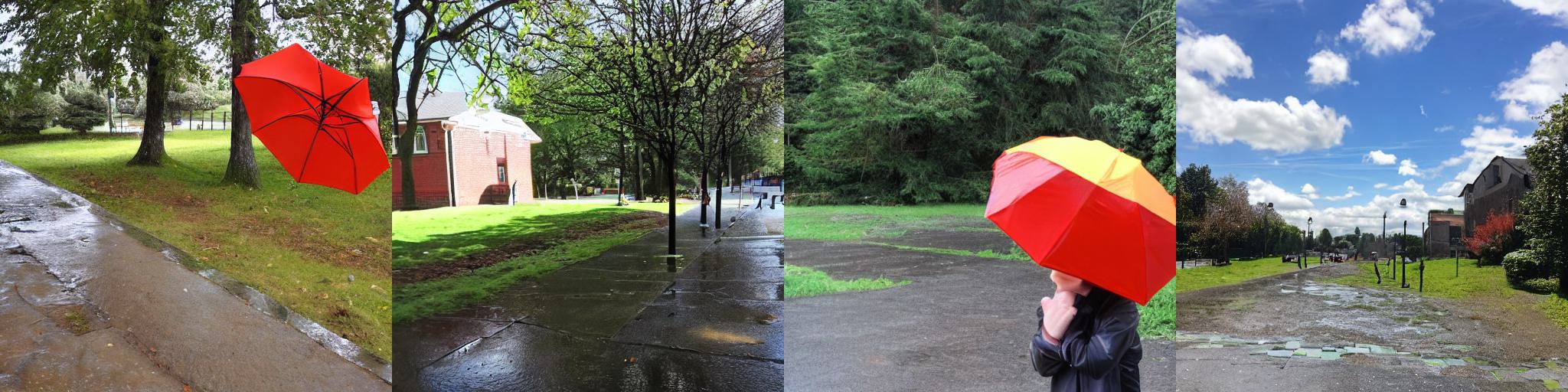}
    \\ \addlinespace
    The bottle is taller {\color{red} \textbf{than}} the cup. & \includegraphics[width=\linewidth]{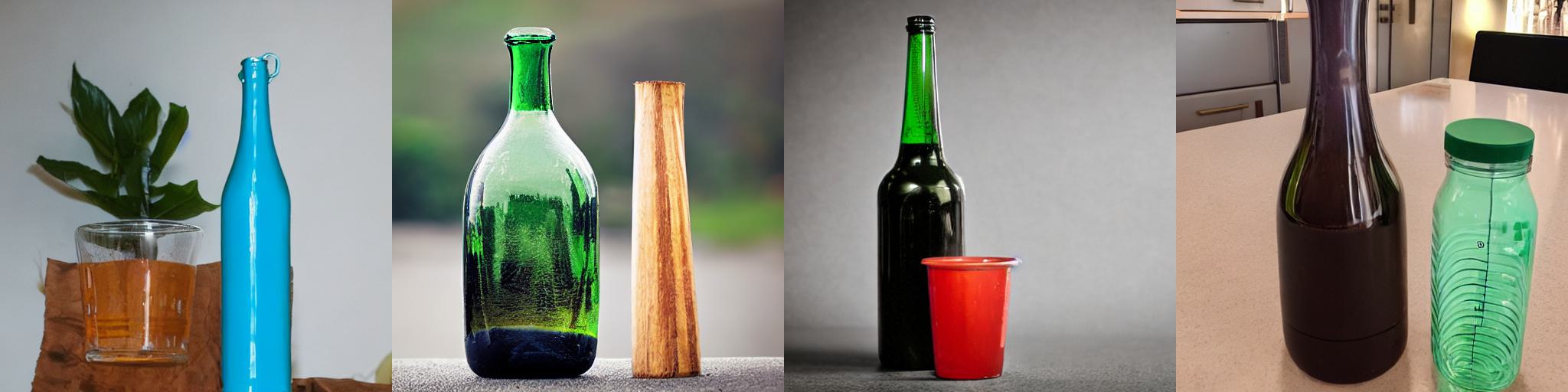}  
    \\ \addlinespace
    The cup is taller {\color{red} \textbf{than}} the bottle. & \includegraphics[width=\linewidth]{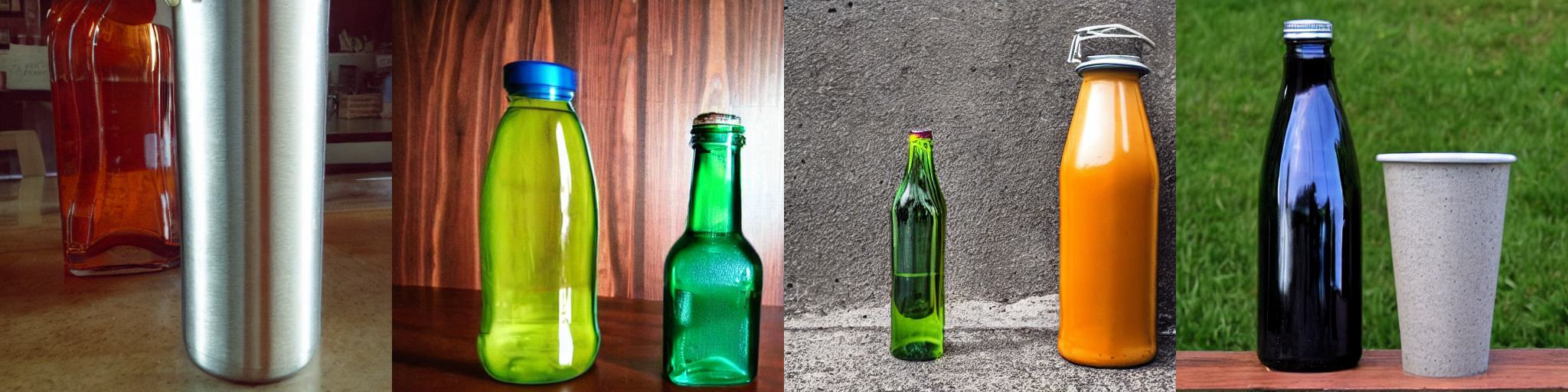}  
    \\ \bottomrule 
    \end{tabular}
    \vspace{4mm}
    \caption{Images generated by Stable Diffusion Model for prompts with conjunctions. The conjunction words are colored in red. }
    \label{fig:app_conjunction_1}
\end{table}

\end{document}